\documentclass{article}

\PassOptionsToPackage{numbers, sort&compress}{natbib}
\usepackage[final]{neurips_2024}
\usepackage[utf8]{inputenc} % allow utf-8 input
\usepackage[T1]{fontenc}    % use 8-bit T1 fonts
\usepackage{hyperref}       % hyperlinks
\usepackage{url}            % simple URL typesetting
\usepackage{booktabs}       % professional-quality tables
\usepackage{amsfonts}       % blackboard math symbols
\usepackage{nicefrac}       % compact symbols for 1/2, etc.
\usepackage{microtype}      % microtypography
\usepackage[dvipsnames]{xcolor}         % colors
\usepackage{caption}
\usepackage{algorithm}
\usepackage{algorithmic}
\usepackage{bm}
\usepackage{graphicx}
\usepackage{amsmath}
\usepackage{amssymb}
\usepackage{mathtools}
\usepackage{amsthm}
\usepackage{multirow}
\usepackage{tabularx}
\usepackage{subcaption}
\usepackage{wrapfig}

\usepackage{tcolorbox}
\usepackage[formats]{listings} % code formatting package
\usepackage{lscape} % horizontal formatting for code
\usepackage[nottoc]{tocbibind}
\usepackage{minitoc}
\usepackage{bbm}
\usepackage{svg}
\usepackage{siunitx}
\usepackage[capitalize,noabbrev]{cleveref}
\usepackage{hhline}

\usepackage{enumitem}
\setlist[itemize]{leftmargin=5mm}

\usepackage{pdfpages}

\newcommand{\mytitle}{Diffusion Imitation from Observation}

\definecolor{URL}{HTML}{0000EE}
\definecolor{RayColor}{rgb}{0,0.08,1}

\newcommand{\sun}[1]{}
\newcommand{\yang}[1]{}
\newcommand{\ray}[1]{}
\newcommand{\daijie}[1]{}
\newcommand{\cml}[1]{}

\newcommand{\Skip}[1]{}

\newcommand{\pointmaze}{\textsc{PointMaze}}
\newcommand{\antmaze}{\textsc{AntMaze}}
\newcommand{\walker}{\textsc{Walker}}
\newcommand{\fetchpush}{\textsc{FetchPush}}
\newcommand{\adroitdoor}{\textsc{AdroitDoor}}
\newcommand{\carracing}{\textsc{CarRacing}}
\newcommand{\sine}{\textsc{Sine}}
\newcommand{\drawerclose}{\textsc{CloseDrawer}}
\newcommand{\frankakitchen}{\textsc{OpenMicrowave}}

\newcommand{\ie}{\textit{i}.\textit{e}.,\ }
\newcommand{\eg}{\textit{e}.\textit{g}.,\ }
\newcommand{\method}{DIFO}
\newcommand{\methodFull}{Diffusion Imitation from Observation}

\newcommand{\expertDenoisingLossspace}{\(\mathcal{L}_{d}^{E}\) }
\newcommand{\agentDenoisingLoss}{\(\mathcal{L}_{d}^{A}\)}

\newcommand{\myfig}[1]{Figure~\ref{#1}}

\newcommand{\myeq}[1]{Eq.~\ref{#1}}

\newcommand{\dotieconcat}[2]{% auxiliary macro, don't use it directly
  \text{\raisebox{.8ex}{$\smallfrown$}}%
}

\makeatletter
\newcommand\dslfontsize{\@setfontsize\dslfontsize\@viipt\@viiipt}
\makeatother

\newcommand{\myparagraph}[1]{\noindent \textbf{#1.}}
\newcommand{\vspacesection}[1]{\vspace{-0.00em}
\section{#1}
\vspace{-0.0em}}
\newcommand{\vspacesubsection}[1]{\vspace{-0.0em}
\subsection{#1}
\vspace{-0.0em}}

\newcommand\blfootnote[1]{%
  \begingroup
  \renewcommand\thefootnote{}\footnote{#1}%
  \addtocounter{footnote}{-1}%
  \endgroup
}

\let\svthefootnote\thefootnote
\newcommand\freefootnote[1]{%
  \let\thefootnote\relax%
  \footnotetext{#1}%
  \let\thefootnote\svthefootnote%
}

\usepackage{color}
\usepackage{xcolor}
\definecolor{codegray}{rgb}{0.5,0.5,0.5}
\title{\mytitle}

\author{%
  Bo-Ruei Huang
  \quad Chun-Kai Yang
  \quad Chun-Mao Lai
  \quad Dai-Jie Wu
  \quad Shao-Hua Sun \\
  Department of Electrical Engineering, National Taiwan University
}

\begin{document}

\maketitle

\begin{abstract}
Learning from observation (LfO) aims to imitate experts by learning from state-only demonstrations without requiring action labels. 
Existing adversarial imitation learning approaches learn a generator agent policy to produce state transitions that are indistinguishable to a discriminator that learns to classify agent and expert state transitions. 
Despite its simplicity in formulation, these methods are often sensitive to hyperparameters and brittle to train. Motivated by the recent success of diffusion models in generative modeling, we propose to integrate a diffusion model into the adversarial imitation learning from observation framework. 
Specifically, we employ a diffusion model to capture expert and agent transitions by generating the next state, given the current state. 
Then, we reformulate the learning objective to train the diffusion model as a binary classifier and use it to provide ``realness'' rewards for policy learning. 
Our proposed framework, Diffusion Imitation from Observation (DIFO), demonstrates superior performance in various continuous control domains, including navigation, locomotion, manipulation, and games.
Project page: \color{URL}{\url{https://nturobotlearninglab.github.io/DIFO}}
% \url{https://nturobotlearninglab.github.io/DIFO/}

\begingroup
\hypersetup{colorlinks=false, linkcolor=black}
\hypersetup{pdfborder={0 0 0}}
\blfootnote{Correspondence to: Shao-Hua Sun <shaohuas@ntu.edu.tw>}
\endgroup
\end{abstract}

\vspace{-2.0em}
\vspacesection{Introduction}
\vspace{-0.5em}
\label{sec:intro}

Learning from demonstration (LfD)~\citep{schaal1997learning, pomerleau1991efficient_bc, Ziebart2008MaximumEI, hussein2017imitation, osa2018algorithmic} aims to acquire policies that can perform desired skills by imitating expert trajectories represented as sequences of state-action pairs,
eliminating the necessity of reward functions. 
Recent advancements in LfD have enabled the deployment of reliable and robust learned policies in various domains, such as robot learning~\citep{johns2021coarse, Giusti2016Visual, gupta2016learning, jang2022BCZ}, strategy games~\citep{harmer2018imitation, Palma2011Combining, Shih2022Conditional}, and self-driving~\citep{Samak2021Robust, choi2021trajgail, scheel2022urbanDriver, Coelho_Oliveira_Santos_2024}.
LfD's dependence on accurately labeled actions remains a substantial limitation, particularly in scenarios where obtaining expert actions is challenging or costly.
Moreover, most LfD methods assume that the demonstrator and imitator share the same embodiment, inherently preventing cross-embodiment imitation.

To address these issues, learning from observation (LfO) methods~\cite{torabi2018generative, zhu2020off, 2024lapo} seek to imitate experts from state-only sequences, thereby removing the need for action labels and allowing learning from experts with different embodiments.
\citet{2024lapo, torabi2018behavioral, yang2019imitation} proposed learning inverse dynamic models (IDMs) that can infer action labels from state sequences and subsequently reformulate LfO as LfD. Nevertheless, acquiring sufficiently aligned data with the expert's data distribution to train IDMs remains an unresolved challenge. On the other hand, adversarial imitation learning (AIL)~\cite{torabi2018generative, kostrikov2018discriminatoractorcritic, xiao2019wasserstein} employs a generator policy learning to imitate an expert, while a discriminator differentiates between the data produced by the policy and the actual expert data.
Despite its simplicity in formulation, AIL methods can be brittle to learn and are often sensitive to hyperparameters~\citep{fu2018learning, barde2020adversarial}.

Recent works have explored leveraging diffusion models' ability in generative modeling and achieved encouraging results in imitation learning and planning \cite{janner2022diffuser, mishra2023generative, ko2024learning}.
For example, diffusion policies~\citep{pearce2022imitating, chi2023diffusionpolicy} learn to denoise actions with injected noises conditioned on states, allowing for modeling multimodal expert behaviors.
Moreover, \citet{chen2024diffusion} proposed to model expert state-action pairs with a diffusion model and then provide gradients to train a behavioral cloning policy to improve its generalizability.
Nevertheless, these works require action labels, fundamentally limiting their applicability to learning from observation.

In this work, we introduce Diffusion Imitation from Observation (DIFO), a novel adversarial imitation learning from observation method that
employs a diffusion model as a discriminator to provide rewards for policy learning.
Specifically, we design a diffusion model that learns to capture expert and agent state transitions by generating the subsequent state conditioning on the current state.
We reformulate the denoising objective of diffusion models as a binary classification task, allowing for the diffusion model to distinguish expert and agent transitions.
Then, provided with the ``realness'' rewards from the diffusion model, the policy imitates the expert by producing transitions that look indistinguishable from expert transitions.

We compare our method \method{} to various existing LfO methods in various continuous control domains, including navigation, locomotion, manipulation, and games.
The experimental results show that \method{} consistently exhibits superior performance.
Moreover, \method{} demonstrates better data efficiency.
The visualized learned reward function and generated state distributions verify the effectiveness of our proposed learning objective for the diffusion model.

\vspace{-0.5em}
\vspacesection{Related work}
\label{sec:related_work}

\myparagraph{Learning from demonstration (LfD)} LfD approaches imitate experts from collected demonstrations, consisting of state and action sequences.
Behavioral cloning (BC)~\citep{pomerleau1991efficient_bc, shafiullah2022behavior}
%, Giusti2016Visual, Samak2021Robust, shafiullah2022behavior} 
formulates LfD as a supervised learning problem by learning a state-to-action mapping.
Inverse reinforcement learning (IRL)~\citep{
russell1998learning, 
ng2000algorithms, 
abbeel2004apprenticeship}
% ho2016generative,
% ross2011reduction,
% Choi2013Bayesian,
% brown2018efficient, 
% kostrikov2018discriminatoractorcritic,
% yu2019meta,
% choi2021trajgail,
% fu2018learning,
% xiao2019wasserstein,
% lai2024diffusion,
% wang2024diffail
extracts a reward function from demonstrations and uses it to learn a policy through reinforcement learning.
In contrast, this work aims to learn from state-only demonstrations, requiring no action label.

\myparagraph{Learning from observation (LfO)} 
LfO~\citep{devin2017learning, sun2018neural} learns from state-only demonstrations, \ie state sequences, making it suitable for scenarios where action labels are unobservable or costly to obtain, and allowing for learning from experts with a different embodiment.
To tackle LfO, one popular direction is to learn an inverse dynamics model (IDM) for an agent that can recover an action for a pair of consecutive states~\citep{2024lapo, torabi2018behavioral, yang2019imitation}.
However, there is no apparent mechanism to efficiently collect tuples of state, next state, and action that align with the expert state sequences, which makes it difficult to learn a good IDM.
On the other hand, adversarial imitation learning from observation (AILfO)~\citep{Rafailov2021Visual, Henderson_2018, torabi2018generative, goalprox}  resemble the idea of generative adversarial networks (GANs)~\citep{goodfellow2014generative}, where an agent generator policy is rewarded by a discriminator learning to distinguish the expert state transitions from the agent state transitions.
Despite the encouraging results, the AILfO trainings are often brittle and sensitive to hyperparameters~\citep{fu2018learning, barde2020adversarial}.
Recent works also use generative models to predict state transitions and use the prediction to guide policy learning using log-likelihood~\citep{escontrela2024video}, ELBO~\citep{zhang2024action}, or conditional entropy~\citep{Huang2023DiffusionReward}.
However, these methods depend highly on the accuracy of the generative models.
In contrast, our work aims to improve the sample efficiency and robustness of AILfO by employing a diffusion model as a discriminator.

\myparagraph{Learning from video (LfV)}
Extending from LfO, LfV specifically considers learning from image-based states, \ie videos, 
by leveraging recent advancements in computer vision, \eg 
multi-view learning~\citep{sermanet2018time}, 
image and video comprehension and generation~\citep{schmeckpeper2020reinforcement, escontrela2024video, seo2022reinforcement, ma2022vip, ghosh2023reinforcement, bhateja2023robotic, kumar2023pretraining},
foundation models~\citep{ma2023liv, embodimentcollaboration2024open},
and optical flow and tracking~\citep{ko2024learning, wen2023any}.
Yet, these methods are mostly specifically designed for learning from videos, and cannot be trivially adapted for vectorized states.

\myparagraph{Diffusion models}
Diffusion models are state-of-the-art generative models capable of capturing and generating high-dimensional data distributions~\citep{ho2020ddpm, song2022denoising}.
Diffusion models have been widely adopted for generating images~\citep{saharia2022photorealistic, rombach2021highresolution}, videos~\citep{blattmann2023stable}, 3D structures~\citep{poole2022dreamfusion}, and speech~\citep{liu2022diffgantts, popov2021grad}.
Recent works also have explored using the ability to model multimodal distributions of diffusion models for LfD~\citep{chi2023diffusionpolicy, pearce2022imitating, chen2024diffusion, lai2024diffusion, wang2024diffail}, where expert demonstrations could exhibit significant variability~\citep{li2017infogail}. 
Our work aims to employ the capability of diffusion models for improving AIRLfO.

\vspace{-0.5em}
\vspacesection{Preliminary}
\label{sec:preliminary}

\vspacesubsection{Learning from observation}
\vspace{-0.5em}
Consider environments represented as a Markov decision process (MDP) defined as a tuple ($\mathcal{S}, \mathcal{A}, r, \mathcal{P}, \rho_0, \gamma)$ of state space $\mathcal{S}$, action space $\mathcal{A}$, reward function $r(s, a, s')$, transition dynamics $\mathcal{P}(s'|s, a)$, initial state distribution $\rho_0$ and discounting factor $\gamma$. 
We define a policy $\pi(a|s)$ that takes actions from state inputs and generates trajectories $\tau = (s_0, a_0, s_1,\ldots,s_{|\tau|})$.
The policy is trained to maximize the sum of discounted rewards $\mathbb{E}_{\left(s_0, a_0, \ldots, s_{|\tau|}\right) \sim \pi}\left[\sum_{i=0}^{{|\tau|}-1} \gamma^i r\left(s_i, a_i, s_{i+1}\right)\right]$. 

In imitation learning, the environment rewards cannot be observed. Instead, a set of expert demonstrations $\tau_E = \{\tau_0,\ldots,\tau_N|\tau_i \sim \pi_E\}$ is given, which generated by unknown expert policy $\pi_E$.
We aim to learn the agent policy $\pi_A$ to generate a similar trajectory distribution with expert demonstrations.
Moreover, in the learning from observation (LfO) setting, where expert action labels are absent, agents learn exclusively from state-only observations represented by sequences $\tau = (s_0, s_1,\ldots,s_{{|\tau|}})$. We use the LfO setting in this work.

\myparagraph{Inverse reinforcement learning (IRL)}
One of the general approaches to imitation learning is IRL.
This approach learns a reward function $r$ from transitions, \ie $(s,a)$ in LfD or $(s,s')$ in LfO, that maximizes the reward of expert transitions and minimizes that of agent transitions. 
The learned reward function can thereby be used for reinforcement learning to train the policy to imitate expert.

\vspacesubsection{Denoising Diffusion Probabilistic Models}
Diffusion models have emerged as state-of-the-art generative models capable of producing high-dimensional data and modeling multimodal distributions.
Our work leverages the Denoising Diffusion Probabilistic Model (DDPM)~\citep{ho2020ddpm}, a latent variable model that generates data through a denoising process. 
The training procedure of the diffusion model consists of forward and reverse processes.  In the forward process, Gaussian noise is progressively added to the clean data, following a predefined noise schedule. The process is formulated as
$\mathbf{x}_t=\sqrt{\bar{\alpha_t}}\mathbf{x}_0+\sqrt{1-\bar{\alpha_t}}\boldsymbol{\epsilon}$,
where $\mathbf{x}_0$ is the clean data, $\boldsymbol{\epsilon}$ is the Gaussian noise, $t$ denotes the time step within the whole process with step $T$ and $\bar{\alpha_t}$ is the scheduled noise level at the current time step.
Conversely, the reverse process, denoted by $p_{\phi}(\textbf{x}_{t-1}|\textbf{x}_t)$, is designed to reconstruct the original data by estimating the previously injected noise based on the given noise level. This is achieved by optimizing
$ \mathcal{L}=\mathbb{E}_{t\sim T,\boldsymbol{\epsilon} \sim \mathcal{N}(0,1)}\left[\left\lVert        \boldsymbol{\epsilon}-\boldsymbol{\epsilon}_\phi(\mathbf{x}_t,t)
\right\rVert^{2}
\right]$,
where $\phi$ denotes the diffusion model.

\vspacesection{Approach}
\label{sec:approach}

\begin{figure}
    \centering
    \includegraphics[trim={2.2cm 3.8cm 1cm 1.4cm},clip,width=\linewidth]{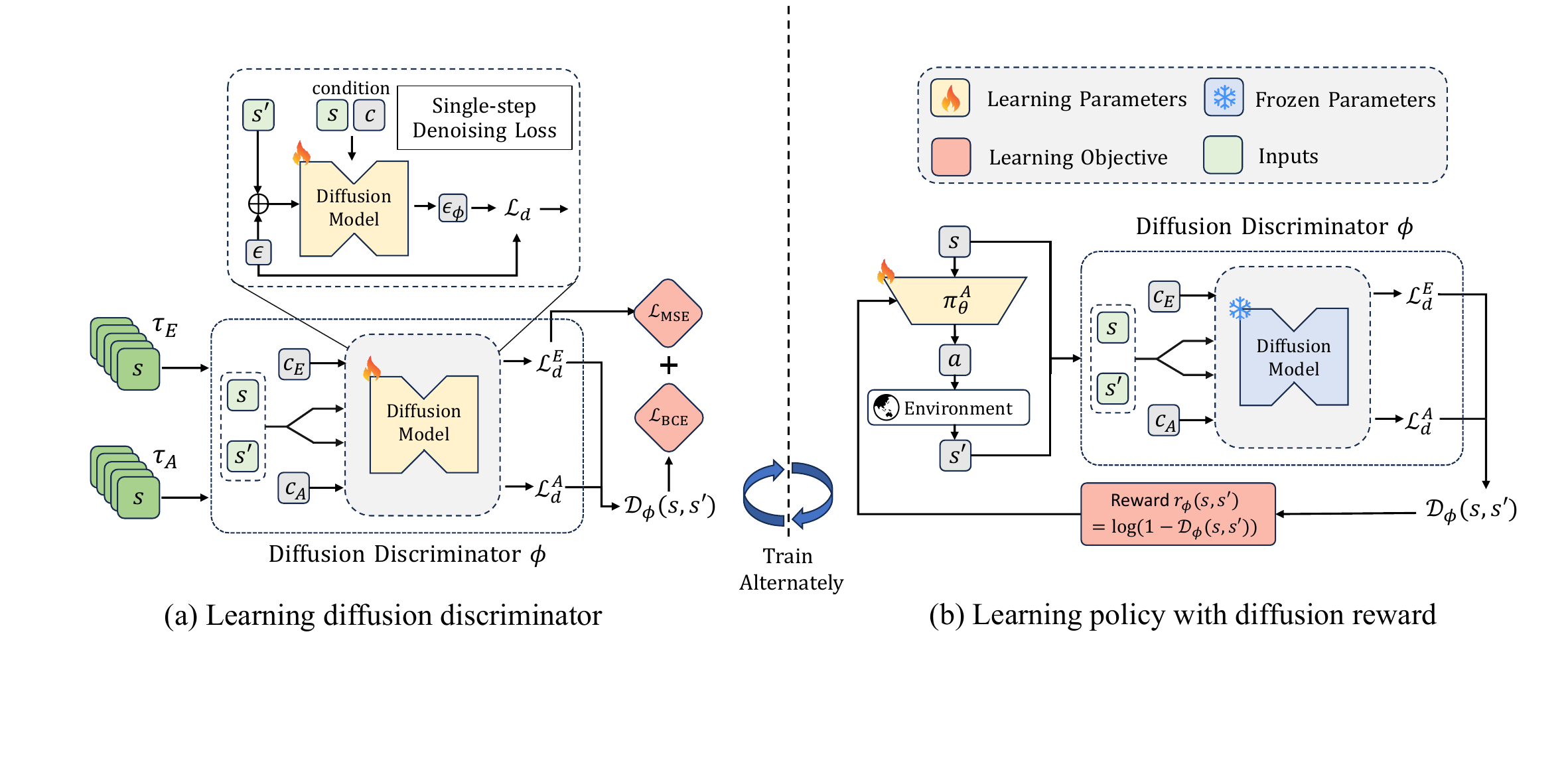}
    % \includegraphics[trim={0cm 1cm 0cm 1cm},clip,width=\linewidth]{figures/4_approach/framework-final.pdf}
    % \vspace{-0.5cm}
    \caption{\textbf{{\methodFull} (\method).}
    We propose {\methodFull} (\method), a novel adversarial imitation learning from observation framework employing a conditional diffusion model. 
    \textbf{(a) Learning diffusion discriminator.} In the \textit{discriminator step} the diffusion model learns to model a state transition $(\mathbf{s}, \mathbf{s}')$ by conditioning on the current state $\mathbf{s}$ and generates the next state $\mathbf{s}'$.
    With the additional condition on binary expert and agent labels ($c_E/c_A$), we construct the diffusion discriminator to distinguish expert and agent transitions by leveraging the single-step denoising loss as a likelihood approximation.
    \textbf{(b) Learning policy with diffusion reward.}
    In the \textit{policy step}, we optimize the policy with reinforcement learning according to
    rewards calculated based on the diffusion discriminator's output $\log(1 - \mathcal{D}_{\phi}(\mathbf{s},\mathbf{s'}))$.
    }
    \label{fig:DIFO}
\end{figure}

We propose {\methodFull} (\method), a novel learning from observation framework integrating a diffusion model into the AIL framework, which is illustrated in \myfig{fig:DIFO}. %\ref{fig:DIFO_a}
Specifically, we utilize a diffusion model to model expert and agent state transitions;
then, we learn an agent policy to imitate the expert via reinforcement learning by using the diffusion model to provide rewards based on how ``real'' agent state transitions are.

\vspacesubsection{Modeling expert transitions via diffusion model}
\label{sec:4.1}

Motivated by the recent success in using diffusion models for generative modeling, 
we use a conditional diffusion model to model expert state transitions.
Specifically, given a state transition $(\mathbf{s}, \mathbf{s}')$, the diffusion model conditions on the current state $\mathbf{s}$ and generates the next state $\mathbf{s}'$.
We adopt DDPM~\citep{ho2020ddpm} and define the reverse process as $p_{\phi}(\mathbf{s'}_{t-1}|\mathbf{s'}_{t}, \textbf{s})$, where $t \in T$ and $\phi$ is the diffusion model, which is trained by minimizing the denoising MSE loss:
\begin{align}
    \mathcal{L}_{d}(\mathbf{s},\mathbf{s'})&=
    \mathbb{E}_{t\sim T,\boldsymbol{\epsilon} \sim \mathcal{N}(0,1)}
    \left[
        \left\lVert
            \boldsymbol{\epsilon} - \boldsymbol{\epsilon}_\phi(\boldsymbol{s'}_t,t|\boldsymbol{s})
        \right\rVert^{2}
    \right],
    \label{eq:denoising_loss}
\end{align}
where $\epsilon$ denotes the noise sampled from a Gaussian distribution and $\epsilon_\phi$ denotes the noise predicted by the diffusion model.
Once the diffusion model is trained, we can generate an expert next state conditioned on any given state by going through the diffusion generation process.

\myparagraph{State-distance reward} 
To train a policy $\pi$ to imitate the expert from a given state $\mathbf{s}$, 
% we can first sample an action from the policy $a$, and feed it to the environment to obtain the policy next state $\mathbf{s_\pi}'$.
we can first sample an action from the policy and obtain the next state $\mathbf{s_\pi}'$ by interacting with the environment.
Next, we generate a predicted next state $\mathbf{s_\phi}'$ using the diffusion model.
Then, to bring the state distribution of the policy closer to the expert's, we can optimize the policy using reinforcement learning by setting the distance of the two next states $d(\mathbf{s_\pi}', \mathbf{s_\phi}')$ as a reward,
where $d$ denotes some distance function that evaluates how close two states are.
However, a good distance function varies from one domain to another.
Moreover, predicting the diffusion model next state $\mathbf{s_\phi}'$ can be very time-consuming since it requires $T$ denoising steps.

\myparagraph{Denoising reward} 
We aim to provide rewards for policy learning while avoiding choosing distance function and going through the diffusion generation process.
To this end, we take inspiration from \citet{li2023diffusion}, which shows that the denoising loss approximates the evidence lower bound (ELBO) of the likelihood.
Our key insight is to leverage the denoising loss calculated from a state and the policy next state $\mathcal{L}_{d}(\mathbf{s}, \mathbf{s_\pi}')$, or $\mathcal{L}_{d}$ in short, as an indicator of how well the policy next state fits the expert distribution.
That said, a low $\mathcal{L}_{d}$ means that the policy produces a  next state close to the expert next state, while a high $\mathcal{L}_{d}$ means that the diffusion model does not recognize this policy next state.
Hence, we can use $-\mathcal{L}_{d}$ as reward to learn a policy to imitate the expert by taking actions to produce next states that can be recognized by the diffusion model. Note that this denoising reward can be computed using a single denoising step. 

\vspacesubsection{Diffusion model as a discriminator}
\label{sec:4.2}

The previous section describes how we can use the denoising loss as a reward for policy learning via reinforcement learning.
However, the policy can learn to exploit a frozen diffusion model by discovering states that lead to a low denoising loss while being drastically different from expert states.
To mitigate this issue, we incorporate principles from the AIL framework by training the diffusion model to recognize both the transitions from the expert and agent.
To this end, we additionally condition the model on a binary label $c \in \{ c_E, c_A\}$, where $c_E$ represents the expert label and $c_A$ represents the agent label, both implemented as one-hot encoding, resulting in the following denoising losses given a state transition $(\mathbf{s}, \mathbf{s}')$:
\begin{align}
    \mathcal{L}_{d}^{E}(\mathbf{s},\mathbf{s'})&=
    \mathbb{E}_{t\sim T,\boldsymbol{\epsilon} \sim \mathcal{N}(0,1)}
    \left[
        \left\lVert
            \boldsymbol{\epsilon} - 
            \boldsymbol{\epsilon}_\phi(\boldsymbol{s'}_t,t|\boldsymbol{s},c_E)
        \right\rVert^{2}
    \right],
    \label{eq:expert_denoising_loss}\\
    {\mathcal{L}_{d}^{A}}(\mathbf{s},\mathbf{s'})&=
    \mathbb{E}_{t\sim T,\boldsymbol{\epsilon} \sim \mathcal{N}(0,1)}
    \left[
        \left\lVert
            \boldsymbol{\epsilon} - 
            \boldsymbol{\epsilon}_\phi(\boldsymbol{s'}_t,t|\boldsymbol{s},c_A)
        \right\rVert^{2}
    \right].
    \label{eq:agent_denoising_loss}
\end{align}
With this formulation and an optimized diffusion model, an expert transition should yield a low  \expertDenoisingLossspace and a high \agentDenoisingLoss, while an agent transition should yield a high \expertDenoisingLossspace and a low \agentDenoisingLoss.
Thus, we construct a diffusion discriminator that can determine if a transition is close to expert as follows:
\begin{align}
    \mathcal{D}_{\phi}(\mathbf{s},\mathbf{s'}) &= 
    \sigma(\lambda_{\sigma}(\mathcal{L}_{d}^{A}(\mathbf{s},\mathbf{s'}) - \mathcal{L}_{d}^{E}(\mathbf{s},\mathbf{s'}))),
    \label{eq:diffusion_discriminator}
\end{align}
where $\sigma$ is the sigmoid function for normalization and $\lambda_{\sigma}$ is a hyperparameter to control the sensitivity.
To turn this diffusion discriminator as a binary classifier to classify agent and expert transitions, we train it to optimize
the binary cross entropy (BCE):
\begin{align}
    \mathcal{L}_{\text{BCE}} &=
    \mathbb{E}_{(\mathbf{s},\mathbf{s'}) \sim \tau_E}
    \left[
        \log(
            1 - \mathcal{D}_{\phi}(\mathbf{s},\mathbf{s'})
        )
    \right] + 
    \mathbb{E}_{(\mathbf{s},\mathbf{s'}) \sim \pi_A}
    \left[
        \log(
            \mathcal{D}_{\phi}(\mathbf{s},\mathbf{s'})
        )
    \right].
    \label{eq:bce_objective}
\end{align}
By optimizing \(\mathcal{L}_{\text{BCE}}\), online interactions with the agent are leveraged as negative samples. Given expert transitions, the model should minimize \expertDenoisingLossspace and maximize \agentDenoisingLoss, resulting in a higher score closer to 1. Conversely, when the input is sampled from the agent, the model aims to maximize \expertDenoisingLossspace and minimize \agentDenoisingLoss, outputting a lower score closer to 0. 
The higher the score is, the more likely a transition is expert.
Hence, we can learn a policy to imitate the expert using $\mathcal{D}_{\phi}$ as rewards.
In contrast to MLP binary discriminators used in existing AIL works like GAIL, which maps high-dimensional inputs to a one-dimensional logit, our diffusion discriminator learns to predict high-dimensional noise patterns.
This is inherently more challenging to overfit, addressing one of the key instabilities in GAIL.

\vspacesubsection{\methodFull}
\label{sec:4.3}

We present \methodFull{} (\method{}) an adversarial imitation learning from observation framework that trains a policy and a discriminator in turns.
In the \textit{discriminator step},
the discriminator learns to classify expert and agent transition by optimizing $\mathcal{L}_{\text{BCE}}$. 
Furthermore, to ensure the diffusion loss of expert data is optimized so that it approximates the ELBO, 
the diffusion model also optimizes 
\expertDenoisingLossspace by sampling from expert demonstrations. 
\footnote{We experiment with optimizing $\mathcal{L}_{\text{MSE}}$ with agent data (\agentDenoisingLoss{}), leading to unstable training (see \cref{appendix:agent_mse}).}
\begin{align}
    \mathcal{L}_{\text{MSE}} &=
    \mathbb{E}_{t\sim T,\boldsymbol{\epsilon} \sim \mathcal{N}(0,1), (\mathbf{s},\mathbf{s'}) \sim \tau_E}
    \left[
        \left\lVert
            \boldsymbol{\epsilon} - 
            \boldsymbol{\epsilon}_\phi(\boldsymbol{s'}_t,t|\boldsymbol{s},c_E)
        \right\rVert^{2}
    \right],
    \label{eq:mse_objective}
\end{align}
resulting in the overall objective:
\begin{align}
    \mathcal{L}_{D} &= 
    \lambda_{\text{MSE}} \mathcal{L}_{\text{MSE}} + 
    \lambda_{\text{BCE}} \mathcal{L}_{\text{BCE}},
    \label{eq:overall_objective}
\end{align}
where $\lambda_{\text{MSE}}$ and $\lambda_{\text{BCE}}$ are hyperparameters adjusting the importance of each term.
In the \textit{policy step}, to provide the policy rewards based on the ``realness'' $\mathcal{D}_{\phi}$ of the agent transitions, we adopt the GAIL reward function~\citep{ho2016generative}:
\begin{align}
    r_{\phi}(\mathbf{s},\mathbf{s'}) &= \log(1 - \mathcal{D}_{\phi}(\mathbf{s},\mathbf{s'})),
    \label{eq:GAIL_reward}
\end{align}
where $\mathcal{D}_{\phi}$ is computed with a single denoising step.
We justify the feasibility of sampling only one denoising step in~\cref{sec:sample-timestep}.
We can optimize the policy using any RL algorithm.
The \method{} framework is illustrated in~\cref{fig:DIFO} and the algorithm is presented in~\cref{appendix:algo}.

\vspacesection{Experiments}
\label{sec:experiment}

\vspacesubsection{Environments}
\label{sec:5env}

\definecolor{FetchYellow}{RGB}{241,194,50}

\begin{figure*}[t]
    \centering
    \begin{minipage}{0.85\textwidth}
        \begin{subfigure}[b]{0.235\textwidth}
            \centering
            \includegraphics[width=\textwidth, height=\textwidth]{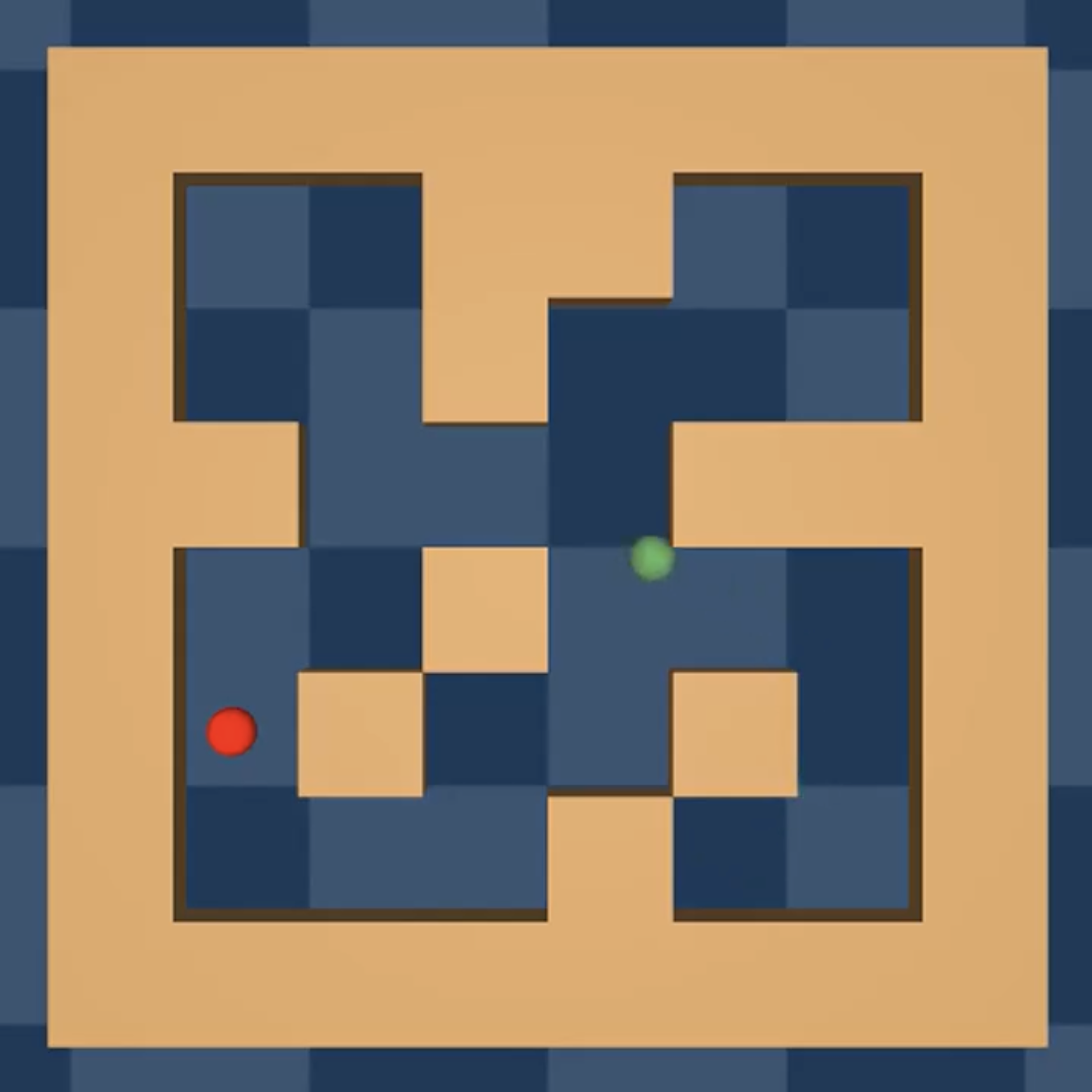}
            \caption{\pointmaze{}}
            \label{fig:5-1-point_maze}
        \end{subfigure}
        \hfill
        \begin{subfigure}[b]{0.235\textwidth}
            \centering
            \includegraphics[width=\textwidth, height=\textwidth]{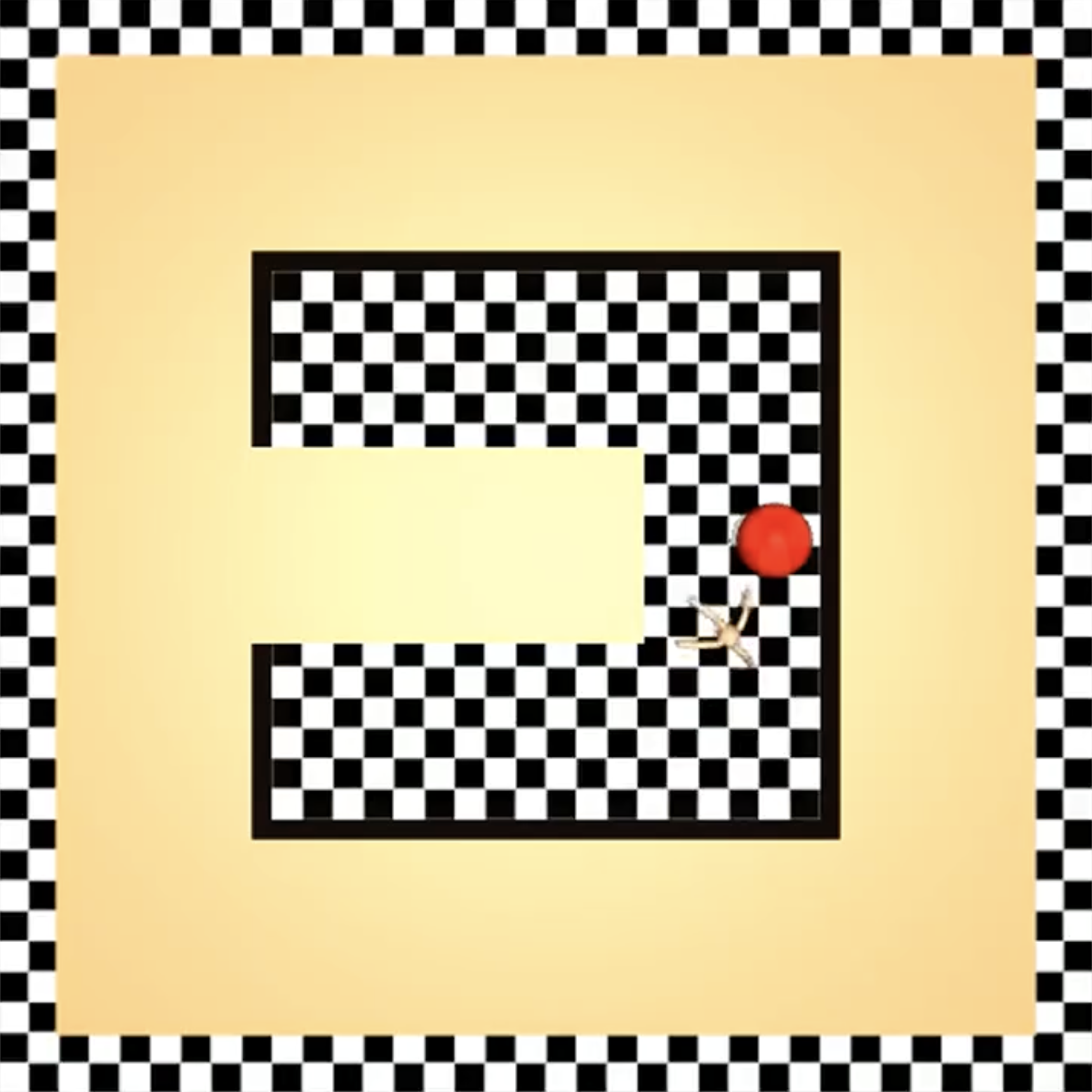}
            \caption{\antmaze{}}
            \label{fig:5-2-ant_maze} 
        \end{subfigure}
        \hfill
        \begin{subfigure}[b]{0.235\textwidth}
            \centering
            \includegraphics[width=\textwidth, height=\textwidth]{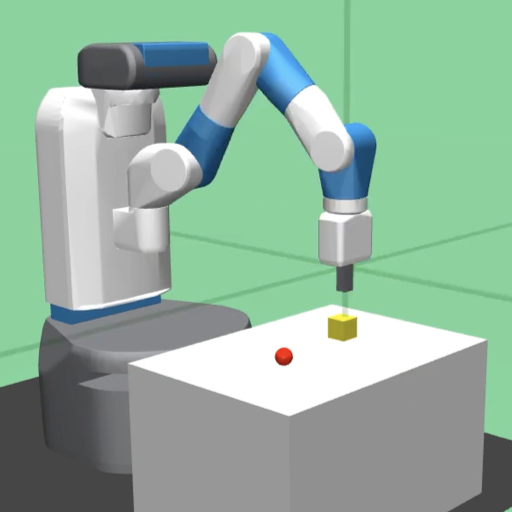}
            \caption{\fetchpush{}}
            \label{fig:5-3-fetch_push}
        \end{subfigure}    
        \hfill
        \begin{subfigure}[b]{0.235\textwidth}
            \centering
            \includegraphics[width=\textwidth, height=\textwidth]{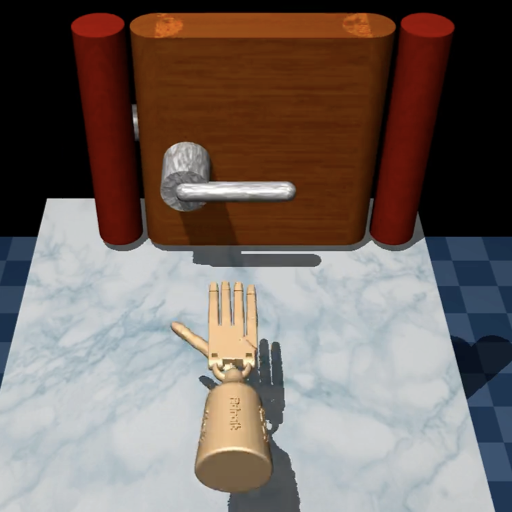}
            \caption{\scriptsize \adroitdoor{}}
            \label{fig:5-4-door}
        \end{subfigure}
    \end{minipage}
    \begin{minipage}{0.85\textwidth}
        \begin{subfigure}[b]{0.235\textwidth}
            \centering
            \includegraphics[width=\textwidth, height=\textwidth]{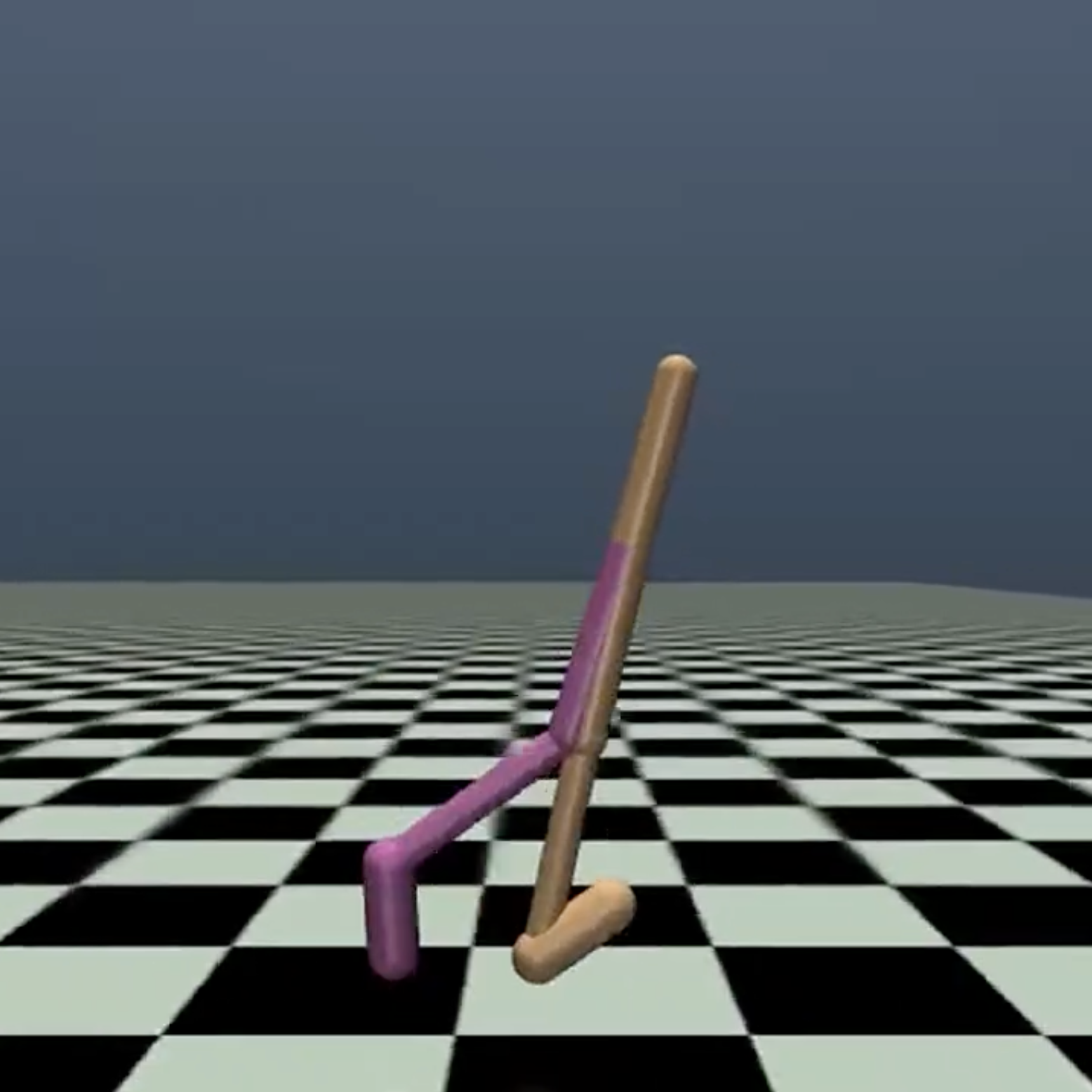}
            \caption{\walker{}}
            \label{fig:5-5-walker}  
        \end{subfigure}
        \hfill
        \begin{subfigure}[b]{0.235\textwidth}
            \centering
            \includegraphics[width=\textwidth, height=\textwidth]{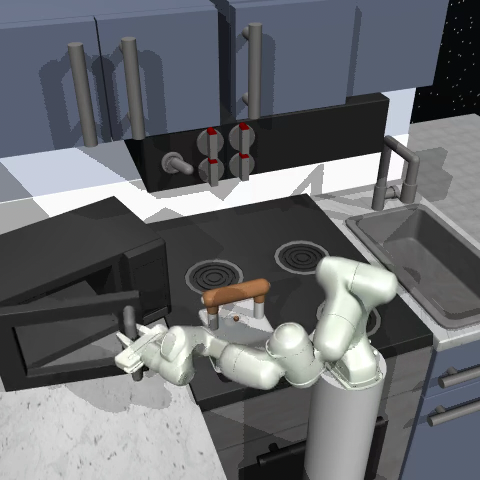}
            \caption{\frankakitchen{}}
            \label{fig:5-6-franka_kitchen}
        \end{subfigure}
        \hfill
        \begin{subfigure}[b]{0.235\textwidth}
            \centering
            \includegraphics[width=\textwidth, height=\textwidth]{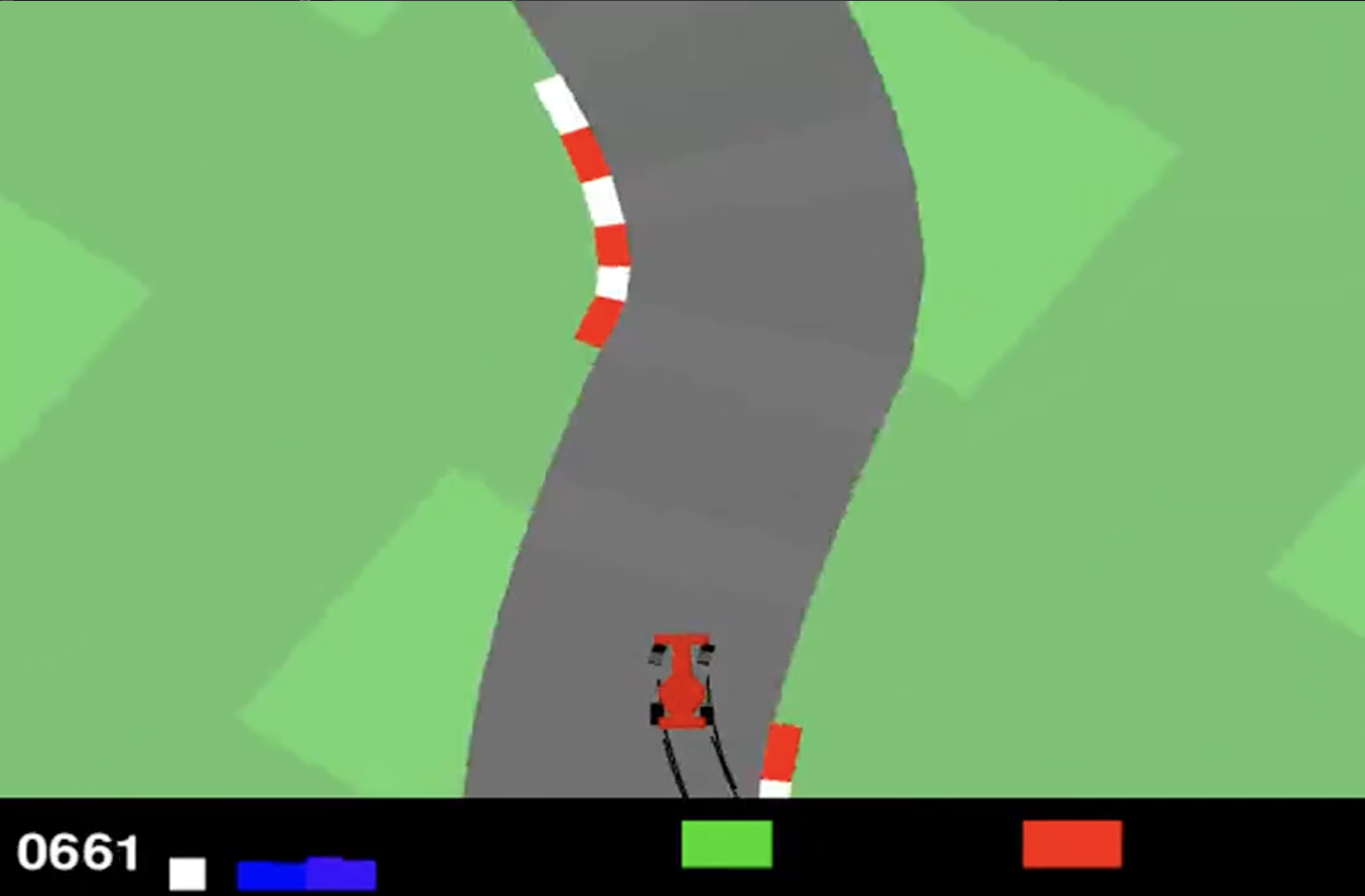}
            \caption{\carracing{}}
            \label{fig:5-7-car_racing}
        \end{subfigure}
        \hfill
        \begin{subfigure}[b]{0.235\textwidth}
            \centering
            \includegraphics[width=\textwidth, height=\textwidth]{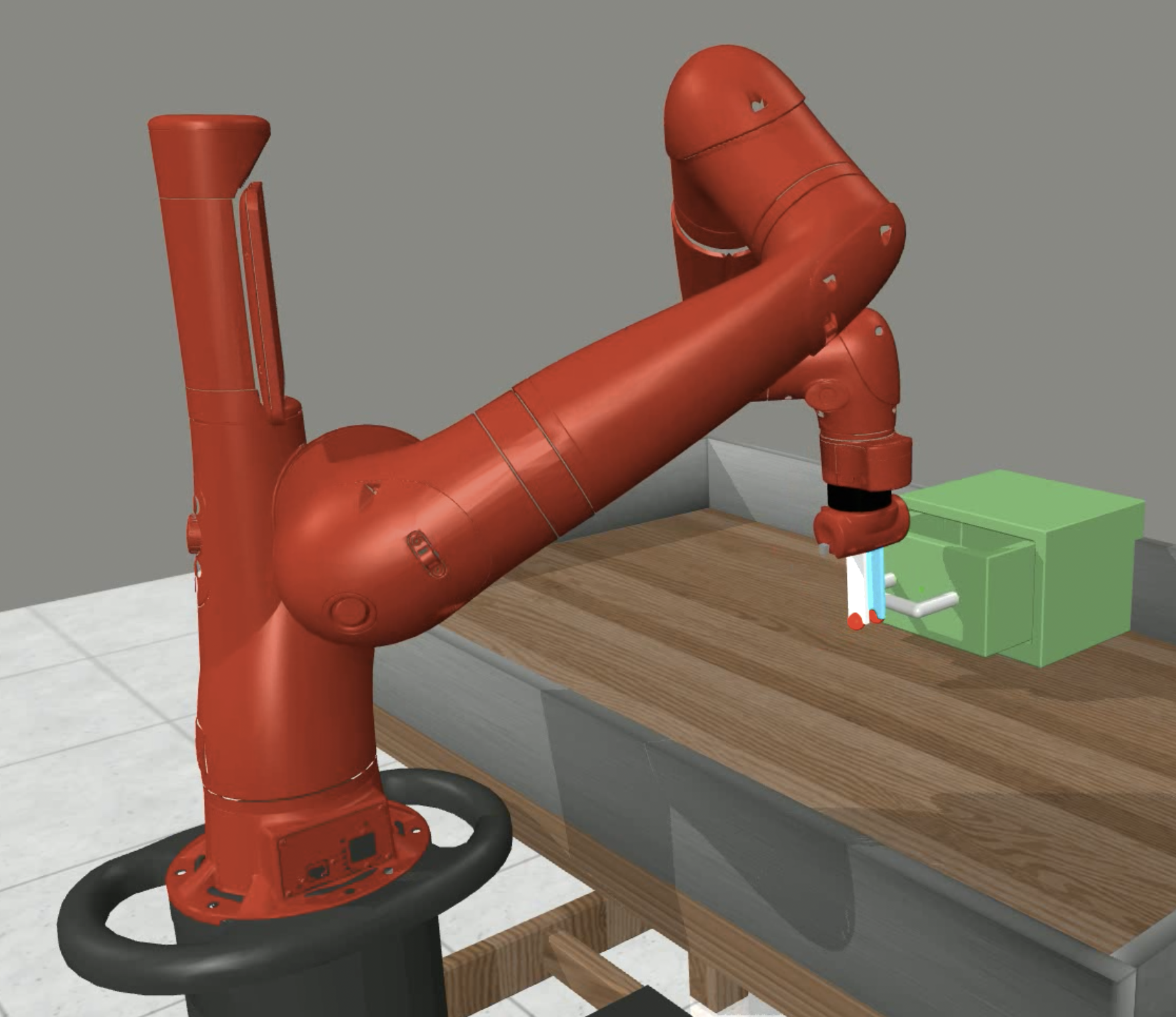}
            \caption{\drawerclose{}}
            \label{fig:5-8-drawer_close}
        \end{subfigure}
    \end{minipage}
    
    \caption[]{\textbf{Environments \& tasks.} 
    \textbf{(a) \pointmaze{}}: A point agent (\textcolor{ForestGreen}{green}) is trained to navigate to the goal (\textcolor{red}{red}).
    \textbf{(b) \antmaze{}}: A high-dimensional locomotion navigation task for an 8-DoF quadruped ant to navigate to the goal (\textcolor{red}{red}).
    \textbf{(c) \fetchpush{}}: A manipulation task to move a block (\textcolor{FetchYellow}{yellow}) to the target (\textcolor{red}{red}).
    \textbf{(d) \adroitdoor{}}: A high-dimension manipulation task to undo the latch and swing the door open.
    \textbf{(e) \walker{}}: A locomotion task for a 6-DoF hopper to maintain at the highest speed while keeping balance.
    \textbf{(f) \frankakitchen{}}: A manipulation task to control the robot arm to open the microwave with joint space control.
    \textbf{(g) \carracing{}}: An image-based task to control the car to complete the track in the shortest time.
    \textbf{(h) \drawerclose{}}: An image-based manipulation task to control the robot arm to close the drawer.
    }
    \label{fig:environment}
\end{figure*}

In this section, we introduce environments, tasks, and how expert demonstrations are collected. All environment trajectories, except {\carracing}, are fixed-horizon to prevent biased information about success~\citep{kostrikov2018discriminatoractorcritic}.
Further details can be found in \cref{appendix:environment}. %Appendix.\ref{}.\sun{TODO}
\begin{itemize}
    \item \textbf{\pointmaze}: A navigation task for a 2-DoF agent with the medium maze, see \cref{fig:5-1-point_maze}.
    A point agent is trained to navigate from an initial position to a goal.
    The goal and initial position of the agent are randomly sampled.
    The agent observes its position, velocity, and goal position.
    The agent applies linear forces in the $x$ and $y$ directions to navigate the maze and reach the goal.
    We collect $60$ demonstrations (\num{36000} transitions) using a controller from~\citet{fu2020d4rl}.
    
    \item \textbf{\antmaze}: A task containing both locomotion and navigation, which presents a significantly more challenging variant of the \textbf{\pointmaze}, as shown in \cref{fig:5-2-ant_maze}. 
    The quadruped ant learns to navigate from an initial position to a goal by controlling the torque of its legs, where both the goal and initial position of the ball are also randomly sampled.
    Notice that this environment serves as a high-dimensional state space task with 29-dimension state space.
    We use $100$ demonstrations (\num{7000} transitions) from Minari~\citep{minari2023minari}.
    
    \item \textbf{\fetchpush}: The goal is to control a 7-DoF Fetch robot arm to push a block to a target position on a table, see \cref{fig:5-3-fetch_push}. 
    Both the block and target positions are randomly sampled.
    The robot is controlled by small displacements of the gripper in XYZ coordinates, which has a 28-dimension state space and a 4-dimension action space.
    We generate 50 demonstrations (\num{2500} transitions) using an expert policy trained by SAC~\cite{tuomas2018sac}. 
    
    \item \textbf{\adroitdoor}: A manipulation task to undo the latch and swing the door open, see \cref{fig:5-4-door}.
    The position of the door is randomly placed.
    It is based on the Adroit manipulation platform~\citep{Kumar2016thesis}, with 39-dimension state space and 28-dimension action space containing all the joints.
    It serves as a high-dimensional state and action space task.
    We use 50 demonstrations (\num{10000} transitions) from the dataset released by~\citet{fu2020d4rl}.

    \item \textbf{\walker}: A locomotion task of a 6-DoF Walker2D in MuJoCo~\cite{todorov2012mujoco}, as shown in \cref{fig:5-5-walker}.
    The goal is to walk forward by applying torques on the six hinges.
    Initial joint states are added with uniform noise.
    We generate \num{1000} transitions using an expert policy trained by SAC~\cite{tuomas2018sac}.

    \item \textbf{\frankakitchen}: A manipulation task to control a 9-DoF Franka robot
    arm to open the microwave door, as shown in \cref{fig:5-6-franka_kitchen}. 
    The environment has a 59-dimension state space and a 9-dimension continuous action space to control the angular velocity of each joint.
    It serves as a high-dimensional state and action space task.
    We use 5 demonstrations (\num{300} transitions) from the dataset released by~\citet{fu2020d4rl}.

    \item \textbf{\carracing}: An image-based control task aimed at directing a car to complete a track as quickly as possible. Observations consist of top-down frames, as shown in \cref{fig:5-7-car_racing}. 
    Tracks are generated randomly in every episode. 
    The car has continuous action space to control the throttle, steering, and breaking.
    We generate 340 transitions using an expert policy trained by PPO~\cite{schulman2017proximal}.

    \item \textbf{\drawerclose}: An image-based manipulation task from Meta-World~\cite{yu2019metaworld} requires the agent to control a Sawyer robot arm to close a drawer. 
    Observations consist of fixed perspective frames, as shown in \cref{fig:5-8-drawer_close}. 
    The robot has continuous action space to control the gripper in XYZ coordinates, and the initial poses of the robot and the drawer are randomized in every episode.
    We generate 100 transitions using a scripted policy.
\end{itemize}

\begin{figure}[htbp]
     \centering
    \captionsetup[subfigure]{aboveskip=0.05cm,belowskip=0.05cm}
    \includegraphics[trim={0cm 2.5cm 0cm 2.6cm},clip,width=\textwidth]{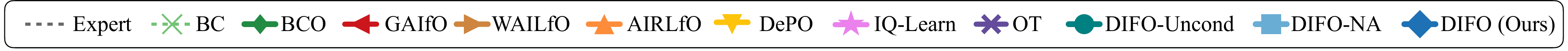}
     \begin{subfigure}[b]{0.245\textwidth}
         \centering
         \includegraphics[width=\textwidth]{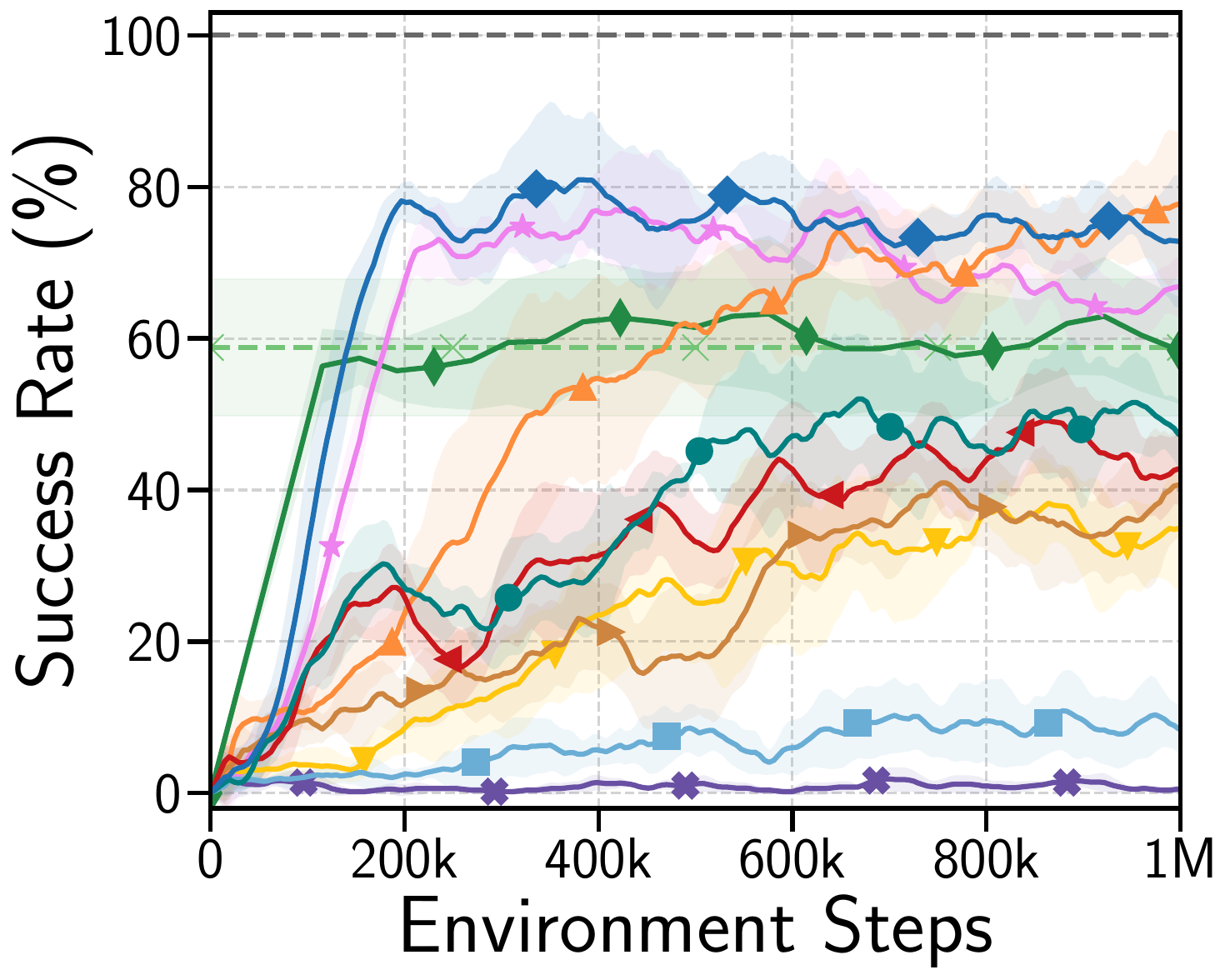}
         \caption{\pointmaze}
         \label{fig:main_result_point_maze}
     \end{subfigure}
     \begin{subfigure}[b]{0.245\textwidth}
         \centering
         \includegraphics[width=\textwidth]{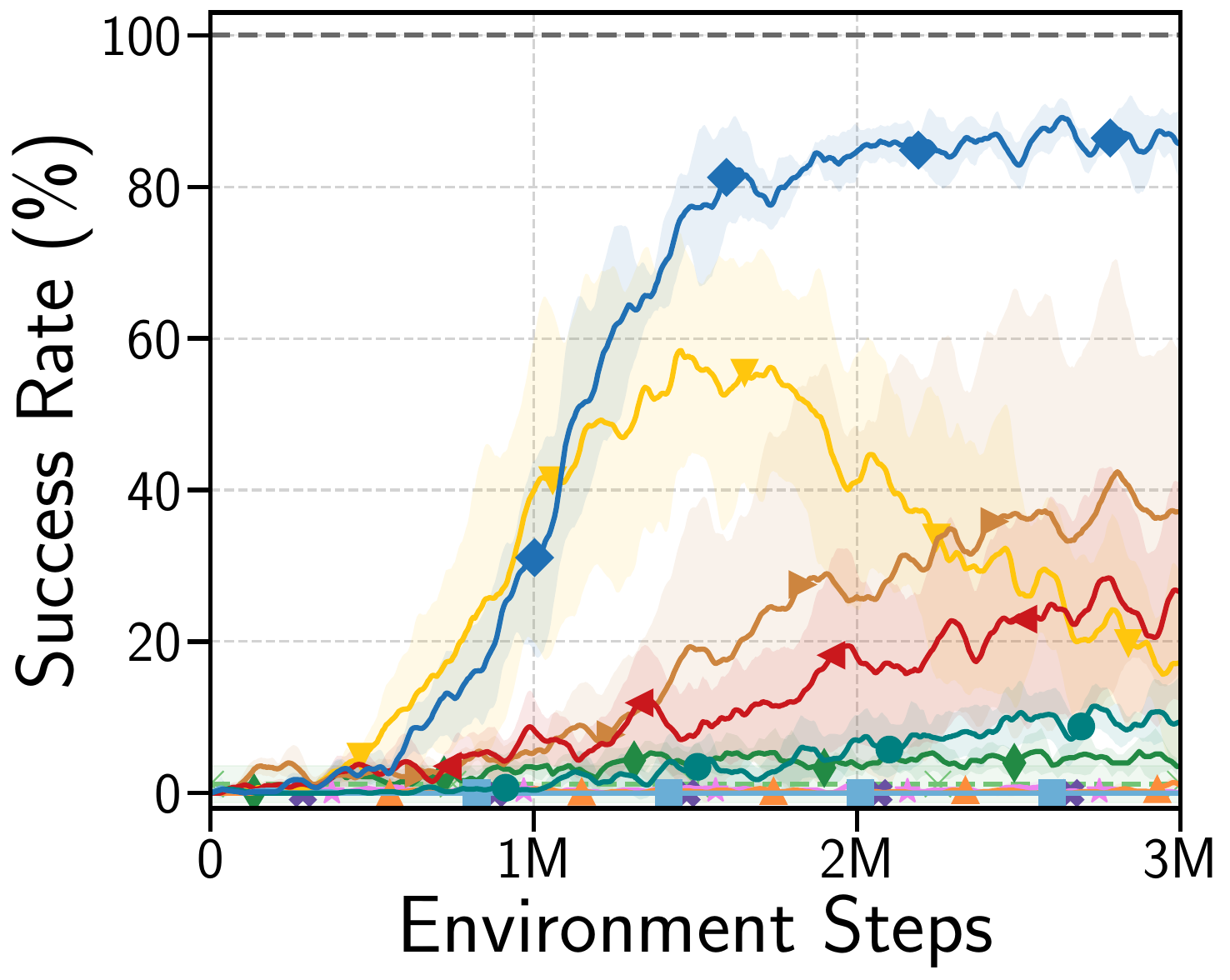}
         \caption{\antmaze}
         \label{fig:main_result_ant_maze}
     \end{subfigure}
     \begin{subfigure}[b]{0.245\textwidth}
         \centering
         \includegraphics[width=\textwidth]{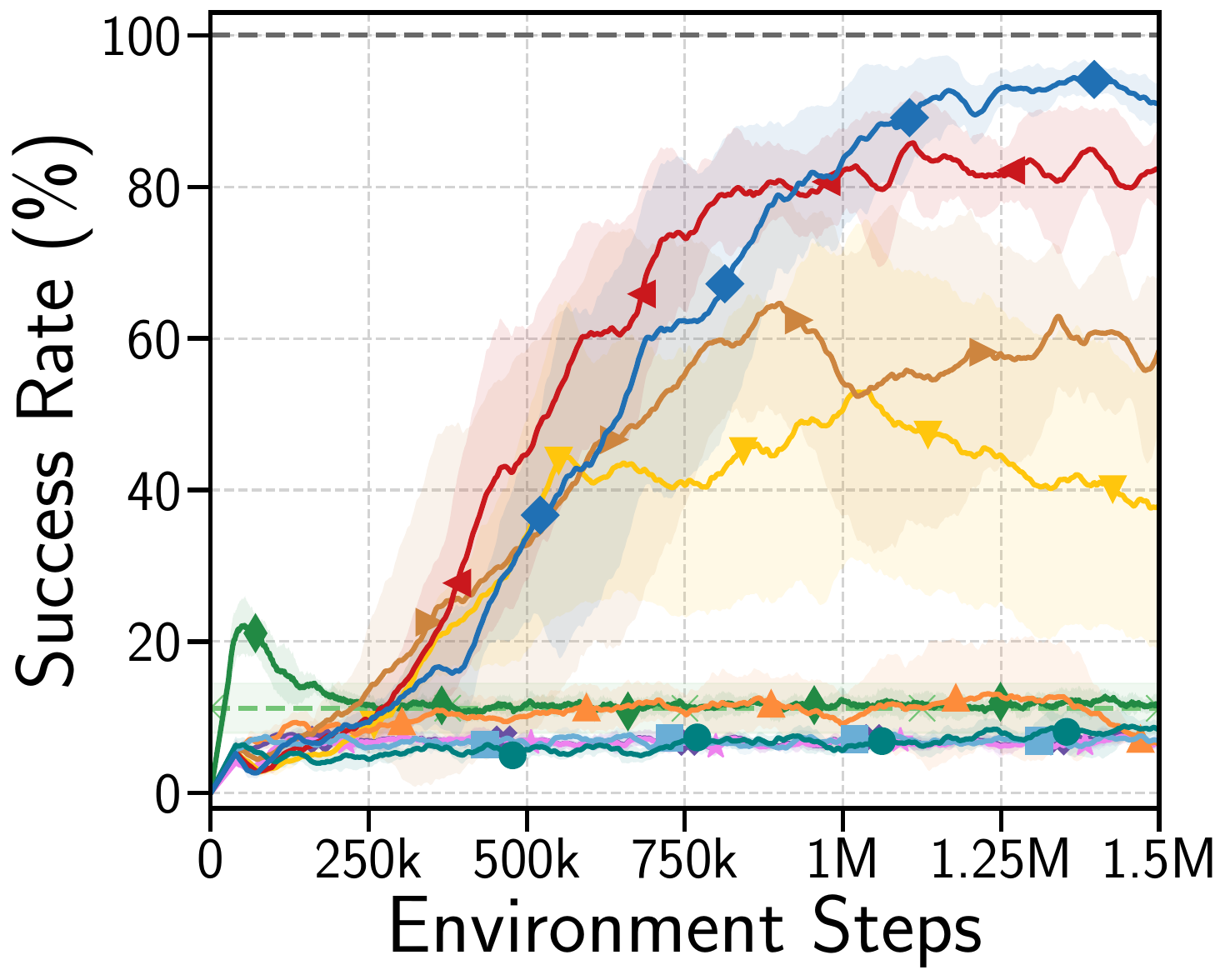}
         \caption{\fetchpush}
         \label{fig:main_result_fetch_push}
     \end{subfigure}
     \begin{subfigure}[b]{0.245\textwidth}
         \centering
         \includegraphics[width=\textwidth]{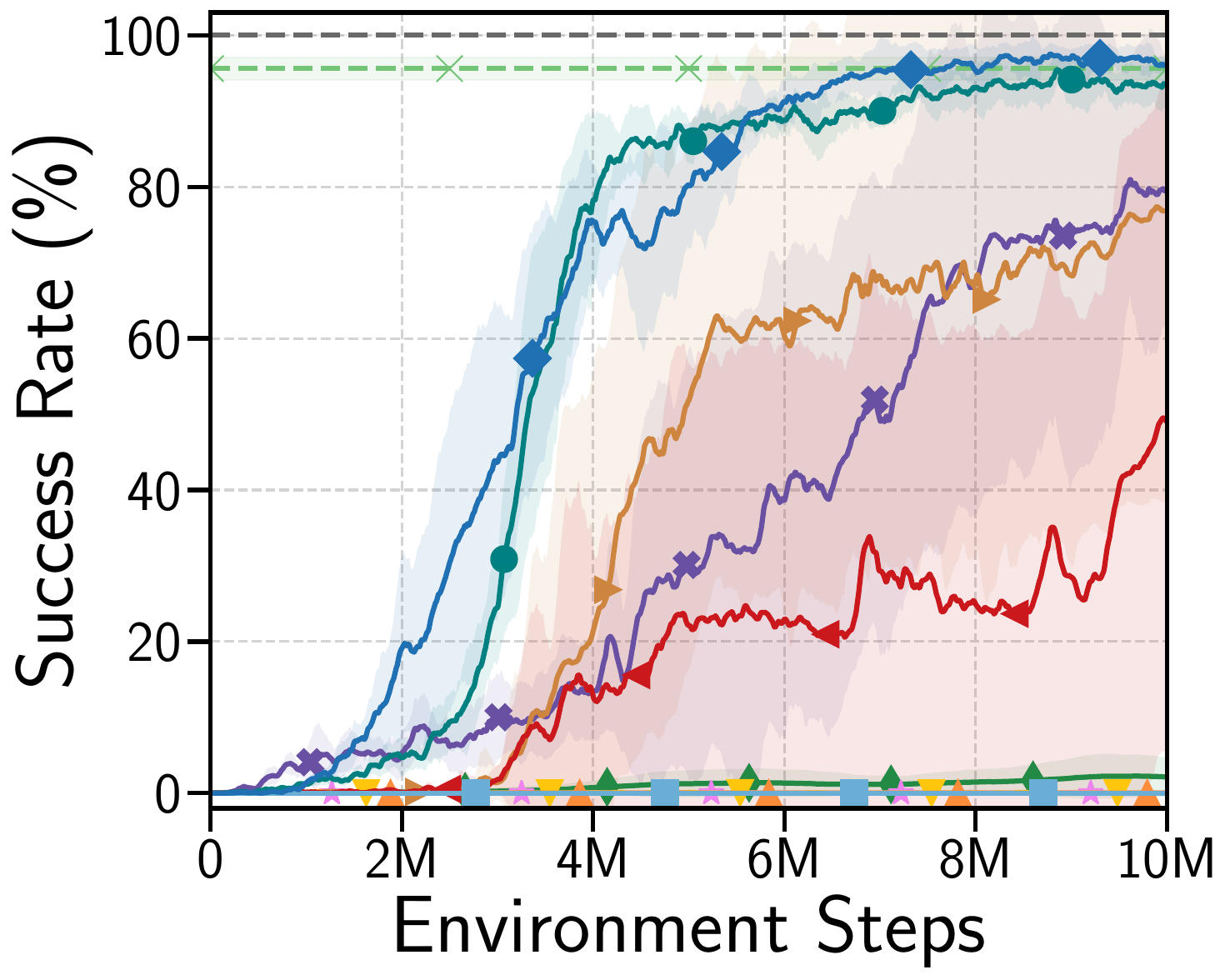}
         \caption{\adroitdoor}
         \label{fig:main_result_door}
     \end{subfigure}
     
     \begin{subfigure}[b]{0.245\textwidth}
         \centering
         \includegraphics[width=\textwidth]{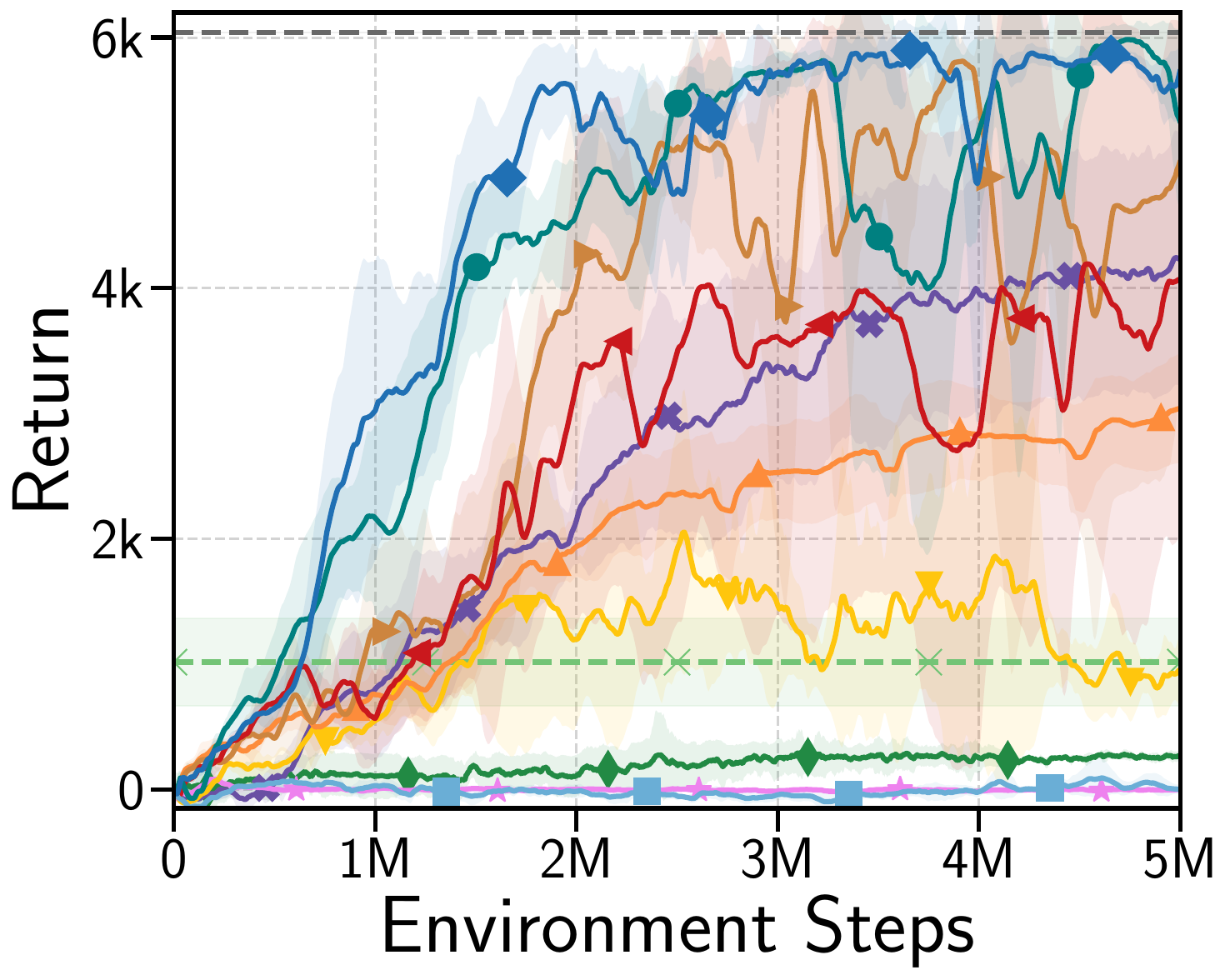}
         \caption{\walker}
         \label{fig:main_result_walker}
     \end{subfigure}
     \begin{subfigure}[b]{0.245\textwidth}
         \centering
         \includegraphics[width=\textwidth]{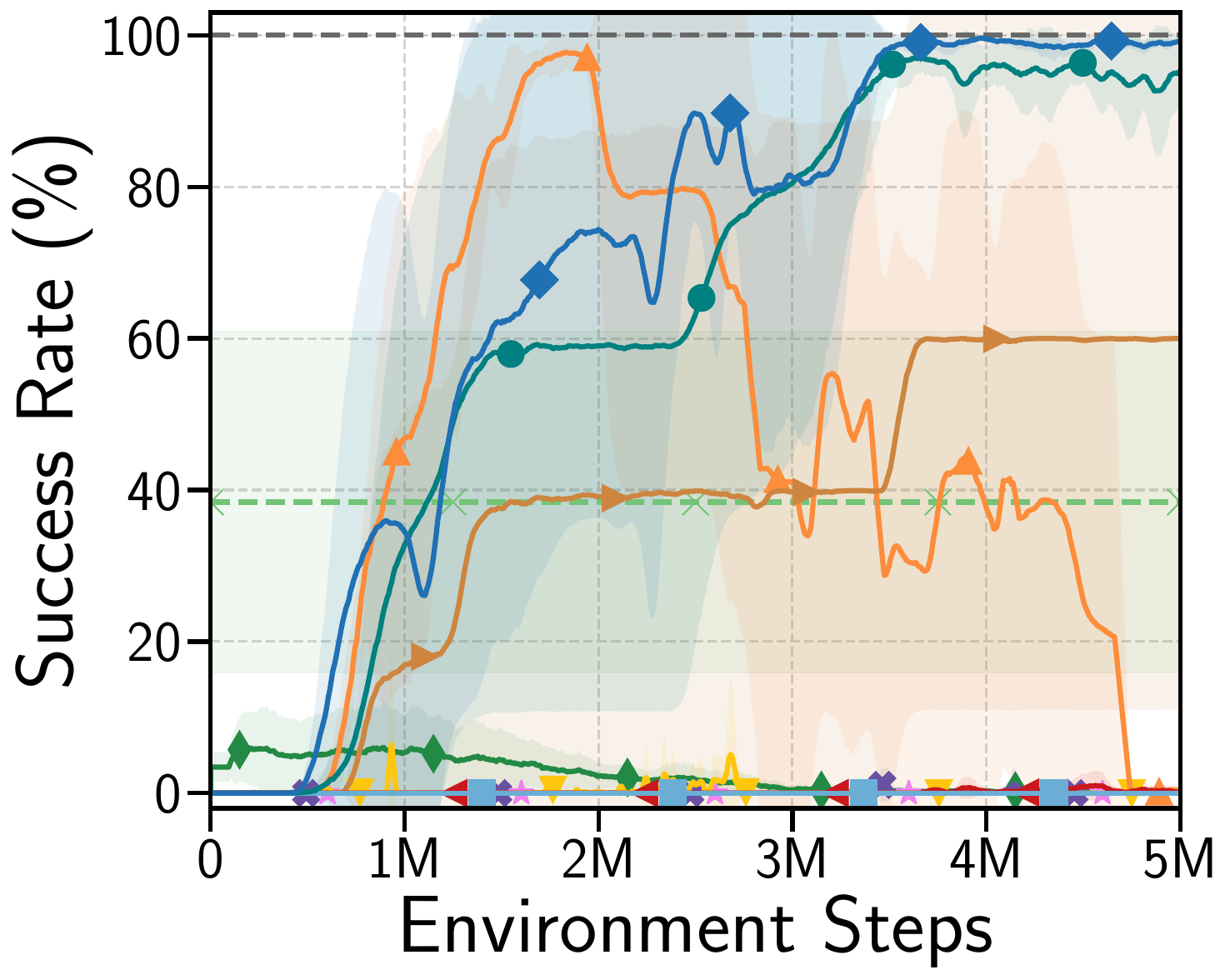}
         \caption{\frankakitchen{}}
         \label{fig:main_result_franka_kitchen}
     \end{subfigure}
     \begin{subfigure}[b]{0.245\textwidth}
         \centering
         \includegraphics[width=\textwidth]{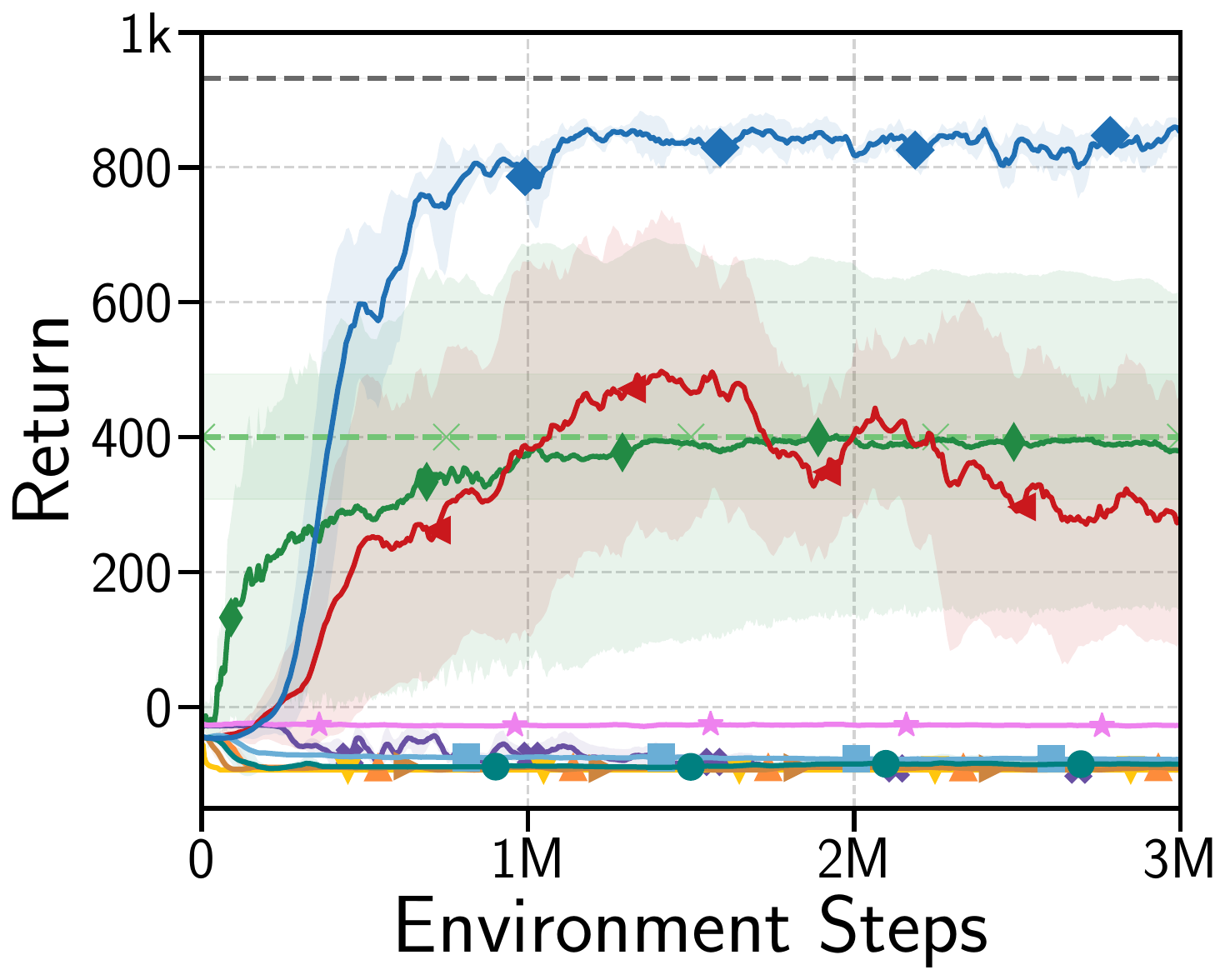}
         \caption{\carracing}
         \label{fig:main_result_car_racing}
     \end{subfigure}
     \begin{subfigure}[b]{0.245\textwidth}
         \centering
         \includegraphics[width=\textwidth]{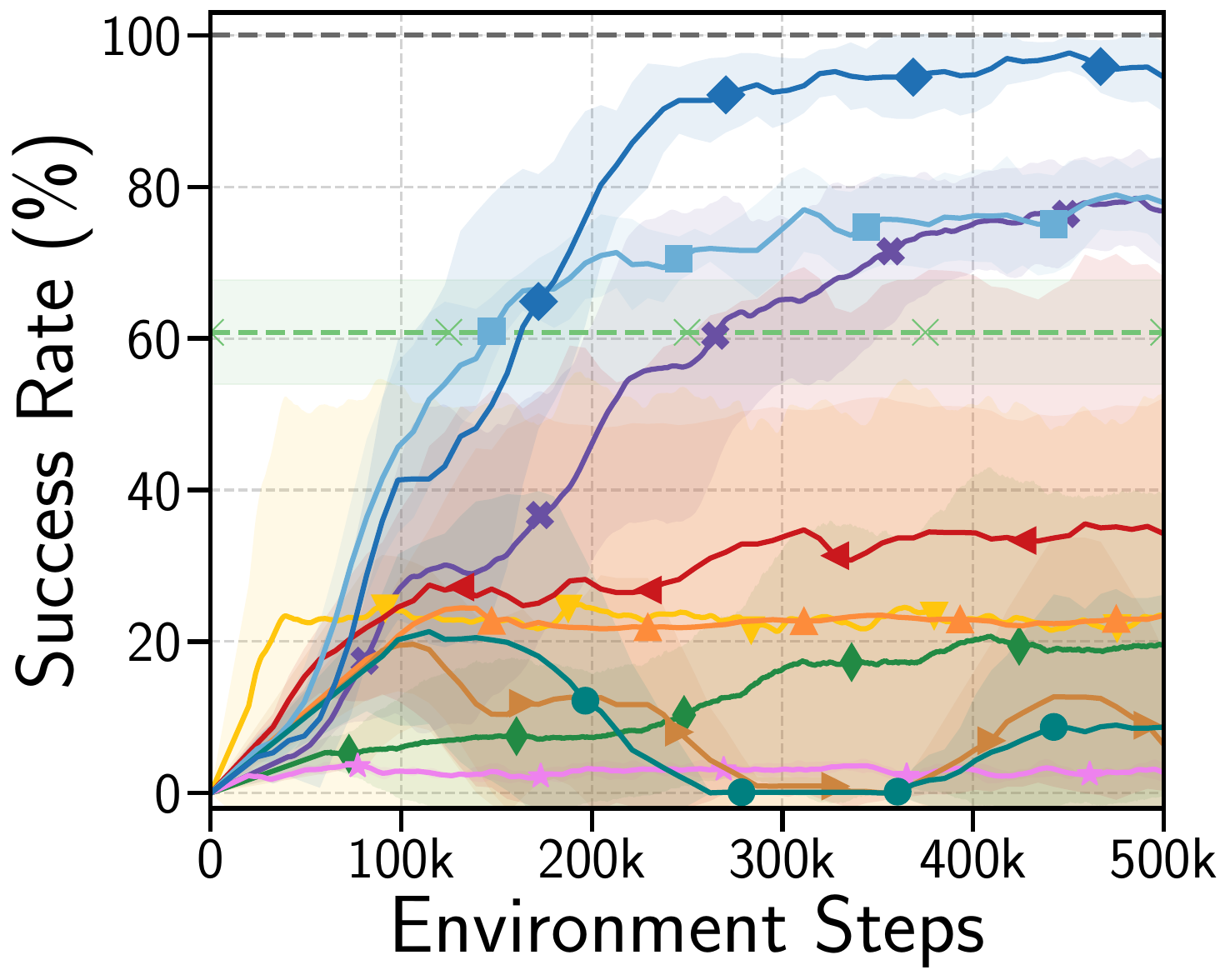}
         \caption{\drawerclose{}}
         \label{fig:main_result_drawer_close}
     \end{subfigure}
    \caption{\myparagraph{Learning performance and efficiency}
    We evaluate all the methods with five random seeds and report their success rates in \pointmaze{}, \antmaze{}, \fetchpush{}, \adroitdoor{}, \frankakitchen{}, and \drawerclose{}, 
    and their returns in \walker{}, and \carracing{}.
    The standard deviation is shown as the shaded area.
    Our proposed method, \method{}, demonstrates more stable and faster learning performance compared to the baselines.
    }
    \label{fig:main_result}
\end{figure}

\vspacesubsection{Baselines and variants}
\label{sec:baselines}
We compare our method {\method} with the following baselines:
\begin{itemize}
  \item \textbf{Behavioral Cloning (BC)}~\citep{pomerleau1991efficient_bc} learns a state-to-action mapping using supervised learning without any interaction with the environment. 
  Note that BC is the only baseline having privileged access to ground truth action labels.
  \item \textbf{Behavioral Cloning from Observation (BCO)}~\citep{torabi2018behavioral} first learns an inverse dynamic model through self-supervised exploration, and uses it to reconstruct action from state-only observation. BCO then uses these action labels to perform behavioral cloning.
  \item \textbf{Generative Adversarial Imitation from Observation (GAIfO)}~\citep{torabi2018generative}, 
  trains a GAIL MLP discriminator taking state transitions $(s,s')$ as input, instead of state-action pairs $(s,a)$.
  \item \textbf{Wasserstein Adversarial Imitation from Observation (WAIfO)} is a LfO variant of WAIL~\citep{xiao2019wasserstein}, taking $(s,s')$ as input. WAIL replaces the learning objective of the discriminator from Jensen-Shannon divergence (GAIL) to Wasserstein distance. 
  \item \textbf{Adverserial Inverse Reinforcement Learning from Observation (AIRLfO)} is a LfO variant of AIRL~\citep{fu2018learning}. AIRL modifies the discriminator output to disentangle task-relevant information from transition dynamics. Similarly to GAIfO, AIRLfO takes $(s,s')$ as input instead of $(s,a)$.
  \item \textbf{Decoupled Policy Optimization (DePO)}~\cite{liu2022plan} decouples the policy into a high-level state planner and an inverse dynamics model, utilizing embedded decoupled policy gradient and generative adversarial training.
  \item \textbf{Inverse soft-Q Learning for Imitation (IQ-Learn)}~\cite{garg2021iq} directly learns a policy in Q-space from demonstrations without explicit reward construction. We use the state-only setting for LfO.
  \item \textbf{Optimal Transport (OT)}~\cite{papagiannis2022imitation} derives a proxy reward function for RL by measuring the distance between probability distributions. We use the state-only setting for LfO.
\end{itemize}

In addition to the existing methods, we also compare DIFO with its variants:
\begin{itemize}
  \item \textbf{DIFO-Non-Adversarial (DIFO-NA)} follows the method introduced in \cref{sec:4.1}, which first pretrains a conditional diffusion model on expert demonstrations, and simply takes the denoising reward $-\mathcal{L}_{d}(s,s')$  for policy training.
  \item \textbf{DIFO-Unconditioned (DIFO-Uncond)} removes the condition on $s$, and denoises both $s$ and $s'$. It is optimized only with $\mathcal{L}_{\text{BCE}}$. Namely, replacing the MLP discriminator with a diffusion discriminator from GAIfO. It serves as a baseline showing the effect of network architecture.
\end{itemize}

\vspacesubsection{Experimental results}
\label{sec:main_experimental}
We report the success rates in \pointmaze{}, \antmaze{}, \fetchpush{}, and \adroitdoor{}, and return in \walker{} and \carracing{} of all the methods in~\cref{fig:main_result}.
Each method is reported with the mean value and standard deviation with five random seeds for all the tasks.
BC's performance is shown as horizontal lines since BC does not leverage environmental interactions.
The expert's performance (gray horizontal lines) in goal-directed tasks, \ie \pointmaze, \antmaze, \fetchpush, \adroitdoor, is $100\%$.
More details of training and evaluation can be found in \cref{appendix:training}.

Our proposed method DIFO consistently outperforms or matches the performance of the best-performing baseline in all the tasks, highlighting the effectiveness of integrating a conditional diffusion model into the AIL framework. 
In \antmaze, \adroitdoor, and \carracing, DIFO outperforms the baselines and converges significantly faster, indicating its efficiency in modeling expert behavior and providing effective rewards even in high-dimensional state and action spaces. 
Moreover, DIFO presents more stable training results, with relatively low variance compared to other AIL methods. 
Notably, although BC has access to action labels, it still fails in most tasks with more randomness.
This is because BC relies solely on learning from the observed expert dataset, unlike the LfO methods that utilize online interaction with environments, BC is susceptible to covariate shifts~\citep{ross2010efficient,ross2011reduction,laskey2016SHIV} and requires a substantial amount of expert data to achieve coverage of the dataset.
The result indicates the significance of online interactions.
OT only successfully learns in environments like \adroitdoor{}, \walker{}, and \drawerclose{}, where trajectory variety is limited.
OT computes distances at the trajectory level rather than the transition level, which requires monotonic trajectories, making it struggle in tasks with diverse trajectories. 
In contrast, our method generates rewards at the transition level, allowing us to identify transition similarities even when facing substantial trajectory variability. 

Variants of DIFO, \ie DIFO-Uncond, and DIFO-NA, perform poorly in most tasks.
DIFO-NA learns poorly in most of the tasks except \drawerclose, underscoring diffusion loss could be a reasonable metric for the discriminator while it is still necessary to model agent online interaction data to prevent the diffusion model from being exploited by the policy. 
On the other hand, DIFO-Uncond performs comparably to other AIL baselines but shows instability across different tasks, this highlighting the importance of modeling transitions using a diffusion model.

We also verify DIFO's capability to model stochastic distribution in~\cref{appendix:stochastic_environment}.

\vspacesubsection{Data efficiency}
\label{sec:data_efficiency}
\begin{figure}
     \centering
     \captionsetup[subfigure]{aboveskip=0.05cm,belowskip=0.05cm}
     
     \includegraphics[trim={0cm 2.5cm 0cm 2.6cm},clip,width=\textwidth]{figures/5_experiment/legend-main-times-new-roman.pdf}
     \begin{subfigure}[b]{0.245\textwidth}
         \centering
         \includegraphics[width=\textwidth]{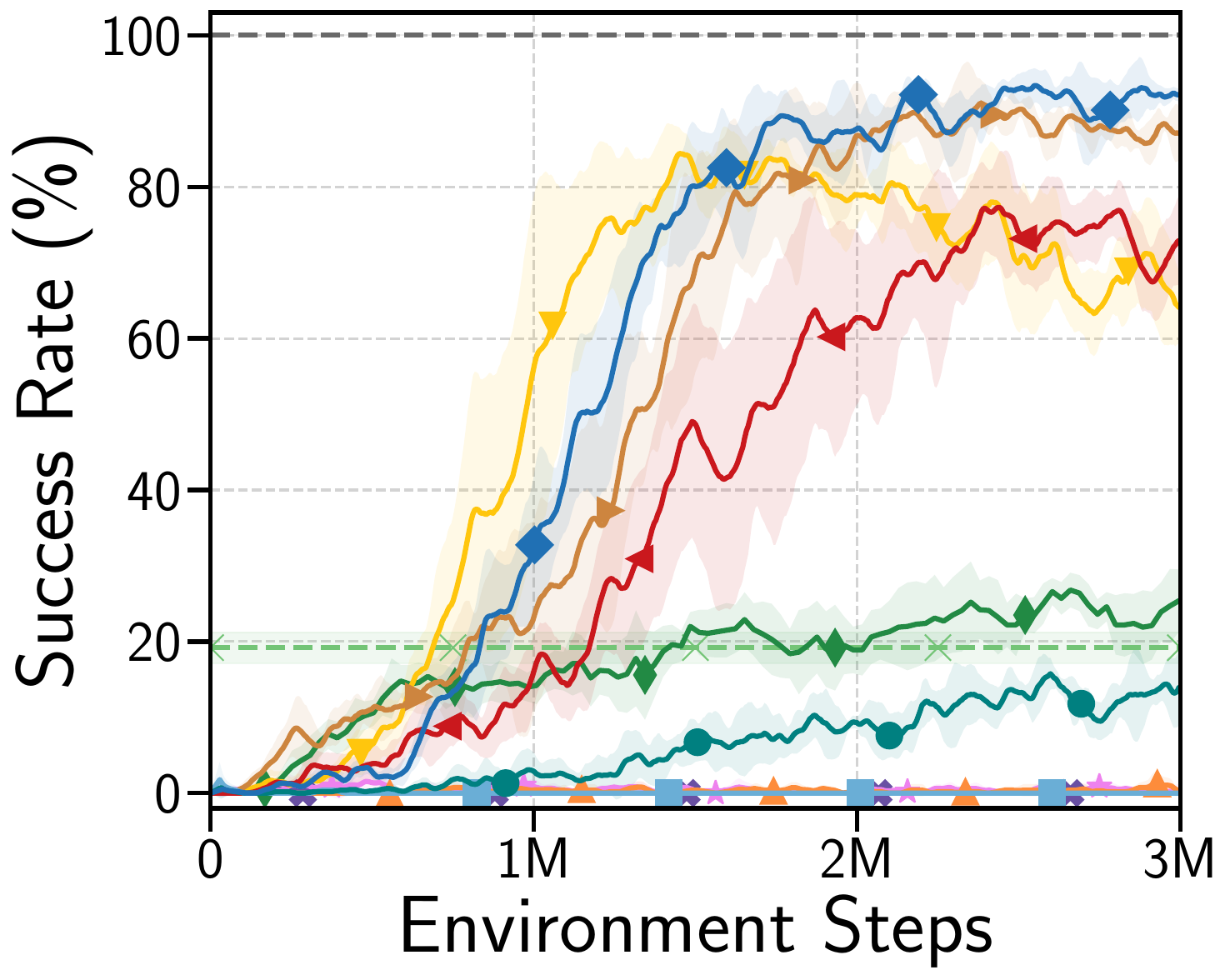}
         \caption{200 Trajectories}
         \label{fig:ant_maze_200}
     \end{subfigure}
     \begin{subfigure}[b]{0.245\textwidth}
         \centering
         \includegraphics[width=\textwidth]{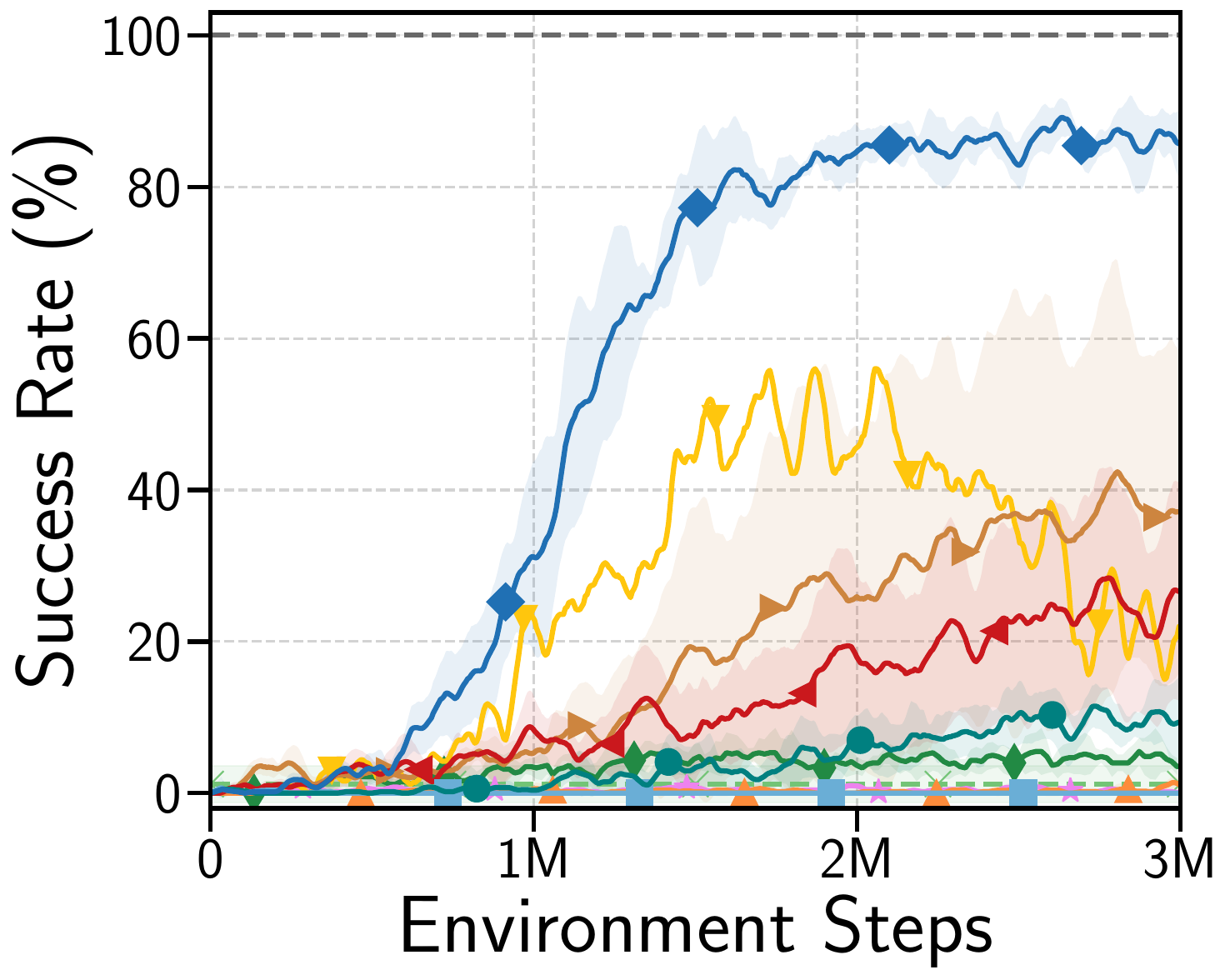}
         \caption{100 Trajectories}
         \label{fig:ant_maze_100}
     \end{subfigure}
     \begin{subfigure}[b]{0.245\textwidth}
         \centering
         \includegraphics[width=\textwidth]{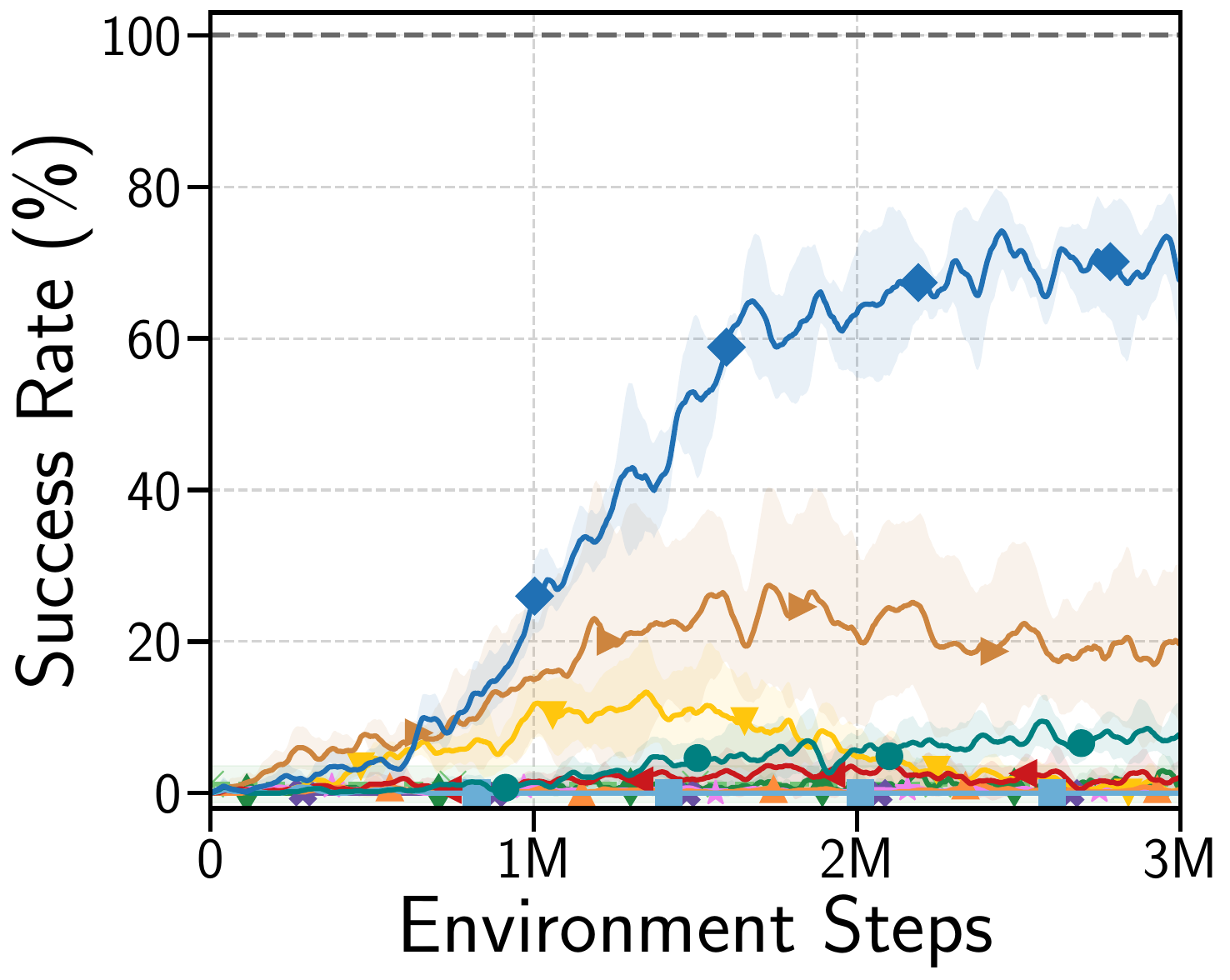}
         \caption{50 Trajectories}
         \label{fig:ant_maze_50}
     \end{subfigure}
     \begin{subfigure}[b]{0.245\textwidth}
         \centering
         \includegraphics[width=\textwidth]{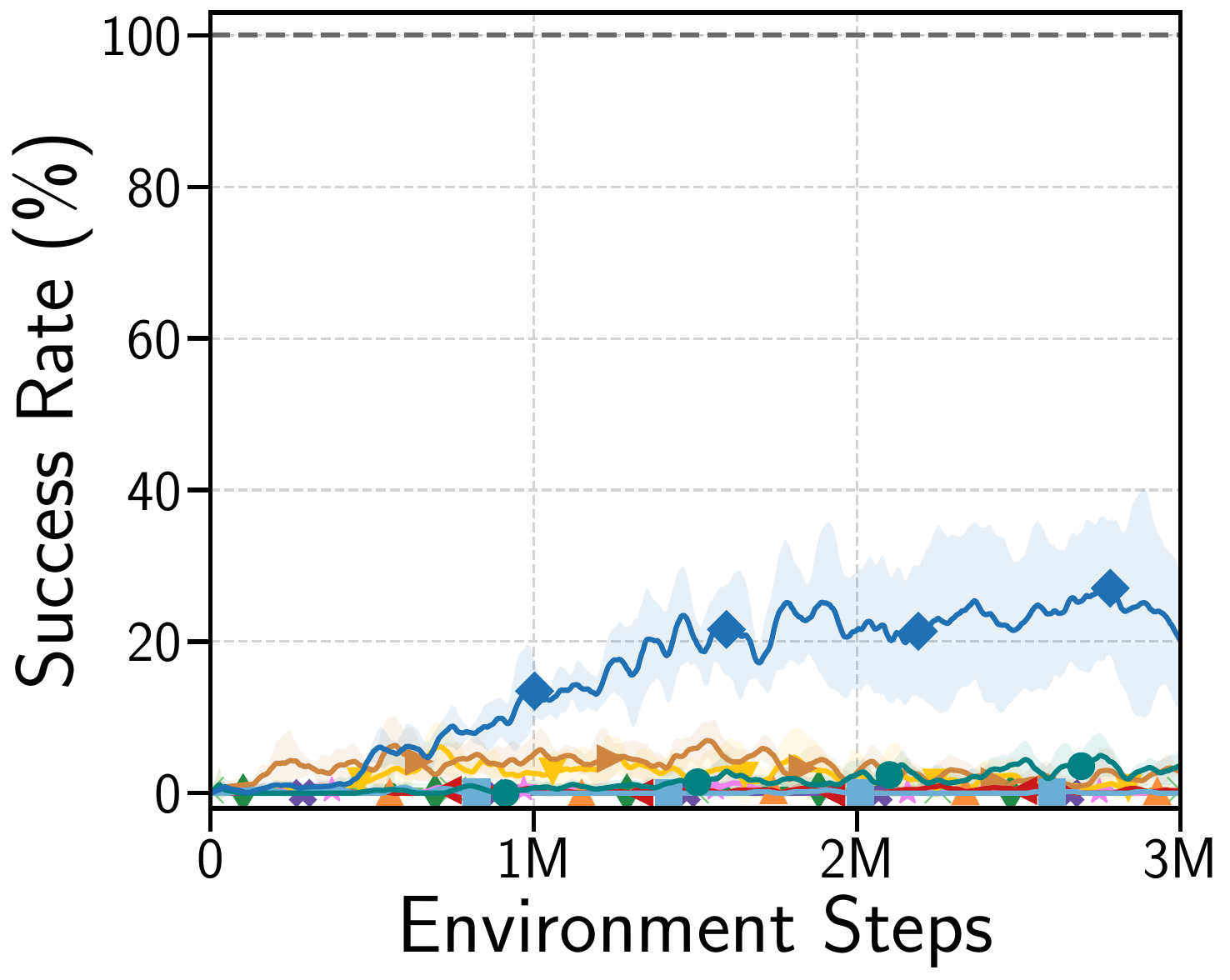}
         \caption{25 Trajectories}
         \label{fig:ant_maze_25}
     \end{subfigure}
    \caption{\myparagraph{Data efficiency}
    We vary the amount of available expert demonstrations in \antmaze{}.
    Our proposed method \method{} consistently outperforms other methods when the number of expert demonstrations decreases, highlighting the data efficiency of \method{}.
    }
    \label{fig:data_efficiency}
\end{figure}

To investigate the data efficiency, \ie how much expert data is required for learning,
we vary the number of expert trajectories in \antmaze~and report the performance of all the methods in~\cref{fig:data_efficiency}. 
Specifically, we use $200$, $100$, $50$, and $25$ expert trajectories, each containing \num{14000}, \num{7000}, \num{3500}, and \num{1750} transitions, respectively.

The results demonstrate that DIFO learns faster compared to all the baselines with various amounts of demonstrations, highlighting its sample efficiency. 
Specifically, as the number of demonstrations decreases from $200$ to $50$, DIFO's performance drops modestly from an $80\%$ success rate to $70\%$, whereas WAILfO, the best-performing baseline when given $200$ expert trajectories, experiences a substantial decline to a $20\%$ success rate. 
Furthermore, when the number of demonstrations is reduced to $25$, all other baselines fail to learn, with success rates nearing zero. In contrast, DIFO maintains a success rate of around $20\%$, underscoring its superior data efficiency.
This data efficiency highlights DIFO's potential for real-world applications, where collecting expert demonstrations can be costly.

\vspacesubsection{Generating data using diffusion models}
\label{sec:generate_data}
\definecolor{MazeRed}{RGB}{236, 8, 10}
\definecolor{MazeBlue}{RGB}{35, 111, 164}
\definecolor{MazeGreen}{RGB}{70, 142, 70}
\definecolor{MazeOrange}{RGB}{208, 111, 15}

\begin{wrapfigure}[18]{R}{0.36\textwidth}
    % \vspace{-4em}
    \vspace{-1.31cm}
    \centering
    \includegraphics[width=\linewidth]{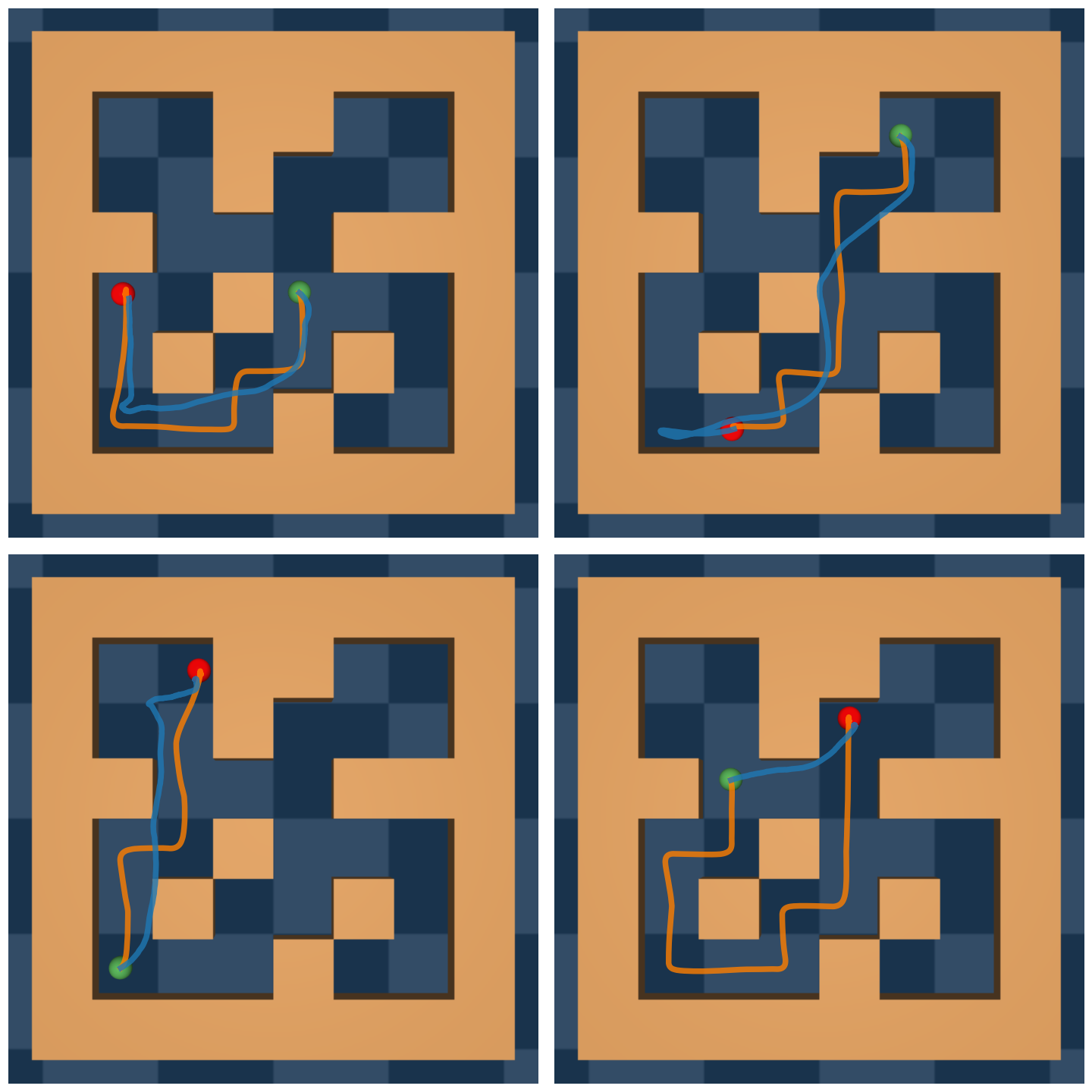}
    \caption{\myparagraph{Generated trajectories under {\pointmaze}}
    The \textcolor{ForestGreen}{green} point marks the initial state.
    The \textcolor{red}{red} point marks the goal.
    The \textcolor{MazeBlue}{blue} trace represents the generated trajectory and the \textcolor{orange}{orange} trace represents the corresponding expert trajectory.
    }
    \label{fig:maze_traj}
\end{wrapfigure}

To investigate whether the DIFO diffusion model can closely capture the expert distribution, we generate trajectories with the diffusion model in \pointmaze{}. 
Specifically, we take a trained diffusion discriminator of DIFO and autoregressively generate a sequence of next states starting from an initial state sampled in the expert dataset.
We visualize four pairs of expert trajectories and the corresponding generated trajectories in~\cref{fig:maze_traj}. 

The results show that our diffusion model can accurately generate trajectories similar to those of the expert.
It is worth noting that the diffusion model can generate trajectories that differ from the expert trajectories while still completing the task, such as the example on the bottom right of~\cref{fig:maze_traj}, where the diffusion model produces even shorter trajectories than the scripted expert policy.
Additional expert trajectories and the corresponding generated trajectories are presented in~\cref{appendix:full_maze}.

\vspacesubsection{Visualized learned reward functions}
\label{sec:visualized_reward}
\begin{figure}
     \centering
     \captionsetup[subfigure]{aboveskip=0.05cm,belowskip=0.05cm}
     \begin{subfigure}[b]{0.215\textwidth}
         \centering
         \includegraphics[height=3.01cm]{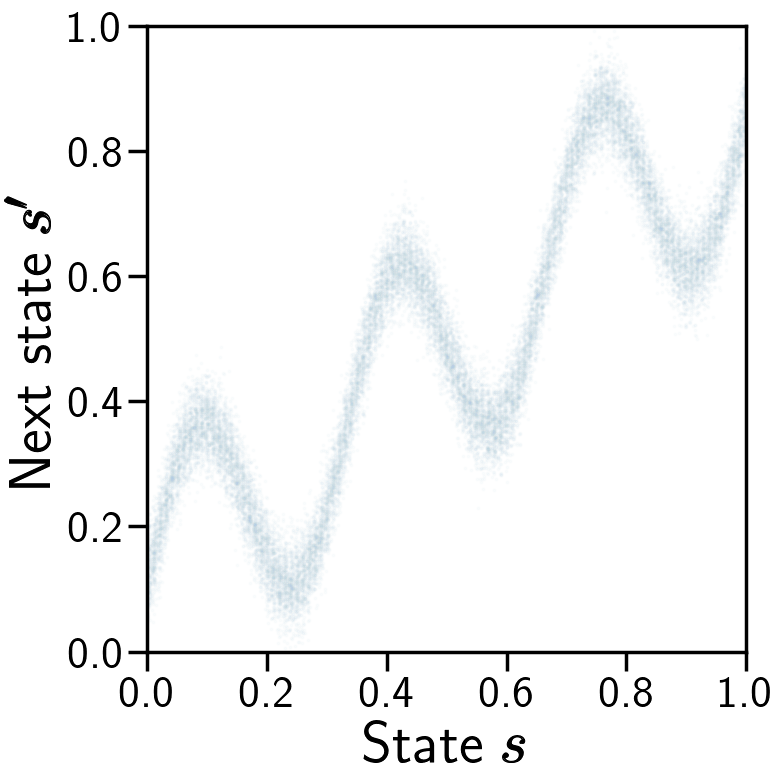}
         \caption{Expert $(s, s')$}
         \label{fig:sine-expert}
     \end{subfigure}
     \begin{subfigure}[b]{0.215\textwidth}
         \centering
         \includegraphics[height=3.01cm]{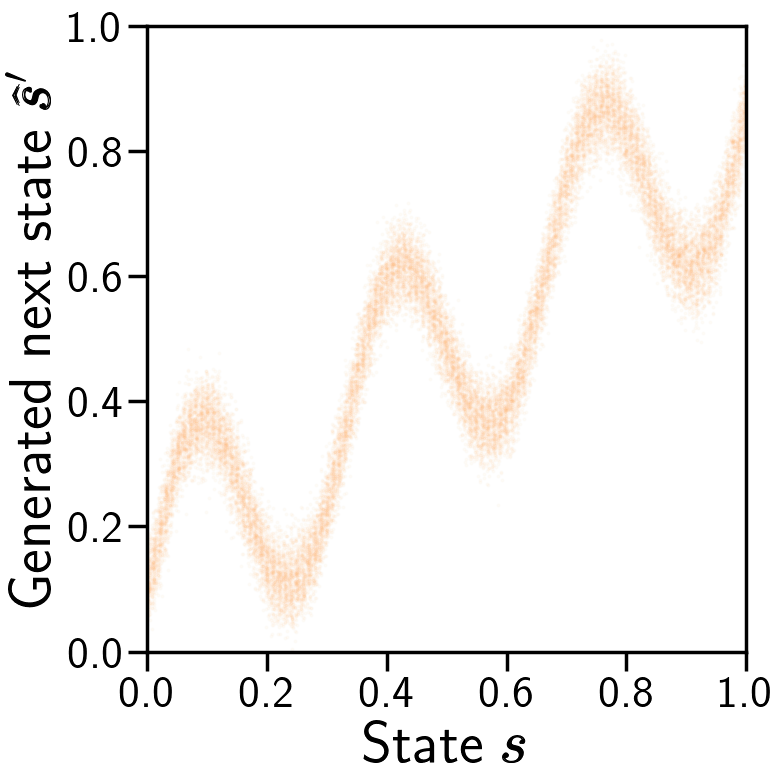}
         \caption{Generated $(s, \widehat{s}')$}
         \label{fig:sine-gen}
     \end{subfigure}
     \begin{subfigure}[b]{0.260\textwidth}
         \centering
         \includegraphics[height=3.01cm]{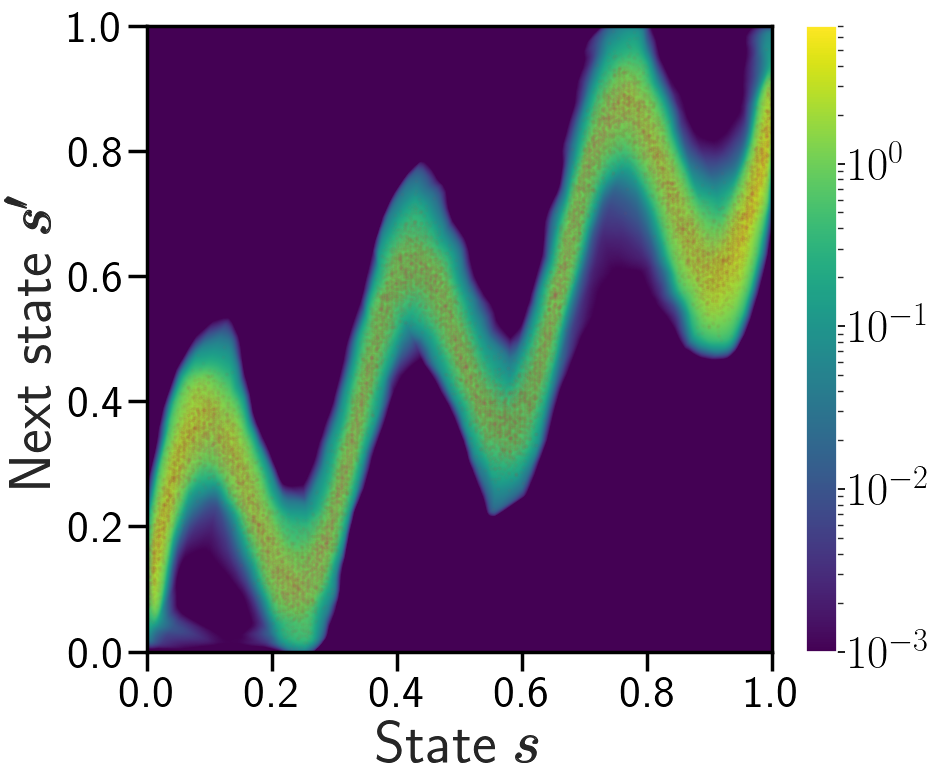}
         \caption{GAIfO reward}
         \label{fig:GAIfO-sine}
     \end{subfigure}
     % \hfill
     \begin{subfigure}[b]{0.260\textwidth}
         \centering
         \includegraphics[height=3.01cm]{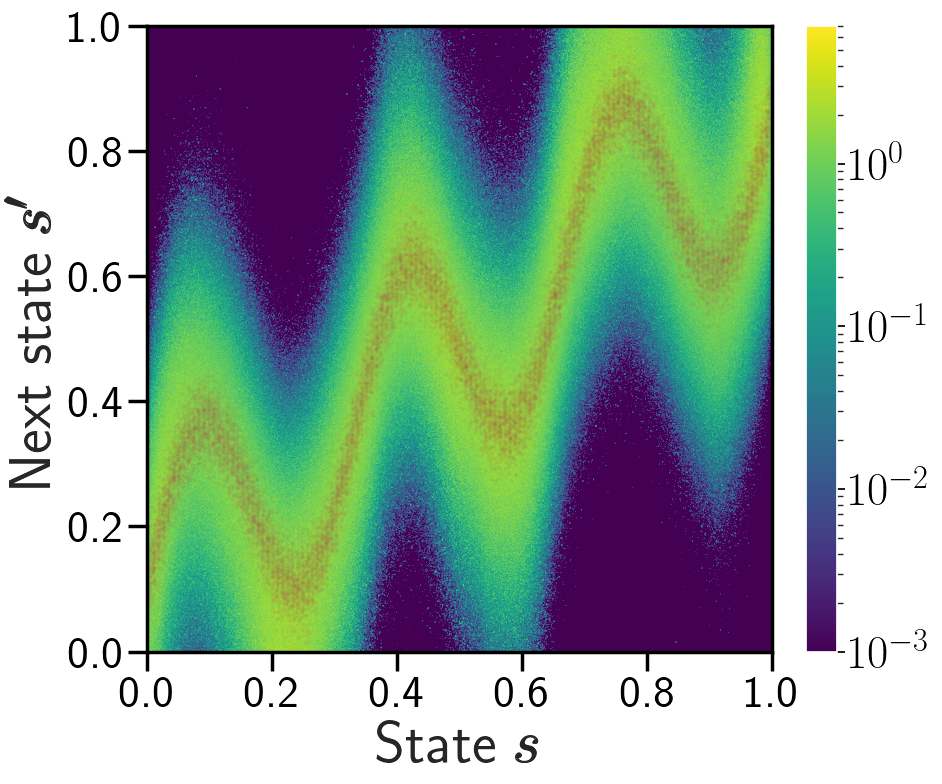}
         \caption{DIFO reward}
         \label{fig:DIFO-sine}
     \end{subfigure}
    \caption{\myparagraph{Reward function visualization and generated distribution on 
    \sine}
    \textbf{(a)} The expert state transition distribution.
    \textbf{(b)} The state transition distribution generated by the DIFO diffusion model.    
    \textbf{(c-d)} The visualized reward functions learned by GAIfO and DIFO, respectively. 
    DIFO produces smoother rewards outside of the expert distribution, allowing for facilitating policy learning.
    }
    \label{fig:sine}
\end{figure}

We aim to visualize and analyze the reward functions learned by DIFO. 
To this end, we introduce a toy environment \sine{} in which both the state and action space are 1-dimension with range $[0,1]$.
We generate expert state-only demonstrations by sampling from the distribution $s' = \sin{6\pi s}+s+\mathcal{N}(0, \,0.05^2)$.
The sampled expert state transitions are plotted in~\cref{fig:sine-expert}.

\myparagraph{Reconstructed distribution}
Given the expert state distribution, we generate a distribution of next states using the diffusion model of DIFO and visualize the distribution in~\cref{fig:sine-gen}.
The generated distribution closely resembles the expert distribution, which again verifies the modeling capability of the conditional diffusion model.

\myparagraph{Visualized learned reward functions} We visualize the reward functions learned by GAIfO and DIFO in {\sine} in~\cref{fig:GAIfO-sine} and~\cref{fig:DIFO-sine}, respectively. 
The result shows that the reward function learned by GAIfO drops dramatically once it deviates from expert distribution, while that of DIFO presents a smoother contour to the region outside the distribution, which allows for bringing a learning agent closer to the expert even when agent's behaviors are far from the expert behavior. 

\begin{figure}[t]
    \centering
    \includegraphics[trim={0cm 2.5cm 0cm 2.6cm},clip,width=\textwidth]{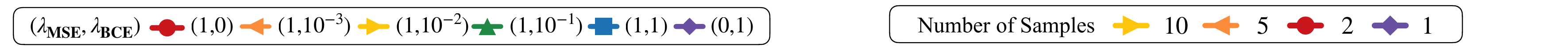}
    \begin{minipage}[t]{0.49\textwidth}
        \centering
        \vtop{
        \begin{subfigure}[t]{0.495\textwidth}
             \centering
             \includegraphics[width=\textwidth]{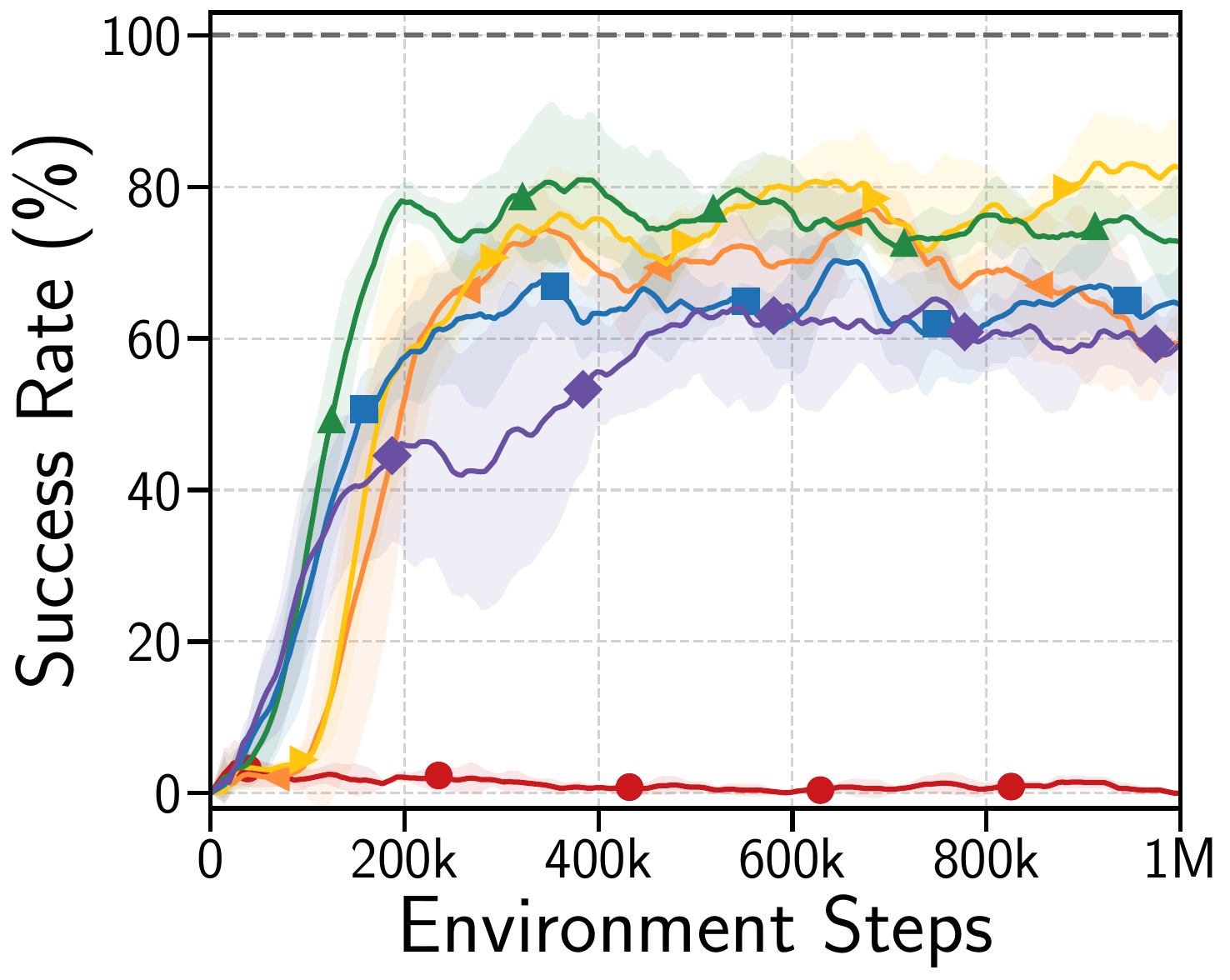}
             \caption{\pointmaze}
             \label{fig:ratio_point_maze}
        \end{subfigure}
        \hspace{-4.5pt}
        \begin{subfigure}[t]{0.495\textwidth}
             \centering
             \includegraphics[width=\textwidth]{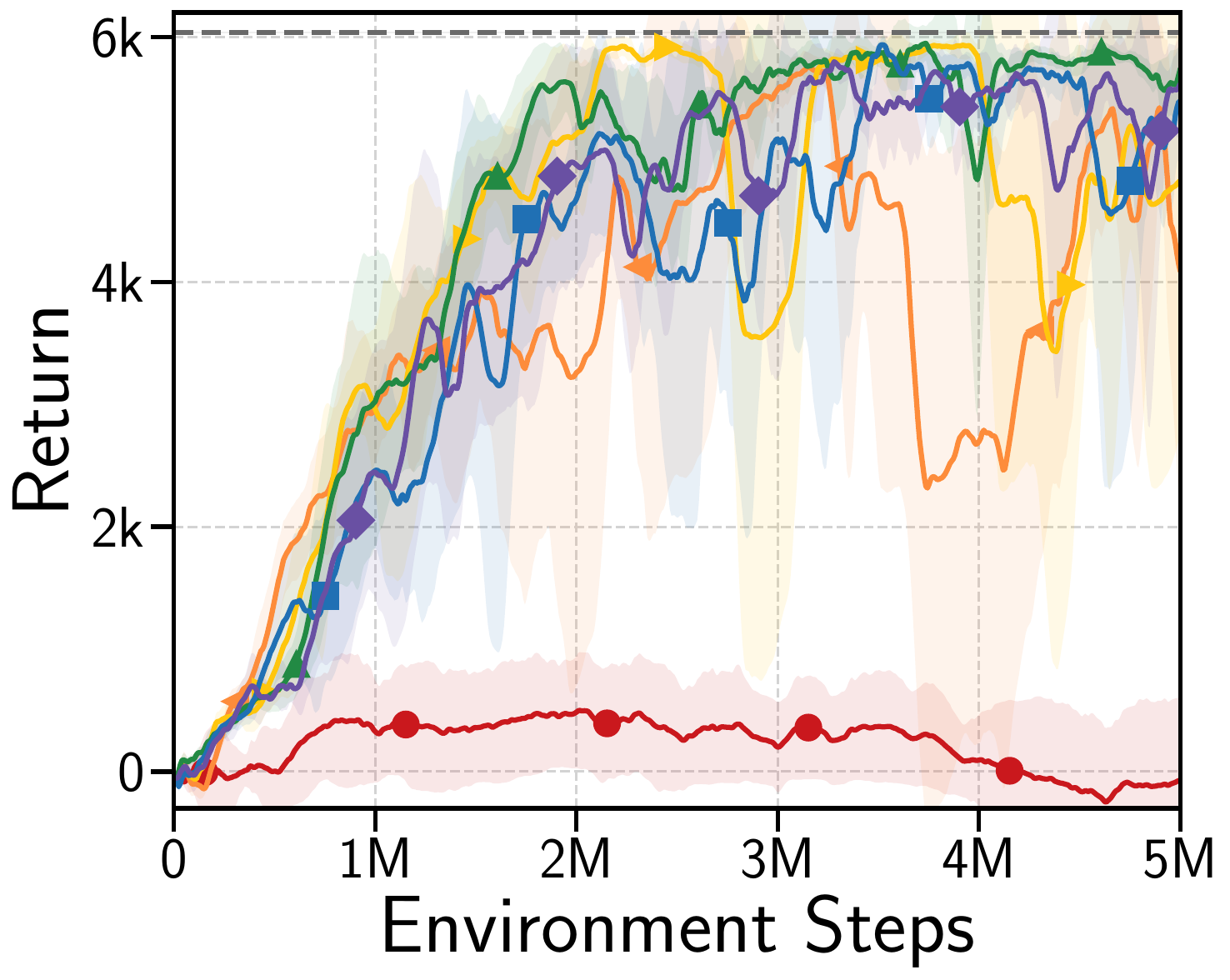}
             \caption{\walker}
             \label{fig:ratio_walker}
        \end{subfigure}
        }
        \caption{\myparagraph{The effect of $\mathbf{\lambda_{\text{MSE}}}$ and $\mathbf{\lambda_{\text{BCE}}}$}
        We vary the values of $\lambda_{\text{MSE}}$ and $\lambda_{\text{BCE}}$ in \pointmaze{} and \walker{}, showcasing DIFO's robustness to hyperparameters and emphasizing the importance of both $\mathcal{L}_{\text{BCE}}$ and $\mathcal{L}_{\text{MSE}}$.
        }
        \label{fig:ablation_ratio}
    \end{minipage}
    \hspace{0.5pt}
    \begin{minipage}[t]{0.49\textwidth}
        \centering
        \vtop{
        \begin{subfigure}[t]{0.495\textwidth}
             \centering
             \includegraphics[width=\textwidth]{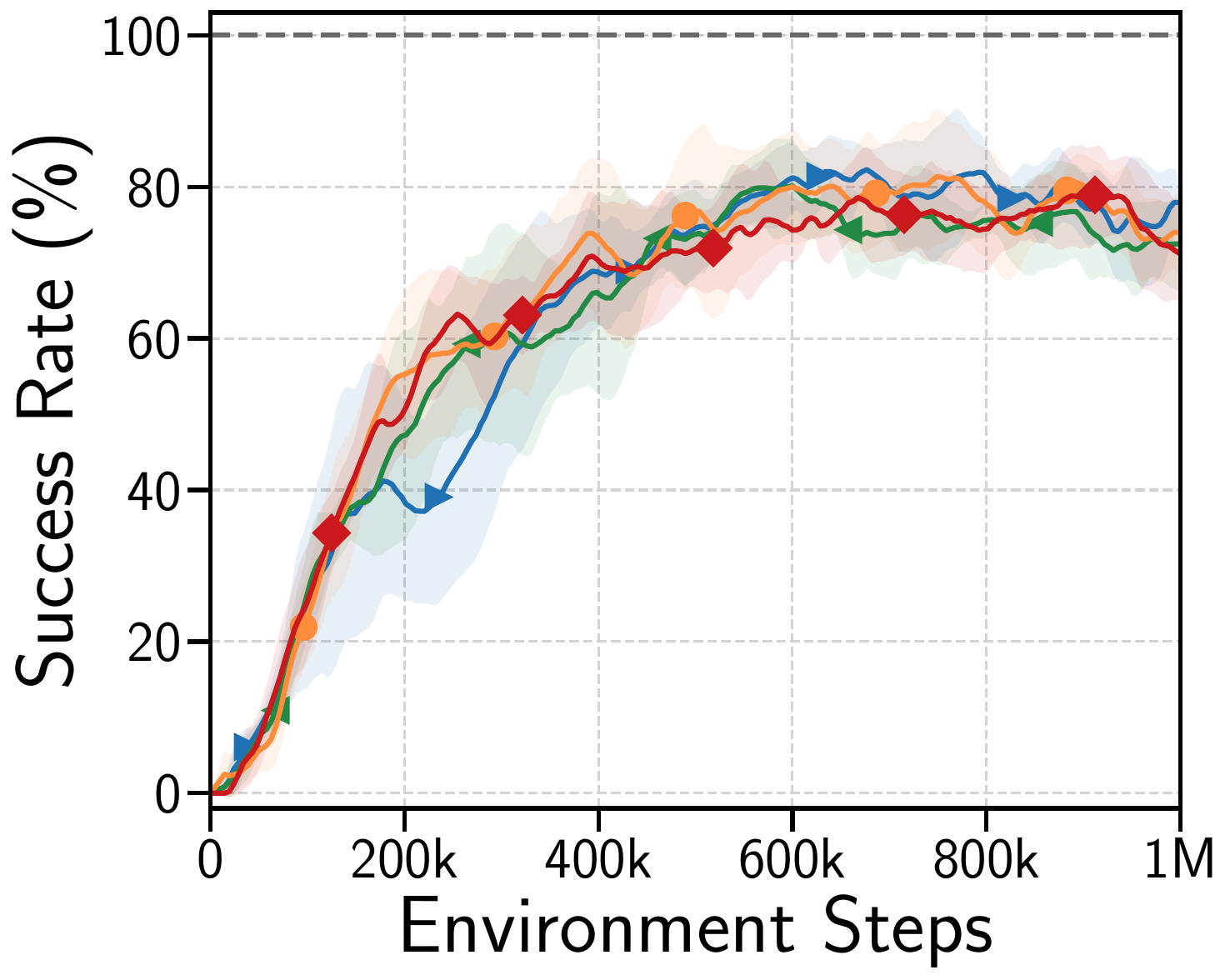}
             \caption{\pointmaze{}}
             \label{fig:sample_step-pointmaze}
        \end{subfigure}
        \hspace{-4.5pt}
        \begin{subfigure}[t]{0.495\textwidth}
             \centering
             \includegraphics[width=\textwidth]{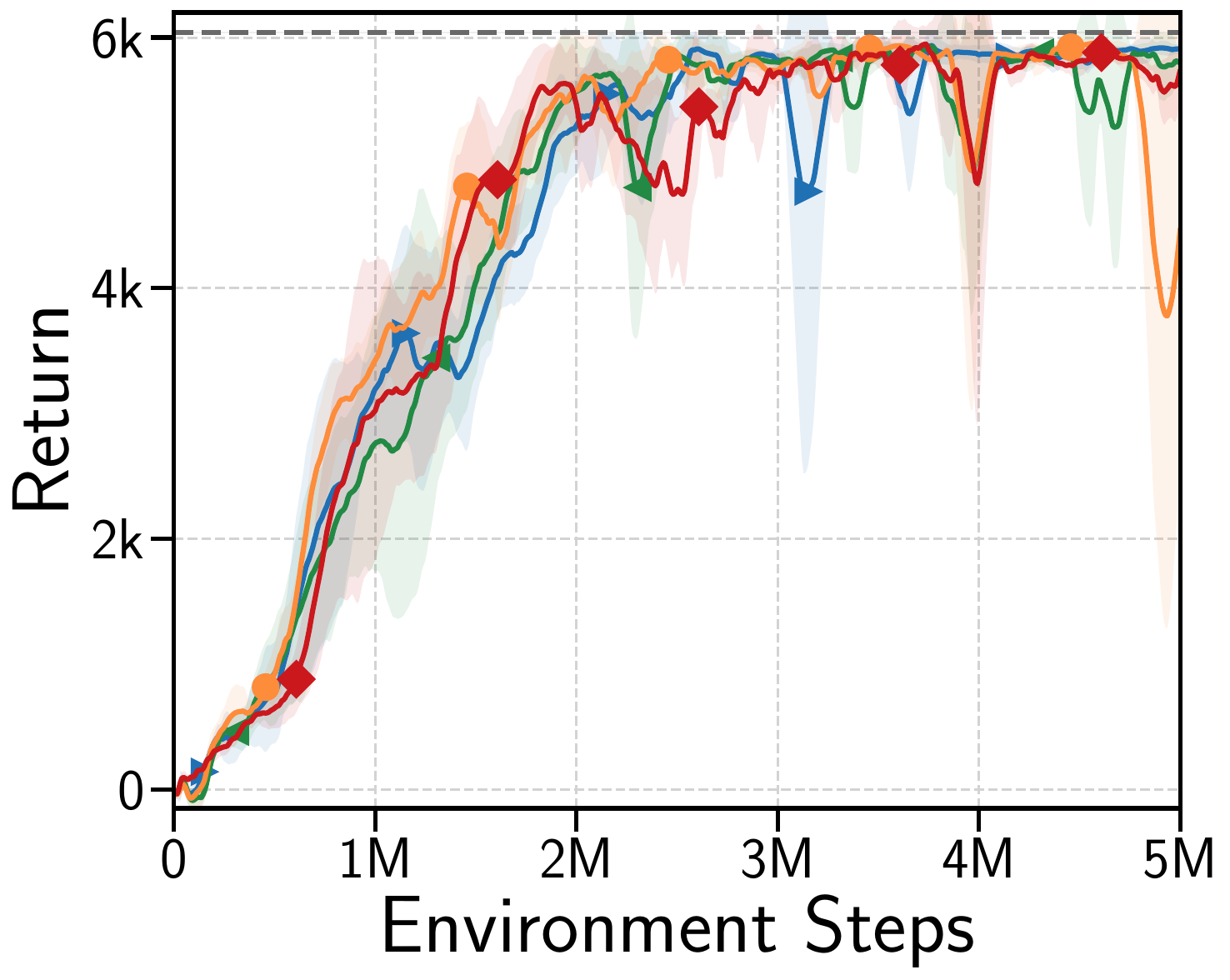}
             \caption{\walker{}}
             \label{fig:sample_step-walker}
        \end{subfigure}
        }
        \caption{\myparagraph{Different numbers of denoising step samples for reward computation} 
        We vary the number of denoising step samples to compute rewards. The result indicates the number of samples does not significantly affect the performance.
        }
        \label{fig:sample_step}
    \end{minipage}
\end{figure}
\vspacesubsection{Ablation study on \texorpdfstring{$\lambda_{\text{MSE}}$}{λMSE} and \texorpdfstring{$\lambda_{\text{BCE}}$}{λBCE}}
\label{sec:ablation_Study}

We hypothesize that both $\mathcal{L}_{\text{MSE}}$ and $\mathcal{L}_{\text{BCE}}$ are important for efficiency learning.
To examine the effect of $\mathcal{L}_{\text{MSE}}$ and $\mathcal{L}_{\text{BCE}}$ and verify the hypothesis, we vary the ratio of $\lambda_{\text{MSE}}$ and $\lambda_{\text{BCE}}$ in {\pointmaze} and {\walker},
including $\mathcal{L}_{\text{BCE}}$ only
and $\mathcal{L}_{\text{MSE}}$ only, \ie $\lambda_{\text{MSE}}=0$ and $\lambda_{\text{BCE}}=0$.
As shown in~\cref{fig:ablation_ratio}, the results emphasize the significance of introducing both $\mathcal{L}_{\text{MSE}}$ and $\mathcal{L}_{\text{BCE}}$, since they enable the model to simultaneously model expert behavior ($\mathcal{L}_{\text{MSE}}$) and perform binary classification ($\mathcal{L}_{\text{BCE}}$).
Without $\mathcal{L}_{\text{MSE}}$, the performance slightly decreases as it does not modeling expert behaviors. 
Without $\mathcal{L}_{\text{BCE}}$, the model fails to learn as it does not utilize negative samples, \ie agent data.
Moreover, when we vary the ratio of $\lambda_{\text{MSE}}$ and $\lambda_{\text{BCE}}$, DIFO maintains stable performance, demonstrating {\method} is relatively insensitive to hyperparameter variations.

\vspacesubsection{Ablation study on the number of samples for reward computation}
\label{sec:sample-timestep}

To investigate the robustness of our rewards, we conducted experiments with varying numbers of denoising step samples in \pointmaze{} and \walker{}.
We take the mean of losses computed from different numbers of samples, \ie multiple $t$, to compute rewards.
As presented in the~\cref{fig:sample_step}, the performance of \method{} is stable under different numbers of samples.
As a result, we use a single denoising step sample to compute the reward for the best efficiency. 
We also investigate the stability of rewards under different numbers of samples in~\cref{appendix:sample_timestep}.
\vspacesection{Conclusion}
\label{sec:conclusion}
We present Diffusion Imitation from Observation (DIFO), a novel adversarial imitation learning from observation framework.
DIFO leverages a conditional diffusion model as a discriminator to distinguish expert state transitions from those of the agent, while the agent policy learns to produce state transitions that are indistinguishable from the expert's for the diffusion discriminator.
Experimental results demonstrate that DIFO outperforms existing learning from observation methods, including BCO, GAIfO, WAIfO, AIRLfO, DePO, OT, and IQ-Learn, across continuous control tasks in various domains, including navigation, manipulation, locomotion, and image-based games.
The visualization of the reward function learned by DIFO shows that it can generalize well to states unseen from expert state transitions by utilizing agent data.
Also, the DIFO diffusion model is able to accurately capture expert state transitions and can generate predicted trajectories that are similar to those of expert's.

\section*{Acknowledgement}
This work was supported by the National Science and
Technology Council, Taiwan (113-2222-E-002-007-).
Shao-Hua Sun was supported by the Yushan Fellow Program by the Ministry of Education, Taiwan.

%%%%%%%%%%%%%%%%%%%%%%%%%%%%%%%%%%%%%%%%%%%%%%%%%%%%%%%%%%%%
% \clearpage

\renewcommand{\bibname}{References}
\bibliographystyle{plainnat}
\bibliography{ref}

\begin{thebibliography}{86}
\providecommand{\natexlab}[1]{#1}
\providecommand{\url}[1]{\texttt{#1}}
\expandafter\ifx\csname urlstyle\endcsname\relax
  \providecommand{\doi}[1]{doi: #1}\else
  \providecommand{\doi}{doi: \begingroup \urlstyle{rm}\Url}\fi

\bibitem[Abbeel and Ng(2004)]{abbeel2004apprenticeship}
Pieter Abbeel and Andrew~Y Ng.
\newblock Apprenticeship learning via inverse reinforcement learning.
\newblock In \emph{International Conference on Machine Learning}, 2004.

\bibitem[Barde et~al.(2020)Barde, Roy, Jeon, Pineau, Pal, and Nowrouzezahrai]{barde2020adversarial}
Paul Barde, Julien Roy, Wonseok Jeon, Joelle Pineau, Chris Pal, and Derek Nowrouzezahrai.
\newblock Adversarial soft advantage fitting: Imitation learning without policy optimization.
\newblock In \emph{Neural Information Processing Systems}, 2020.

\bibitem[Bhateja et~al.(2023)Bhateja, Guo, Ghosh, Singh, Tomar, Vuong, Chebotar, Levine, and Kumar]{bhateja2023robotic}
Chethan Bhateja, Derek Guo, Dibya Ghosh, Anikait Singh, Manan Tomar, Quan Vuong, Yevgen Chebotar, Sergey Levine, and Aviral Kumar.
\newblock Robotic offline rl from internet videos via value-function pre-training.
\newblock In \emph{International Conference on Robotics and Automation}, 2023.

\bibitem[Blattmann et~al.(2023)Blattmann, Dockhorn, Kulal, Mendelevitch, Kilian, Lorenz, Levi, English, Voleti, Letts, et~al.]{blattmann2023stable}
Andreas Blattmann, Tim Dockhorn, Sumith Kulal, Daniel Mendelevitch, Maciej Kilian, Dominik Lorenz, Yam Levi, Zion English, Vikram Voleti, Adam Letts, et~al.
\newblock Stable video diffusion: Scaling latent video diffusion models to large datasets.
\newblock \emph{arXiv:2311.15127}, 2023.

\bibitem[Chen et~al.(2024)Chen, Wang, Hsu, Lai, and Sun]{chen2024diffusion}
Shang-Fu Chen, Hsiang-Chun Wang, Ming-Hao Hsu, Chun-Mao Lai, and Shao-Hua Sun.
\newblock Diffusion model-augmented behavioral cloning.
\newblock In \emph{International Conference on Machine Learning}, 2024.

\bibitem[Chi et~al.(2023)Chi, Feng, Du, Xu, Cousineau, Burchfiel, and Song]{chi2023diffusionpolicy}
Cheng Chi, Siyuan Feng, Yilun Du, Zhenjia Xu, Eric Cousineau, Benjamin Burchfiel, and Shuran Song.
\newblock Diffusion policy: Visuomotor policy learning via action diffusion.
\newblock In \emph{Robotics: Science and Systems}, 2023.

\bibitem[Choi et~al.(2021)Choi, Kim, and Yeo]{choi2021trajgail}
Seongjin Choi, Jiwon Kim, and Hwasoo Yeo.
\newblock Trajgail: Generating urban vehicle trajectories using generative adversarial imitation learning.
\newblock \emph{Transportation Research Part C: Emerging Technologies}, 2021.

\bibitem[Coelho et~al.(2024)Coelho, Oliveira, and Santos]{Coelho_Oliveira_Santos_2024}
Daniel Coelho, Miguel Oliveira, and Vitor Santos.
\newblock Rlfold: Reinforcement learning from online demonstrations in urban autonomous driving.
\newblock In \emph{Association for the Advancement of Artificial Intelligence}, 2024.

\bibitem[Collaboration et~al.(2024)Collaboration, O'Neill, Rehman, Maddukuri, Gupta, Padalkar, Lee, Pooley, et~al.]{embodimentcollaboration2024open}
Embodiment Collaboration, Abby O'Neill, Abdul Rehman, Abhiram Maddukuri, Abhishek Gupta, Abhishek Padalkar, Abraham Lee, Acorn Pooley, et~al.
\newblock Open x-embodiment: Robotic learning datasets and rt-x models.
\newblock In \emph{International Conference on Robotics and Automation}, 2024.

\bibitem[de~Lazcano et~al.(2023)de~Lazcano, Andreas, Tai, Lee, and Terry]{gymnasium_robotics2023github}
Rodrigo de~Lazcano, Kallinteris Andreas, Jun~Jet Tai, Seungjae~Ryan Lee, and Jordan Terry.
\newblock Gymnasium robotics, 2023.

\bibitem[Devin et~al.(2017)Devin, Gupta, Darrell, Abbeel, and Levine]{devin2017learning}
Coline Devin, Abhishek Gupta, Trevor Darrell, Pieter Abbeel, and Sergey Levine.
\newblock Learning modular neural network policies for multi-task and multi-robot transfer.
\newblock In \emph{International Conference on Robotics and Automation}, 2017.

\bibitem[Escontrela et~al.(2023)Escontrela, Adeniji, Yan, Jain, Peng, Goldberg, Lee, Hafner, and Abbeel]{escontrela2024video}
Alejandro Escontrela, Ademi Adeniji, Wilson Yan, Ajay Jain, Xue~Bin Peng, Ken Goldberg, Youngwoon Lee, Danijar Hafner, and Pieter Abbeel.
\newblock Video prediction models as rewards for reinforcement learning.
\newblock In \emph{Neural Information Processing Systems}, 2023.

\bibitem[Fu et~al.(2018)Fu, Luo, and Levine]{fu2018learning}
Justin Fu, Katie Luo, and Sergey Levine.
\newblock Learning robust rewards with adverserial inverse reinforcement learning.
\newblock In \emph{International Conference on Learning Representations}, 2018.

\bibitem[Fu et~al.(2020)Fu, Kumar, Nachum, Tucker, and Levine]{fu2020d4rl}
Justin Fu, Aviral Kumar, Ofir Nachum, George Tucker, and Sergey Levine.
\newblock D4rl: Datasets for deep data-driven reinforcement learning.
\newblock \emph{arXiv:2004.07219}, 2020.

\bibitem[Garg et~al.(2021)Garg, Chakraborty, Cundy, Song, and Ermon]{garg2021iq}
Divyansh Garg, Shuvam Chakraborty, Chris Cundy, Jiaming Song, and Stefano Ermon.
\newblock Iq-learn: Inverse soft-q learning for imitation.
\newblock In \emph{Neural Information Processing Systems}, 2021.

\bibitem[Ghosh et~al.(2023)Ghosh, Bhateja, and Levine]{ghosh2023reinforcement}
Dibya Ghosh, Chethan~Anand Bhateja, and Sergey Levine.
\newblock Reinforcement learning from passive data via latent intentions.
\newblock In \emph{International Conference on Learning Representations}, 2023.

\bibitem[Giusti et~al.(2016)Giusti, Guzzi, Cireşan, He, Rodríguez, Fontana, Faessler, Forster, Schmidhuber, Caro, Scaramuzza, and Gambardella]{Giusti2016Visual}
Alessandro Giusti, Jérôme Guzzi, Dan~C. Cireşan, Fang-Lin He, Juan~P. Rodríguez, Flavio Fontana, Matthias Faessler, Christian Forster, Jürgen Schmidhuber, Gianni~Di Caro, Davide Scaramuzza, and Luca~M. Gambardella.
\newblock A machine learning approach to visual perception of forest trails for mobile robots.
\newblock \emph{IEEE Robotics and Automation Letters}, 2016.

\bibitem[Gleave et~al.(2022)Gleave, Taufeeque, Rocamonde, Jenner, Wang, Toyer, Ernestus, Belrose, Emmons, and Russell]{gleave2022imitation}
Adam Gleave, Mohammad Taufeeque, Juan Rocamonde, Erik Jenner, Steven~H. Wang, Sam Toyer, Maximilian Ernestus, Nora Belrose, Scott Emmons, and Stuart Russell.
\newblock imitation: Clean imitation learning implementations.
\newblock \emph{arXiv:2211.11972}, 2022.

\bibitem[Goodfellow et~al.(2014)Goodfellow, Pouget-Abadie, Mirza, Xu, Warde-Farley, Ozair, Courville, and Bengio]{goodfellow2014generative}
Ian Goodfellow, Jean Pouget-Abadie, Mehdi Mirza, Bing Xu, David Warde-Farley, Sherjil Ozair, Aaron Courville, and Yoshua Bengio.
\newblock Generative adversarial nets.
\newblock In \emph{Advances in Neural Information Processing Systems}, 2014.

\bibitem[Gupta et~al.(2016)Gupta, Eppner, Levine, and Abbeel]{gupta2016learning}
Abhishek Gupta, Clemens Eppner, Sergey Levine, and Pieter Abbeel.
\newblock Learning dexterous manipulation for a soft robotic hand from human demonstrations.
\newblock In \emph{IEEE/RSJ International Conference on Intelligent Robots and Systems}, 2016.

\bibitem[Haarnoja et~al.(2018)Haarnoja, Zhou, Abbeel, and Levine]{tuomas2018sac}
Tuomas Haarnoja, Aurick Zhou, Pieter Abbeel, and Sergey Levine.
\newblock Soft actor-critic: Off-policy maximum entropy deep reinforcement learning with a stochastic actor.
\newblock In \emph{International Conference on Machine Learning}, 2018.

\bibitem[Harmer et~al.(2018)Harmer, Gissl{\'e}n, del Val, Holst, Bergdahl, Olsson, Sj{\"o}{\"o}, and Nordin]{harmer2018imitation}
Jack Harmer, Linus Gissl{\'e}n, Jorge del Val, Henrik Holst, Joakim Bergdahl, Tom Olsson, Kristoffer Sj{\"o}{\"o}, and Magnus Nordin.
\newblock Imitation learning with concurrent actions in 3d games.
\newblock In \emph{IEEE Conference on Computational Intelligence and Games}, 2018.

\bibitem[Henderson et~al.(2018)Henderson, Chang, Bacon, Meger, Pineau, and Precup]{Henderson_2018}
Peter Henderson, Wei-Di Chang, Pierre-Luc Bacon, David Meger, Joelle Pineau, and Doina Precup.
\newblock Optiongan: Learning joint reward-policy options using generative adversarial inverse reinforcement learning.
\newblock In \emph{Association for the Advancement of Artificial Intelligence}, 2018.

\bibitem[Ho and Ermon(2016)]{ho2016generative}
Jonathan Ho and Stefano Ermon.
\newblock Generative adversarial imitation learning.
\newblock In \emph{Advances in Neural Information Processing Systems}, 2016.

\bibitem[Huang et~al.(2024)Huang, Jiang, Ze, and Xu]{Huang2023DiffusionReward}
Tao Huang, Guangqi Jiang, Yanjie Ze, and Huazhe Xu.
\newblock Diffusion reward: Learning rewards via conditional video diffusion.
\newblock In \emph{European Conference on Computer Vision}, 2024.

\bibitem[Hussein et~al.(2017)Hussein, Gaber, Elyan, and Jayne]{hussein2017imitation}
Ahmed Hussein, Mohamed~Medhat Gaber, Eyad Elyan, and Chrisina Jayne.
\newblock Imitation learning: A survey of learning methods.
\newblock \emph{ACM Computing Surveys}, 2017.

\bibitem[J~Ho(2020)]{ho2020ddpm}
A~Jain J~Ho.
\newblock Denoising diffusion probabilistic models.
\newblock In \emph{Advances in Neural Information Processing Systems}, 2020.

\bibitem[Jang et~al.(2022)Jang, Irpan, Khansari, Kappler, Ebert, Lynch, Levine, and Finn]{jang2022BCZ}
Eric Jang, Alex Irpan, Mohi Khansari, Daniel Kappler, Frederik Ebert, Corey Lynch, Sergey Levine, and Chelsea Finn.
\newblock Bc-z: Zero-shot task generalization with robotic imitation learning.
\newblock In \emph{Conference on Robot Learning}, 2022.

\bibitem[Janner et~al.(2022)Janner, Du, Tenenbaum, and Levine]{janner2022diffuser}
Michael Janner, Yilun Du, Joshua Tenenbaum, and Sergey Levine.
\newblock Planning with diffusion for flexible behavior synthesis.
\newblock In \emph{International Conference on Machine Learning}, 2022.

\bibitem[Johns(2021)]{johns2021coarse}
Edward Johns.
\newblock Coarse-to-fine imitation learning: Robot manipulation from a single demonstration.
\newblock In \emph{International Conference on Robotics and Automation}, 2021.

\bibitem[Ko et~al.(2024)Ko, Mao, Du, Sun, and Tenenbaum]{ko2024learning}
Po-Chen Ko, Jiayuan Mao, Yilun Du, Shao-Hua Sun, and Joshua~B Tenenbaum.
\newblock Learning to act from actionless videos through dense correspondences.
\newblock In \emph{International Conference on Learning Representations}, 2024.

\bibitem[Kostrikov et~al.(2019)Kostrikov, Agrawal, Dwibedi, Levine, and Tompson]{kostrikov2018discriminatoractorcritic}
Ilya Kostrikov, Kumar~Krishna Agrawal, Debidatta Dwibedi, Sergey Levine, and Jonathan Tompson.
\newblock Discriminator-actor-critic: Addressing sample inefficiency and reward bias in adversarial imitation learning.
\newblock In \emph{International Conference on Learning Representations}, 2019.

\bibitem[Kumar et~al.(2022)Kumar, Singh, Ebert, Nakamoto, Yang, Finn, and Levine]{kumar2023pretraining}
Aviral Kumar, Anikait Singh, Frederik Ebert, Mitsuhiko Nakamoto, Yanlai Yang, Chelsea Finn, and Sergey Levine.
\newblock Pre-training for robots: Offline rl enables learning new tasks from a handful of trials.
\newblock In \emph{Robotics: Science and Systems}, 2022.

\bibitem[Kumar(2016)]{Kumar2016thesis}
Vikash Kumar.
\newblock \emph{Manipulators and Manipulation in high dimensional spaces}.
\newblock PhD thesis, University of Washington, Seattle, 2016.

\bibitem[Lai et~al.(2024)Lai, Wang, Hsieh, Wang, Chen, and Sun]{lai2024diffusion}
Chun-Mao Lai, Hsiang-Chun Wang, Ping-Chun Hsieh, Yu-Chiang~Frank Wang, Min-Hung Chen, and Shao-Hua Sun.
\newblock Diffusion-reward adversarial imitation learning.
\newblock \emph{arXiv:2405.16194}, 2024.

\bibitem[Laskey et~al.(2016)Laskey, Staszak, Hsieh, Mahler, Pokorny, Dragan, and Goldberg]{laskey2016SHIV}
Michael Laskey, Sam Staszak, Wesley Yu-Shu Hsieh, Jeffrey Mahler, Florian~T. Pokorny, Anca~D. Dragan, and Ken Goldberg.
\newblock Shiv: Reducing supervisor burden in dagger using support vectors for efficient learning from demonstrations in high dimensional state spaces.
\newblock In \emph{International Conference on Robotics and Automation}, 2016.

\bibitem[Lee et~al.(2021)Lee, Szot, Sun, and Lim]{goalprox}
Youngwoon Lee, Andrew Szot, Shao-Hua Sun, and Joseph~J. Lim.
\newblock Generalizable imitation learning from observation via inferring goal proximity.
\newblock In \emph{Neural Information Processing Systems}, 2021.

\bibitem[Li et~al.(2023)Li, Prabhudesai, Duggal, Brown, and Pathak]{li2023diffusion}
Alexander~C. Li, Mihir Prabhudesai, Shivam Duggal, Ellis Brown, and Deepak Pathak.
\newblock Your diffusion model is secretly a zero-shot classifier.
\newblock In \emph{International Conference on Computer Vision}, 2023.

\bibitem[Li et~al.(2017)Li, Song, and Ermon]{li2017infogail}
Yunzhu Li, Jiaming Song, and Stefano Ermon.
\newblock Infogail: Interpretable imitation learning from visual demonstrations.
\newblock In \emph{Neural Information Processing Systems}, 2017.

\bibitem[Liu et~al.(2022{\natexlab{a}})Liu, Zhu, Zhuang, Zhang, Hao, Yu, and Wang]{liu2022plan}
Minghuan Liu, Zhengbang Zhu, Yuzheng Zhuang, Weinan Zhang, Jianye Hao, Yong Yu, and Jun Wang.
\newblock Plan your target and learn your skills: Transferable state-only imitation learning via decoupled policy optimization.
\newblock In \emph{International Conference on Machine Learning}, 2022{\natexlab{a}}.

\bibitem[Liu et~al.(2022{\natexlab{b}})Liu, Su, and Yu]{liu2022diffgantts}
Songxiang Liu, Dan Su, and Dong Yu.
\newblock Diffgan-tts: High-fidelity and efficient text-to-speech with denoising diffusion gans.
\newblock \emph{arXiv:2201.11972}, 2022{\natexlab{b}}.

\bibitem[Ma et~al.(2023{\natexlab{a}})Ma, Kumar, Zhang, Bastani, and Jayaraman]{ma2023liv}
Yecheng~Jason Ma, Vikash Kumar, Amy Zhang, Osbert Bastani, and Dinesh Jayaraman.
\newblock Liv: Language-image representations and rewards for robotic control.
\newblock In \emph{International Conference on Machine Learning}, 2023{\natexlab{a}}.

\bibitem[Ma et~al.(2023{\natexlab{b}})Ma, Sodhani, Jayaraman, Bastani, Kumar, and Zhang]{ma2022vip}
Yecheng~Jason Ma, Shagun Sodhani, Dinesh Jayaraman, Osbert Bastani, Vikash Kumar, and Amy Zhang.
\newblock Vip: Towards universal visual reward and representation via value-implicit pre-training.
\newblock \emph{International Conference on Learning Representations}, 2023{\natexlab{b}}.

\bibitem[Mishra et~al.(2023)Mishra, Xue, Chen, and Xu]{mishra2023generative}
Utkarsh~Aashu Mishra, Shangjie Xue, Yongxin Chen, and Danfei Xu.
\newblock Generative skill chaining: Long-horizon skill planning with diffusion models.
\newblock In \emph{Conference on Robot Learning}, 2023.

\bibitem[Ng and Russell(2000)]{ng2000algorithms}
Andrew~Y. Ng and Stuart~J. Russell.
\newblock Algorithms for inverse reinforcement learning.
\newblock In \emph{International Conference on Machine Learning}, 2000.

\bibitem[Osa et~al.(2018)Osa, Pajarinen, Neumann, Bagnell, Abbeel, Peters, et~al.]{osa2018algorithmic}
Takayuki Osa, Joni Pajarinen, Gerhard Neumann, J~Andrew Bagnell, Pieter Abbeel, Jan Peters, et~al.
\newblock An algorithmic perspective on imitation learning.
\newblock \emph{Foundations and Trends{\textregistered} in Robotics}, 2018.

\bibitem[Palma et~al.(2011)Palma, S{\'a}nchez-Ruiz, G{\'o}mez-Mart{\'\i}n, G{\'o}mez-Mart{\'\i}n, and Gonz{\'a}lez-Calero]{Palma2011Combining}
Ricardo Palma, Antonio~A. S{\'a}nchez-Ruiz, Marco~Antonio G{\'o}mez-Mart{\'\i}n, Pedro~Pablo G{\'o}mez-Mart{\'\i}n, and Pedro~Antonio Gonz{\'a}lez-Calero.
\newblock Combining expert knowledge and learning from demonstration in real-time strategy games.
\newblock In \emph{Case-Based Reasoning Research and Development}, 2011.

\bibitem[Papagiannis and Li(2022)]{papagiannis2022imitation}
Georgios Papagiannis and Yunpeng Li.
\newblock Imitation learning with sinkhorn distances.
\newblock In \emph{Joint European Conference on Machine Learning and Knowledge Discovery in Databases}. Springer, 2022.

\bibitem[Pearce et~al.(2023)Pearce, Rashid, Kanervisto, Bignell, Sun, Georgescu, Macua, Tan, Momennejad, Hofmann, and Devlin]{pearce2022imitating}
Tim Pearce, Tabish Rashid, Anssi Kanervisto, David Bignell, Mingfei Sun, Raluca Georgescu, Sergio~Valcarcel Macua, Shan~Zheng Tan, Ida Momennejad, Katja Hofmann, and Sam Devlin.
\newblock Imitating human behaviour with diffusion models.
\newblock In \emph{International Conference on Learning Representations}, 2023.

\bibitem[Perez-Vicente et~al.(2023)Perez-Vicente, Younis, Balis, and Davey]{minari2023minari}
Rodrigo Perez-Vicente, Omar Younis, John Balis, and Alex Davey.
\newblock Minari: A dataset api for offline reinforcement learning, 2023.

\bibitem[Pomerleau(1991)]{pomerleau1991efficient_bc}
Dean~A. Pomerleau.
\newblock Efficient training of artificial neural networks for autonomous navigation.
\newblock \emph{Neural Computation}, 1991.

\bibitem[Poole et~al.(2023)Poole, Jain, Barron, and Mildenhall]{poole2022dreamfusion}
Ben Poole, Ajay Jain, Jonathan~T Barron, and Ben Mildenhall.
\newblock Dreamfusion: Text-to-3d using 2d diffusion.
\newblock In \emph{International Conference on Learning Representations}, 2023.

\bibitem[Popov et~al.(2021)Popov, Vovk, Gogoryan, Sadekova, and Kudinov]{popov2021grad}
Vadim Popov, Ivan Vovk, Vladimir Gogoryan, Tasnima Sadekova, and Mikhail Kudinov.
\newblock Grad-tts: A diffusion probabilistic model for text-to-speech.
\newblock In \emph{International Conference on Machine Learning}, 2021.

\bibitem[Rafailov et~al.(2021)Rafailov, Yu, Rajeswaran, and Finn]{Rafailov2021Visual}
Rafael Rafailov, Tianhe Yu, Aravind Rajeswaran, and Chelsea Finn.
\newblock Visual adversarial imitation learning using variational models.
\newblock In \emph{Neural Information Processing Systems}, 2021.

\bibitem[Raffin et~al.(2021)Raffin, Hill, Gleave, Kanervisto, Ernestus, and Dormann]{stable-baselines3}
Antonin Raffin, Ashley Hill, Adam Gleave, Anssi Kanervisto, Maximilian Ernestus, and Noah Dormann.
\newblock Stable-baselines3: Reliable reinforcement learning implementations.
\newblock \emph{Journal of Machine Learning Research}, 2021.

\bibitem[Rombach et~al.(2022)Rombach, Blattmann, Lorenz, Esser, and Ommer]{rombach2021highresolution}
Robin Rombach, Andreas Blattmann, Dominik Lorenz, Patrick Esser, and Bj{\"o}rn Ommer.
\newblock High-resolution image synthesis with latent diffusion models.
\newblock In \emph{IEEE Conference on Computer Vision and Pattern Recognition}, 2022.

\bibitem[Ronneberger et~al.(2015)Ronneberger, Fischer, and Brox]{ronneberger2015u}
Olaf Ronneberger, Philipp Fischer, and Thomas Brox.
\newblock U-net: Convolutional networks for biomedical image segmentation.
\newblock In \emph{Medical Image Computing and Computer-Assisted Intervention}, 2015.

\bibitem[Ross and Bagnell(2010)]{ross2010efficient}
St{\'e}phane Ross and Drew Bagnell.
\newblock Efficient reductions for imitation learning.
\newblock In \emph{International Conference on Artificial Intelligence and Statistics}, 2010.

\bibitem[Ross et~al.(2011)Ross, Gordon, and Bagnell]{ross2011reduction}
St{\'e}phane Ross, Geoffrey Gordon, and Drew Bagnell.
\newblock A reduction of imitation learning and structured prediction to no-regret online learning.
\newblock In \emph{International Conference on Artificial Intelligence and Statistics}, 2011.

\bibitem[Russell(1998)]{russell1998learning}
Stuart Russell.
\newblock Learning agents for uncertain environments.
\newblock In \emph{Annual Conference on Computational Learning Theory}, 1998.

\bibitem[Saharia et~al.(2022)Saharia, Chan, Saxena, Li, Whang, Denton, Ghasemipour, Gontijo~Lopes, Karagol~Ayan, Salimans, et~al.]{saharia2022photorealistic}
Chitwan Saharia, William Chan, Saurabh Saxena, Lala Li, Jay Whang, Emily~L Denton, Kamyar Ghasemipour, Raphael Gontijo~Lopes, Burcu Karagol~Ayan, Tim Salimans, et~al.
\newblock Photorealistic text-to-image diffusion models with deep language understanding.
\newblock \emph{Neural Information Processing Systems}, 2022.

\bibitem[Samak et~al.(2021)Samak, Samak, and Kandhasamy]{Samak2021Robust}
Tanmay~Vilas Samak, Chinmay~Vilas Samak, and Sivanathan Kandhasamy.
\newblock Robust behavioral cloning for autonomous vehicles using end-to-end imitation learning.
\newblock \emph{SAE International Journal of Connected and Automated Vehicles}, 2021.

\bibitem[Schaal(1997)]{schaal1997learning}
Stefan Schaal.
\newblock Learning from demonstration.
\newblock In \emph{Advances in Neural Information Processing Systems}, 1997.

\bibitem[Scheel et~al.(2022)Scheel, Bergamini, Wolczyk, Osi\'nski, and Ondruska]{scheel2022urbanDriver}
Oliver Scheel, Luca Bergamini, Maciej Wolczyk, B\l{a}\.{z}ej Osi\'nski, and Peter Ondruska.
\newblock Urban driver: Learning to drive from real-world demonstrations using policy gradients.
\newblock In \emph{Conference on Robot Learning}, 2022.

\bibitem[Schmeckpeper et~al.(2020)Schmeckpeper, Rybkin, Daniilidis, Levine, and Finn]{schmeckpeper2020reinforcement}
Karl Schmeckpeper, Oleh Rybkin, Kostas Daniilidis, Sergey Levine, and Chelsea Finn.
\newblock Reinforcement learning with videos: Combining offline observations with interaction.
\newblock In \emph{Conference on Robot Learning}, 2020.

\bibitem[Schmidt and Jiang(2024)]{2024lapo}
Dominik Schmidt and Minqi Jiang.
\newblock Learning to act without actions.
\newblock In \emph{International Conference on Learning Representations}, 2024.

\bibitem[Schulman et~al.(2017)Schulman, Wolski, Dhariwal, Radford, and Klimov]{schulman2017proximal}
John Schulman, Filip Wolski, Prafulla Dhariwal, Alec Radford, and Oleg Klimov.
\newblock Proximal policy optimization algorithms.
\newblock \emph{arXiv:1707.06347}, 2017.

\bibitem[Seo et~al.(2022)Seo, Lee, James, and Abbeel]{seo2022reinforcement}
Younggyo Seo, Kimin Lee, Stephen~L James, and Pieter Abbeel.
\newblock Reinforcement learning with action-free pre-training from videos.
\newblock In \emph{International Conference on Machine Learning}, 2022.

\bibitem[Sermanet et~al.(2018)Sermanet, Lynch, Chebotar, Hsu, Jang, Schaal, Levine, and Brain]{sermanet2018time}
Pierre Sermanet, Corey Lynch, Yevgen Chebotar, Jasmine Hsu, Eric Jang, Stefan Schaal, Sergey Levine, and Google Brain.
\newblock Time-contrastive networks: Self-supervised learning from video.
\newblock In \emph{International Conference on Robotics and Automation}, 2018.

\bibitem[Shafiullah et~al.(2022)Shafiullah, Cui, Altanzaya, and Pinto]{shafiullah2022behavior}
Nur Muhammad~Mahi Shafiullah, Zichen~Jeff Cui, Ariuntuya Altanzaya, and Lerrel Pinto.
\newblock Behavior transformers: Cloning $k$ modes with one stone.
\newblock In \emph{Neural Information Processing Systems}, 2022.

\bibitem[Shih et~al.(2022)Shih, Ermon, and Sadigh]{Shih2022Conditional}
Andy Shih, Stefano Ermon, and Dorsa Sadigh.
\newblock Conditional imitation learning for multi-agent games.
\newblock In \emph{International Conference on Human-Robot Interaction}, 2022.

\bibitem[Song et~al.(2021)Song, Meng, and Ermon]{song2022denoising}
Jiaming Song, Chenlin Meng, and Stefano Ermon.
\newblock Denoising diffusion implicit models.
\newblock In \emph{International Conference on Learning Representations}, 2021.

\bibitem[Sun et~al.(2018)Sun, Noh, Somasundaram, and Lim]{sun2018neural}
Shao-Hua Sun, Hyeonwoo Noh, Sriram Somasundaram, and Joseph Lim.
\newblock Neural program synthesis from diverse demonstration videos.
\newblock In \emph{International Conference on Machine Learning}, 2018.

\bibitem[Todorov et~al.(2012)Todorov, Erez, and Tassa]{todorov2012mujoco}
Emanuel Todorov, Tom Erez, and Yuval Tassa.
\newblock Mujoco: A physics engine for model-based control.
\newblock In \emph{International Conference On Intelligent Robots and Systems}, 2012.

\bibitem[Torabi et~al.(2018)Torabi, Warnell, and Stone]{torabi2018behavioral}
Faraz Torabi, Garrett Warnell, and Peter Stone.
\newblock Behavioral cloning from observation.
\newblock In \emph{International Joint Conference on Artificial Intelligence}, 2018.

\bibitem[Torabi et~al.(2019)Torabi, Warnell, and Stone]{torabi2018generative}
Faraz Torabi, Garrett Warnell, and Peter Stone.
\newblock Generative adversarial imitation from observation.
\newblock In \emph{International Conference on Machine Learning}, 2019.

\bibitem[Towers et~al.(2023)Towers, Terry, Kwiatkowski, Balis, Cola, Deleu, Goulão, Kallinteris, KG, Krimmel, Perez-Vicente, Pierré, Schulhoff, Tai, Shen, and Younis]{towers_gymnasium_2023}
Mark Towers, Jordan~K. Terry, Ariel Kwiatkowski, John~U. Balis, Gianluca~de Cola, Tristan Deleu, Manuel Goulão, Andreas Kallinteris, Arjun KG, Markus Krimmel, Rodrigo Perez-Vicente, Andrea Pierré, Sander Schulhoff, Jun~Jet Tai, Andrew Tan~Jin Shen, and Omar~G. Younis.
\newblock Gymnasium, 2023.

\bibitem[von Platen et~al.(2022)von Platen, Patil, Lozhkov, Cuenca, Lambert, Rasul, Davaadorj, Nair, Paul, Berman, Xu, Liu, and Wolf]{von-platen-etal-2022-diffusers}
Patrick von Platen, Suraj Patil, Anton Lozhkov, Pedro Cuenca, Nathan Lambert, Kashif Rasul, Mishig Davaadorj, Dhruv Nair, Sayak Paul, William Berman, Yiyi Xu, Steven Liu, and Thomas Wolf.
\newblock Diffusers: State-of-the-art diffusion models, 2022.

\bibitem[Wang et~al.(2024)Wang, Wu, Pang, Zhang, and Yin]{wang2024diffail}
Bingzheng Wang, Guoqiang Wu, Teng Pang, Yan Zhang, and Yilong Yin.
\newblock Diffail: Diffusion adversarial imitation learning.
\newblock In \emph{Association for the Advancement of Artificial Intelligence}, 2024.

\bibitem[Wen et~al.(2023)Wen, Lin, So, Chen, Dou, Gao, and Abbeel]{wen2023any}
Chuan Wen, Xingyu Lin, John So, Kai Chen, Qi~Dou, Yang Gao, and Pieter Abbeel.
\newblock Any-point trajectory modeling for policy learning.
\newblock In \emph{Robotics: Science and Systems}, 2023.

\bibitem[Xiao et~al.(2019)Xiao, Herman, Wagner, Ziesche, Etesami, and Linh]{xiao2019wasserstein}
Huang Xiao, Michael Herman, Joerg Wagner, Sebastian Ziesche, Jalal Etesami, and Thai~Hong Linh.
\newblock Wasserstein adversarial imitation learning.
\newblock \emph{arXiv:1906.08113}, 2019.

\bibitem[Yang et~al.(2019)Yang, Ma, Huang, Sun, Liu, Huang, and Gan]{yang2019imitation}
Chao Yang, Xiaojian Ma, Wenbing Huang, Fuchun Sun, Huaping Liu, Junzhou Huang, and Chuang Gan.
\newblock Imitation learning from observations by minimizing inverse dynamics disagreement.
\newblock In \emph{Neural Information Processing Systems}, 2019.

\bibitem[Yu et~al.(2020)Yu, Quillen, He, Julian, Hausman, Finn, and Levine]{yu2019metaworld}
Tianhe Yu, Deirdre Quillen, Zhanpeng He, Ryan Julian, Karol Hausman, Chelsea Finn, and Sergey Levine.
\newblock Meta-world: A benchmark and evaluation for multi-task and meta reinforcement learning.
\newblock In \emph{Conference on robot learning}, 2020.

\bibitem[Zhang et~al.(2024)Zhang, Becker-Ehmck, van~der Smagt, and Karl]{zhang2024action}
Xingyuan Zhang, Philip Becker-Ehmck, Patrick van~der Smagt, and Maximilian Karl.
\newblock Action inference by maximising evidence: zero-shot imitation from observation with world models.
\newblock In \emph{Neural Information Processing Systems}, 2024.

\bibitem[Zhu et~al.(2020)Zhu, Lin, Dai, and Zhou]{zhu2020off}
Zhuangdi Zhu, Kaixiang Lin, Bo~Dai, and Jiayu Zhou.
\newblock Off-policy imitation learning from observations.
\newblock In \emph{Neural Information Processing Systems}, 2020.

\bibitem[Ziebart et~al.(2008)Ziebart, Maas, Bagnell, and Dey]{Ziebart2008MaximumEI}
Brian~D. Ziebart, Andrew~L. Maas, J.~Andrew Bagnell, and Anind~K. Dey.
\newblock Maximum entropy inverse reinforcement learning.
\newblock In \emph{Association for the Advancement of Artificial Intelligence}, 2008.

\end{thebibliography}

\clearpage
\appendix
\appendix

\section*{Appendix}

% \vspace{-0.5em}
\section{Pseudocode of DIFO}
% \vspace{-0.5em}
\label{appendix:algo}

% In this section, we present the pseudocode of our proposed method, DIFO.
\begin{algorithm}[H]
   \caption{Diffusion Imitation from Observation (DIFO)}
   \label{alg:DIFO}
   \begin{algorithmic}[1]
   \STATE {\bfseries Input:} 
   Expert demonstrations $\tau_E \sim \pi_E$, policy $\pi_A$ and diffusion model  $\mathcal{D}_{\phi}$    
        \WHILE{$\pi_A$ not converges}
            \STATE Sample agent transitions $\tau_i \sim \pi_A$ by interacting with the environment
            \STATE Calculate discriminator loss $\mathcal{L}_D$ with \myeq{eq:overall_objective}
            \STATE Update discriminator parameters $\phi$ with $\nabla \mathcal{L}_D$  
            \STATE Calculate IRL reward $r_{\phi}(\mathbf{s},\mathbf{s'})$ with \myeq{eq:GAIL_reward}
            \STATE Update $\pi_A$ with respect to $r_{\phi}$ with any RL algorithm
        \ENDWHILE
    \end{algorithmic}
\end{algorithm}
% \vspace{-2.0em}

\section{Environment \& task details}
% \vspace{-0.5em}
\label{appendix:environment}
The environments we used for our experiments in \cref{sec:experiment} are from Gymnasium~\citep{gymnasium_robotics2023github, towers_gymnasium_2023}. 
We list names, version numbers, dimensions of state and action spaces in \cref{table:env-description}.
All environments we used are continuous action spaces.

\begin{table*}[!htbp]
\centering
% \small
\caption[Enviroments]{\textbf{Enviroments.} Detailed setting of each Task.}
\scalebox{0.9}{\begin{tabular}{@{}ccccc@{}}\toprule
\textbf{Task}
& ID
& Observation space 
& Action space 
& Fixed horizon \\
% & Success Rate & Episode Length & Success Rate & Episode Length & Success Rate & Episode Length & Return\\
\cmidrule{1-5}
\pointmaze
& PointMaze\_Medium-v3
& 8 
& 2 
& True \\
% \cmidrule{1-4}
\antmaze
& AntMaze\_UMaze-v4
& 31
& 8 
& True \\
% \cmidrule{1-4}
\fetchpush
& FetchPush-v2
& 31
& 4 
& True \\
% \cmidrule{1-4}
\adroitdoor
& AdroitHandDoorCustom-v1
& 39 
& 28 
& True \\
% \cmidrule{1-4}
\walker
& Walker2d-v4
& 17 
& 6 
& True \\
\frankakitchen{}
& FrankaKitchen-v1
& 59 
& 9
& True \\
% \cmidrule{1-4}
\carracing
& CarRacing-v2
& $96\times96\times3$
& 3 
& False \\
\drawerclose{}
& N/A
& $64\times64\times3$
& 4
& True \\
\bottomrule
\end{tabular}
}
\label{table:env-description}
\end{table*}
Notably, we fix the horizon to prevent biased information about termination~\citep{kostrikov2018discriminatoractorcritic} in all environments except {\carracing} since the goal of {\carracing} is to complete the track as fast as possible instead of goal completion. Notice that though the objective of {\walker} is also not goal completion, the termination signal provides additional information about tumbling.

\myparagraph{Preprocessing}
In {\carracing}, we preprocess the observation state by applying frame skipping with a factor of 2, resizing, and converting the image to grayscale, resulting in a $64\times64$ matrix. Finally, we stack 2 frames to form the input state.

We list the details of the expert demonstration datasets we used in \cref{sec:experiment} and how we collect them in \cref{table:env-expertnum}.
All of our expert data incorporates stochasticity to enhance diversity in trajectories. We add a small amount of noise to the experts' actions, providing stochasticity and multimodality in expert behaviors.

\begin{table*}[!htbp]
\centering
% \small
\caption[Enviroments]{\textbf{Expert observations.} Detailed information on collected expert observations in each Task.}
\scalebox{0.9}{\begin{tabular}{@{}cccc@{}}\toprule
\textbf{Task} 
& \# of trajectories
& \# of transitions
& Algorithm or Retrieved Source
\\
% & Success Rate & Episode Length & Success Rate & Episode Length & Success Rate & Episode Length & Return\\
\cmidrule{1-4}
\pointmaze
& 60
& \num{36000}
& D4RL~\citep{fu2020d4rl}
\\
% \cmidrule{1-4}
\antmaze
& 100
& \num{7000}
& Minari~\citep{minari2023minari}
\\
% \cmidrule{1-4}
\fetchpush
& 50
& \num{2500}
& Own SAC expert
\\
% \cmidrule{1-4}
\adroitdoor
& 50
& \num{10000} 
& D4RL~\citep{fu2020d4rl}
\\
% \cmidrule{1-4}
\walker
& 1
& \num{1000}
& Own SAC expert
\\
\frankakitchen
& 5
& \num{300}
& D4RL~\citep{fu2020d4rl}
\\
% \cmidrule{1-4}
\carracing
& 1
& \num{340}
& Own PPO expert
\\
\drawerclose
& 1
& \num{100}
& Own Script expert
\\
\bottomrule
\end{tabular}
}
\label{table:env-expertnum}
\end{table*}

In \cref{sec:visualized_reward}, we introduce a toy environment \sine.
We generate \num{25000} state-only transition $(s,\,s')$, with transition function:
\begin{align}
s' = \sin{6\pi s}+s+\mathcal{N}(0, \,0.05^2)
\end{align}
where $\mathcal{N}(\mu, \,\sigma^2)$ is a normal distribution noise with the mean value $\mu=0$ and the standard deviation $\sigma=0.05$.
\section{Optimizing \texorpdfstring{$\mathcal{L}_{\text{MSE}}$}{LMSE} with agent data}
\label{appendix:agent_mse}

\begin{figure}[htbp]
    \centering
    \captionsetup[subfigure]{aboveskip=0.05cm,belowskip=0.05cm}
     \begin{subfigure}[b]{0.245\textwidth}
         \centering
         \includegraphics[width=\textwidth]{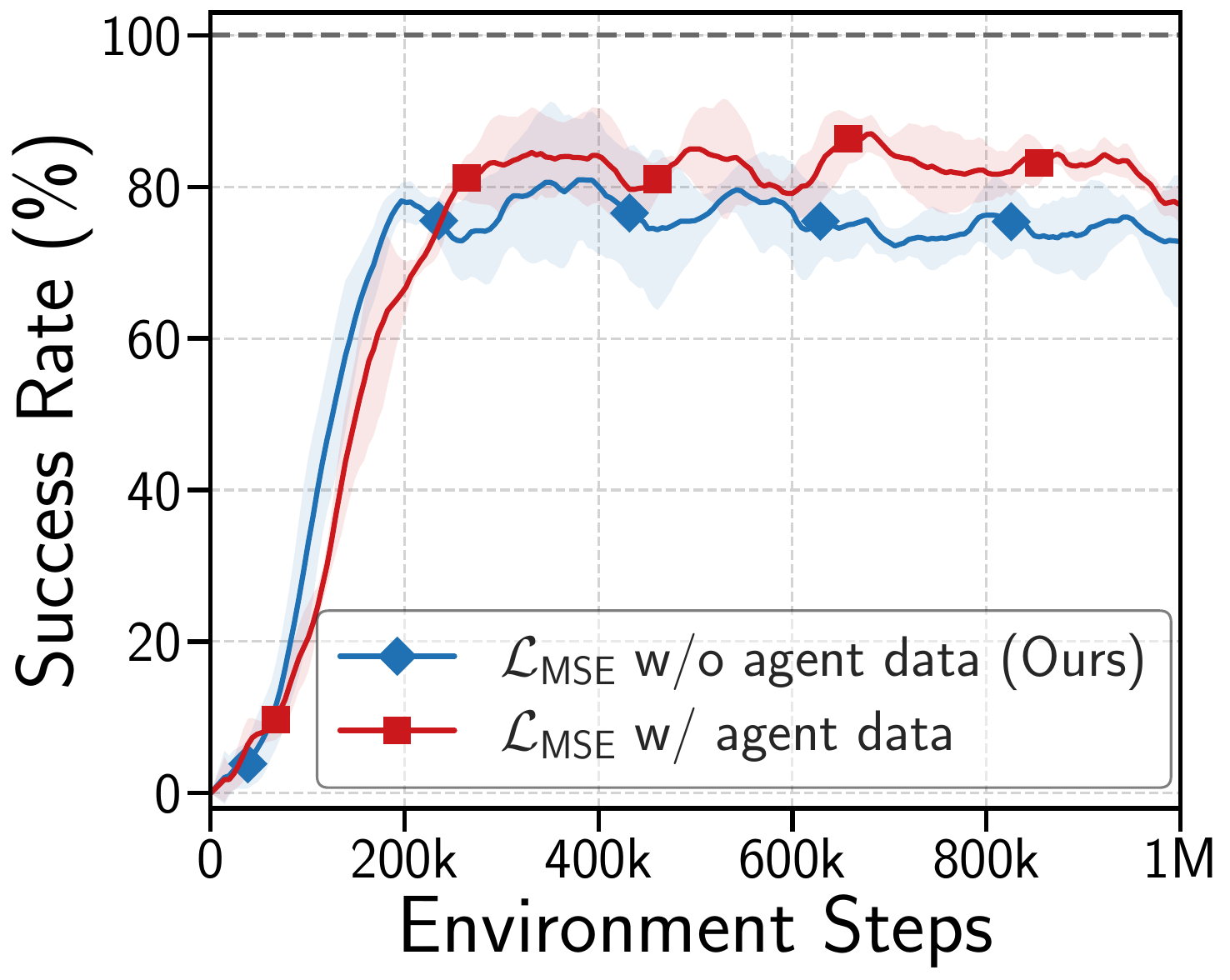}
         \caption{\pointmaze}
         \label{fig:agent_point_maze}
     \end{subfigure}
     \begin{subfigure}[b]{0.245\textwidth}
         \centering
         \includegraphics[width=\textwidth]{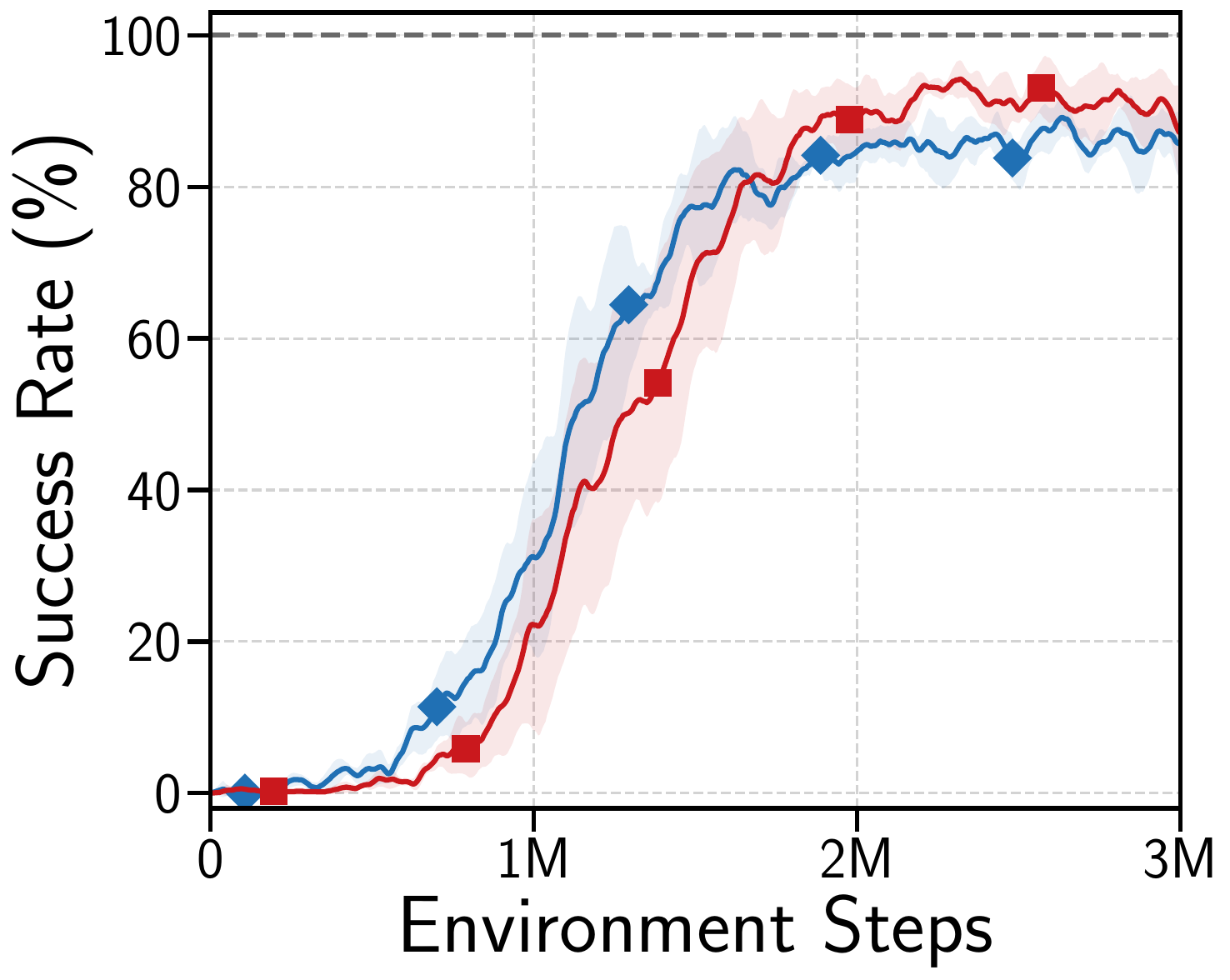}
         \caption{\antmaze}
         \label{fig:agent_ant_maze}
     \end{subfigure}
     \begin{subfigure}[b]{0.245\textwidth}
         \centering
         \includegraphics[width=\textwidth]{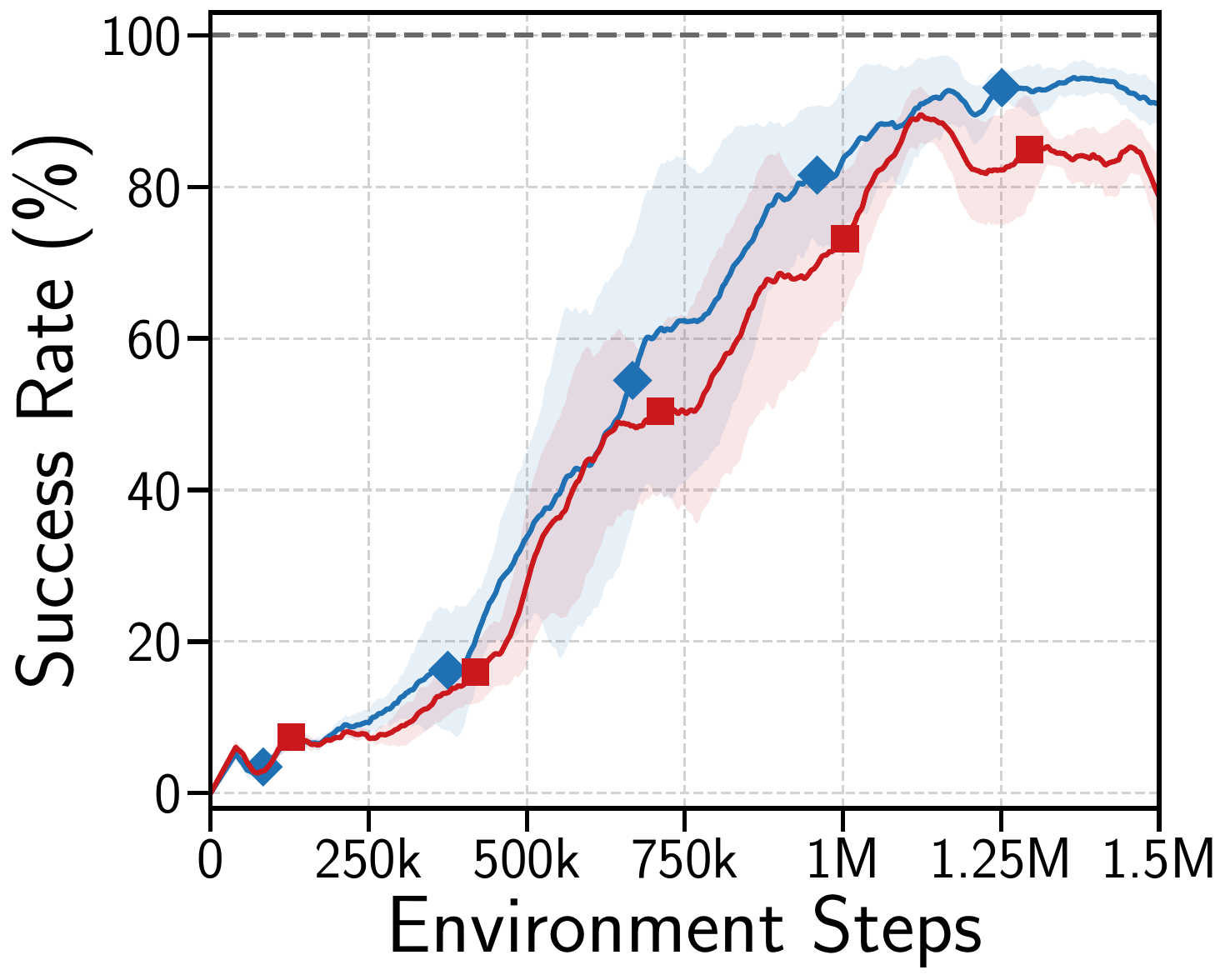}
         \caption{\fetchpush}
         \label{fig:agent_fetch_push}
     \end{subfigure}
     \begin{subfigure}[b]{0.245\textwidth}
         \centering
         \includegraphics[width=\textwidth]{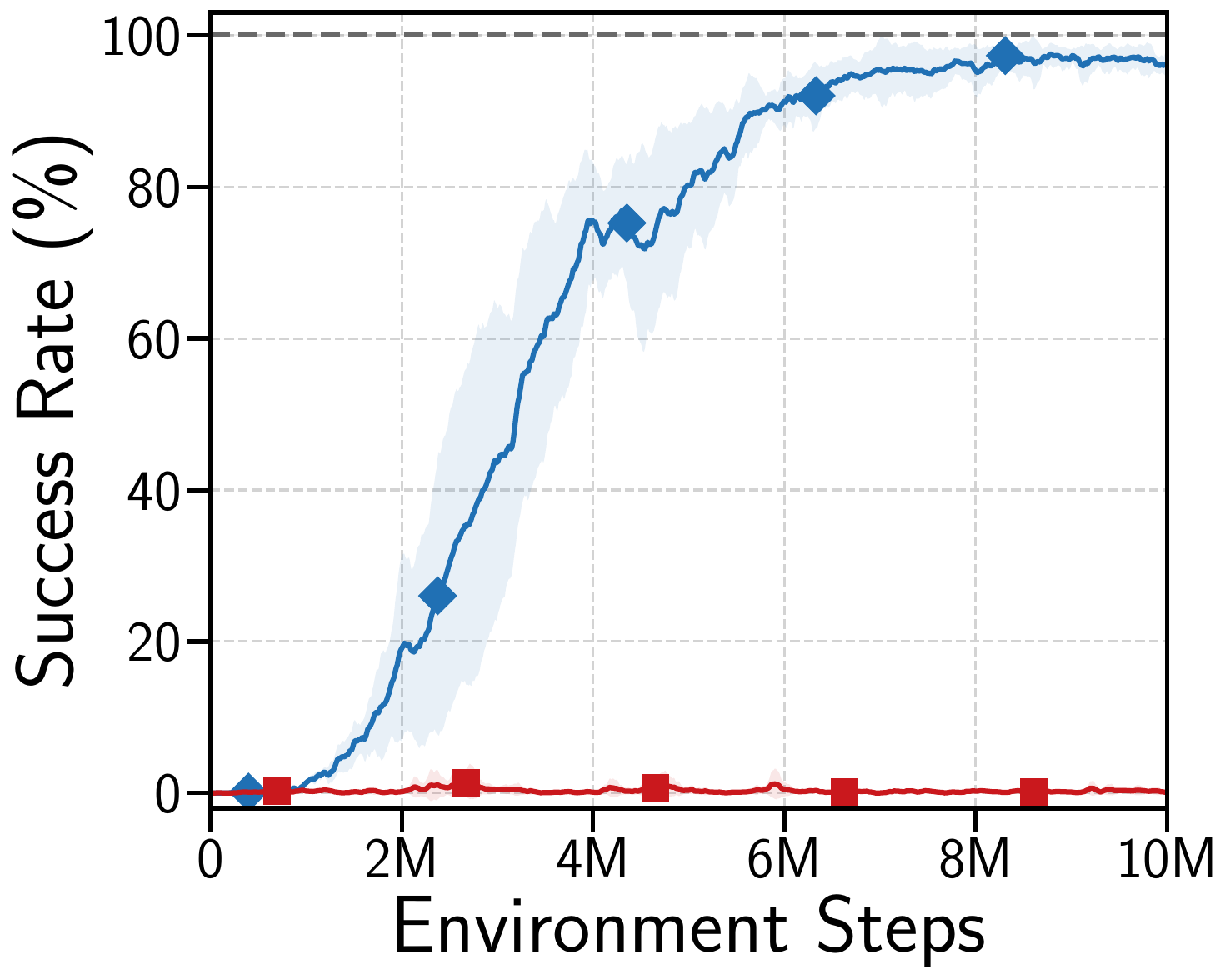}
         \caption{\adroitdoor}
         \label{fig:agent_door}
     \end{subfigure}
     
     \begin{subfigure}[b]{0.245\textwidth}
         \centering
         \includegraphics[width=\textwidth]{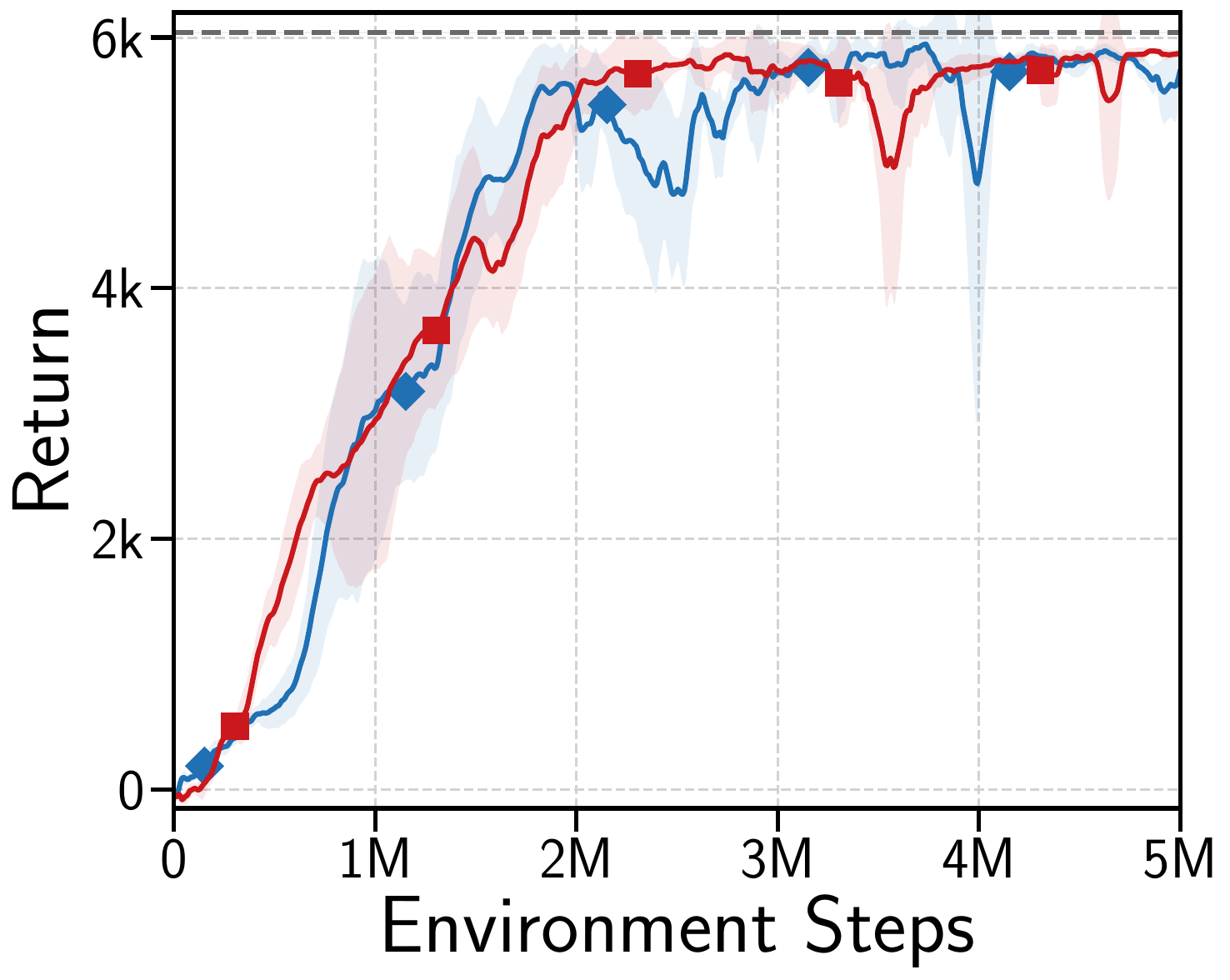}
         \caption{\walker}
         \label{fig:agent_walker}
     \end{subfigure}
     \begin{subfigure}[b]{0.245\textwidth}
         \centering
         \includegraphics[width=\textwidth]{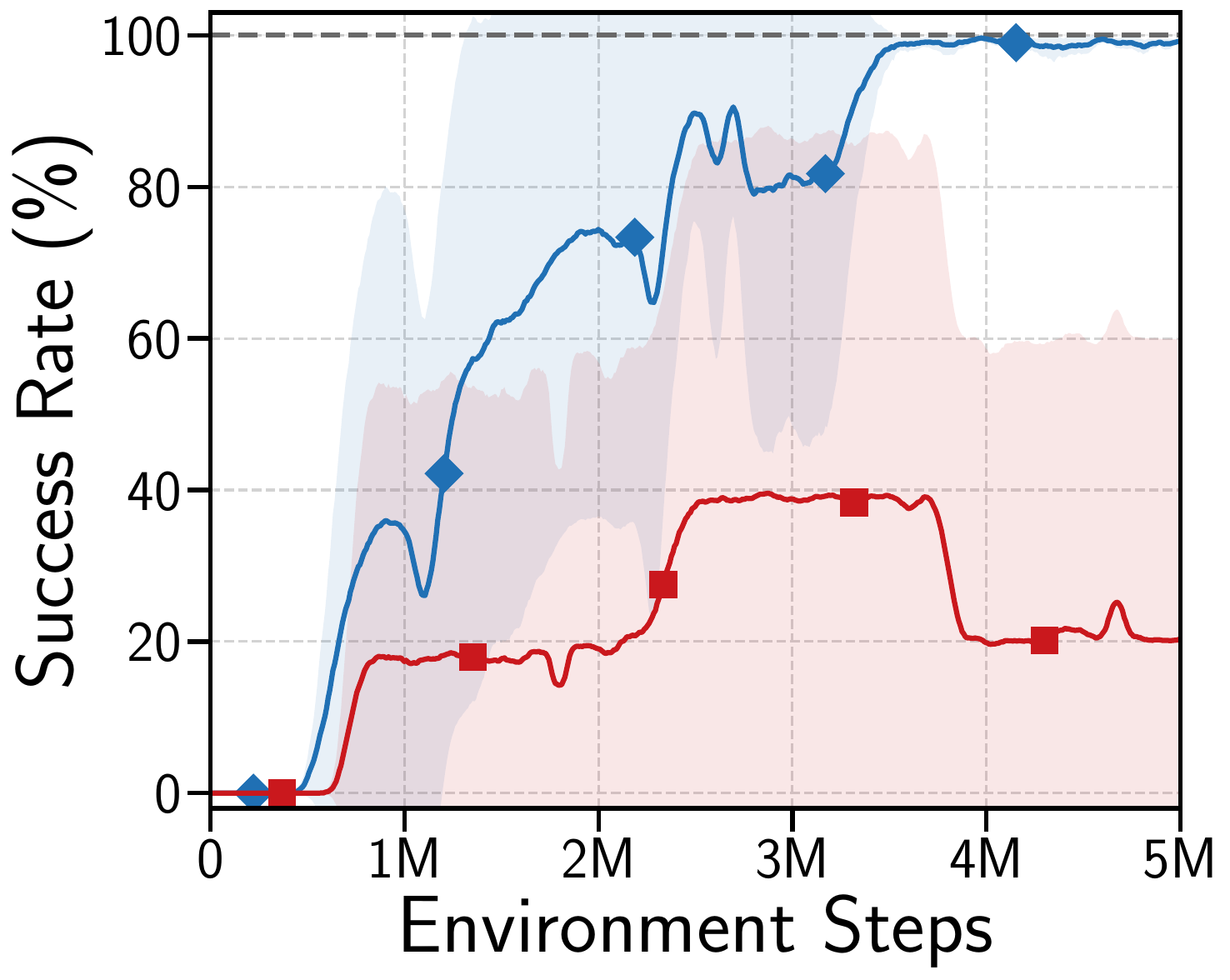}
         \caption{\frankakitchen}
         \label{fig:agent_franka}
     \end{subfigure}     
     % \hfill
     \begin{subfigure}[b]{0.245\textwidth}
         \centering
         \includegraphics[width=\textwidth]{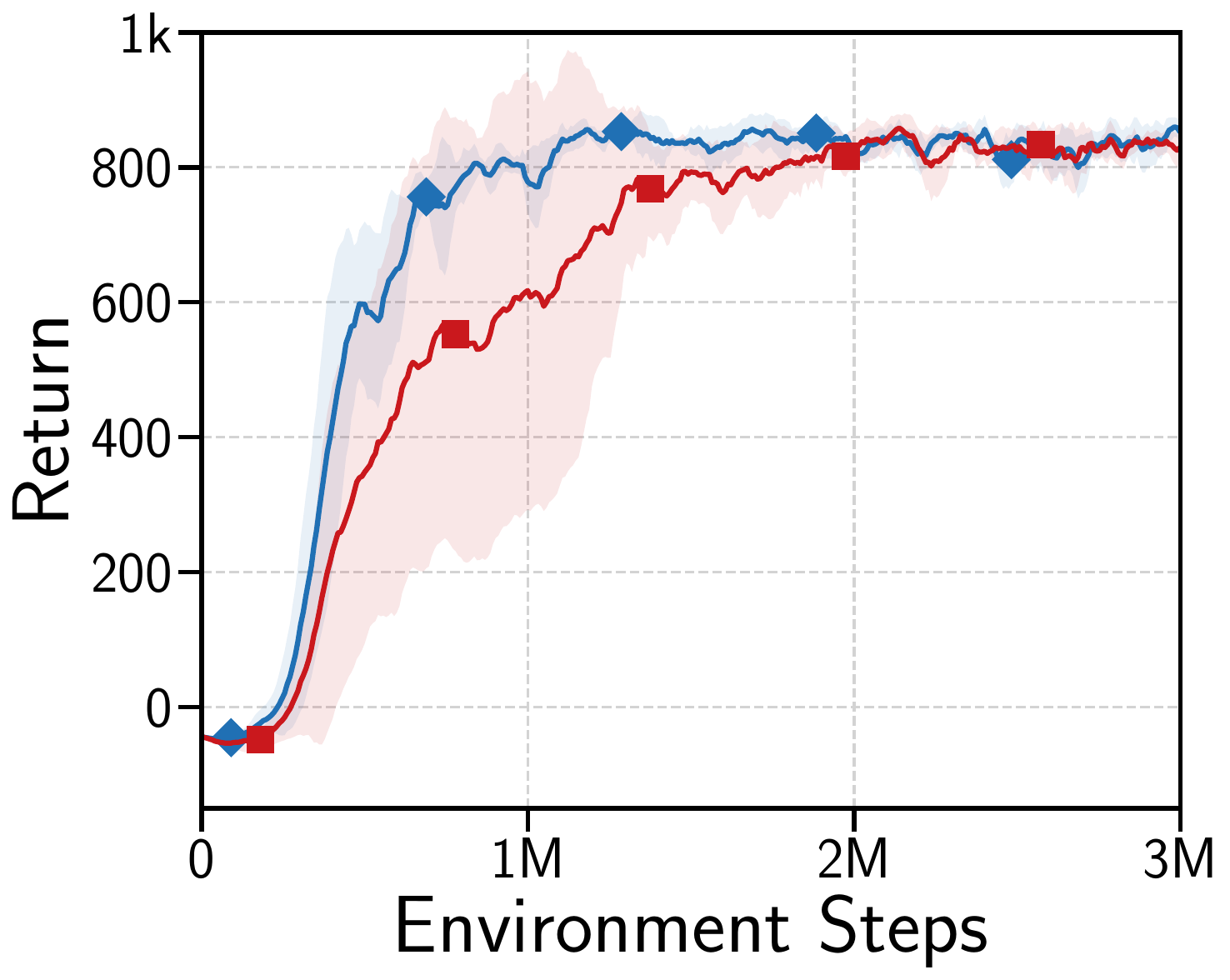}
         \caption{\carracing}
         \label{fig:agent_car_racing}
    \end{subfigure}
    \begin{subfigure}[b]{0.245\textwidth}
         \centering
         \includegraphics[width=\textwidth]{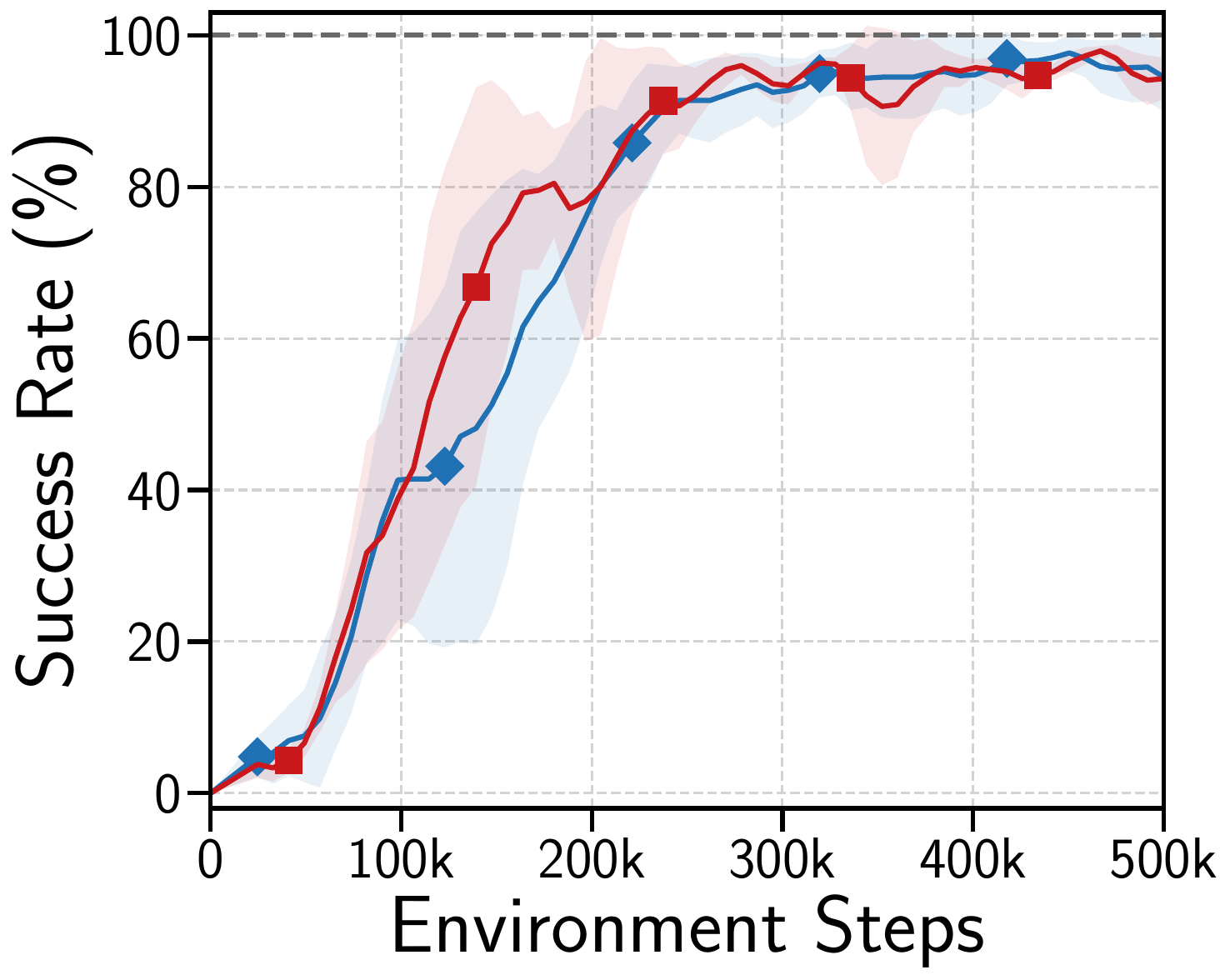}
         \caption{\drawerclose}
         \label{fig:agent_drawerclose}
     \end{subfigure}
    \caption{\myparagraph{Optimizing $\mathcal{L}_{\text{MSE}}$ with and without agent data}
    We evaluate optimizing $\mathcal{L}_{\text{MSE}}$ with and without agent data in all tasks.
    The results show that optimizing $\mathcal{L}_{\text{MSE}}$ with agent data (red) leads to slower convergence and less stable performance.}
    \label{fig:agent_mse}
\end{figure}

Our method optimizes $\mathcal{L}_{\text{MSE}}$ (Eq.~\ref{eq:mse_objective}) to approximate the ELBO only using expert demonstrations.
To investigate the effect of optimizing this MSE loss using agent data, we experiment with optimizing $\mathcal{L}_{\text{MSE}}$ with and without agent data on all tasks.
The results are reported in \cref{fig:agent_mse}.
We found that optimizing $\mathcal{L}_{\text{MSE}}$ with agent data can lead to slower and unstable convergence, especially in tasks with larger state and action spaces, \eg \adroitdoor{}, where optimizing $\mathcal{L}_{\text{MSE}}$ leads to a $0\%$ success rate.
We hypothesize that optimizing $\mathcal{L}_{\text{MSE}}$ leads to unstable training because, during the early stage of training, the agent policy changes frequently and generates diverse transitions.
This diversity leads to a consistent distribution shift for the diffusion model, making the diffusion model unstable to learn.
As a result, the overall performance can be less stable.
Hence, our method is designed to only optimize $\mathcal{L}_{\text{MSE}}$ using expert demonstrations.

\section{Stochastic environment}
\label{appendix:stochastic_environment}
\begin{wrapfigure}[10]{R}{0.4\textwidth}
    \vspace{-2.2cm}
% \begin{figure}[!h]
     \centering
     \includegraphics[width=0.9\linewidth]{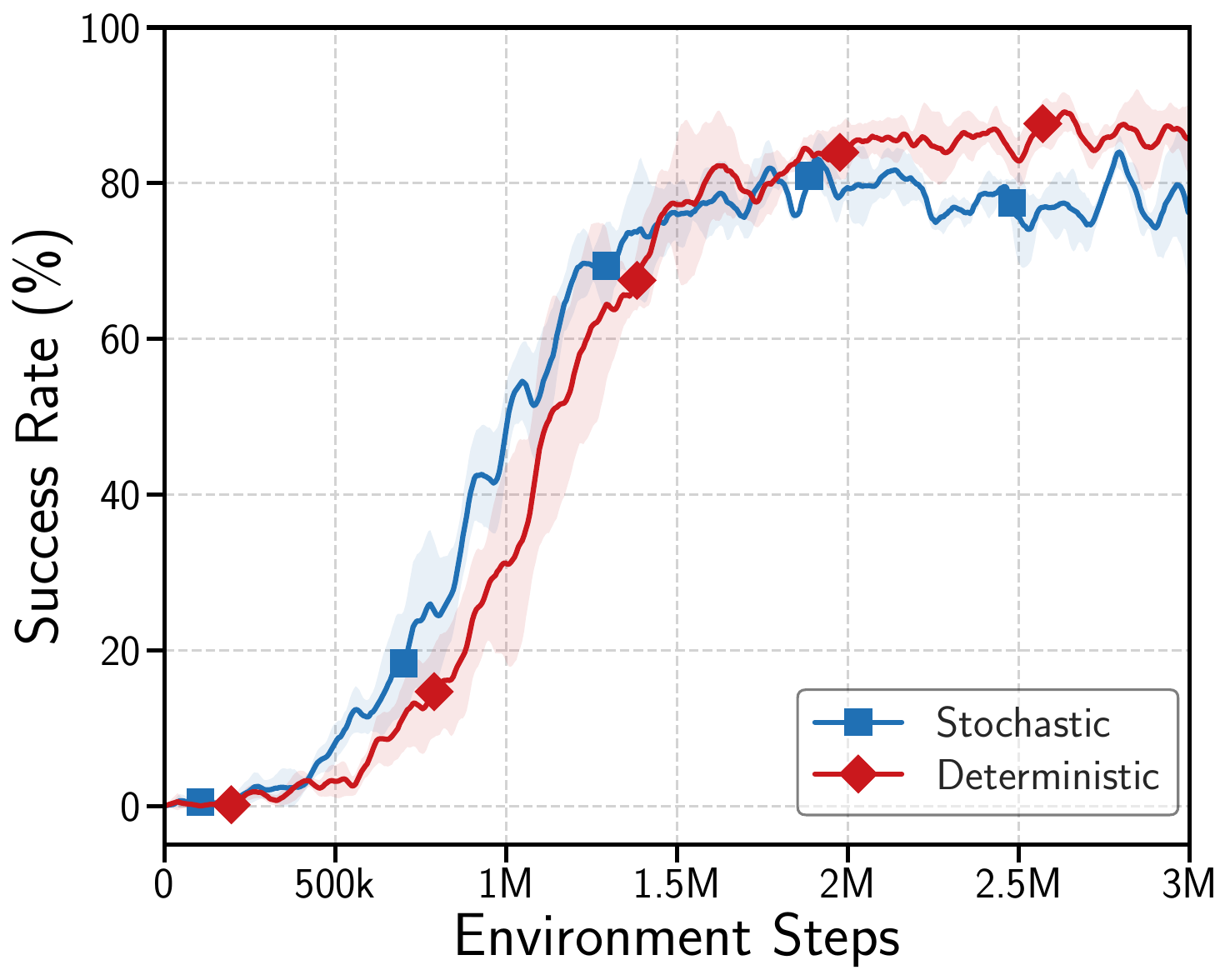}
     % \caption{\myparagraph{Stochastic \antmaze} We apply an addition Gaussian noise $0.5\mathcal{N}(0, 1)$ to each action to create a stochastic \antmaze{} environment. Our method maintains robust performance under this large stochasticity.}
     \caption{\myparagraph{Stochastic environment} We apply addition Gaussian noises to actions to create a stochastic \antmaze{}. Our method \method{} maintains robust performance under large stochasticity.}
     \label{fig:stochastic_ant_maze}
% \end{figure}
\end{wrapfigure}

We create a stochastic version of \antmaze{} environment where Gaussian noise is added to the agent's actions before they are applied in the environment.
The magnitude of the noise is 0.5, resulting in the actual action taken in the environment would be $action = action + 0.5 \mathcal{N}(0,1)$. 
Given the action space of this environment is $[-1, 1]$, this represents a high level of stochasticity. 

\cref{fig:stochastic_ant_maze} shows that the performance of our method remains robust even under such high stochasticity, indicating our model's ability to adapt to stochastic environments effectively.

\section{Full trajectories generations of {\pointmaze}}
\label{appendix:full_maze}

\begin{figure}[!h]
    \centering
    \includegraphics[width=\linewidth]{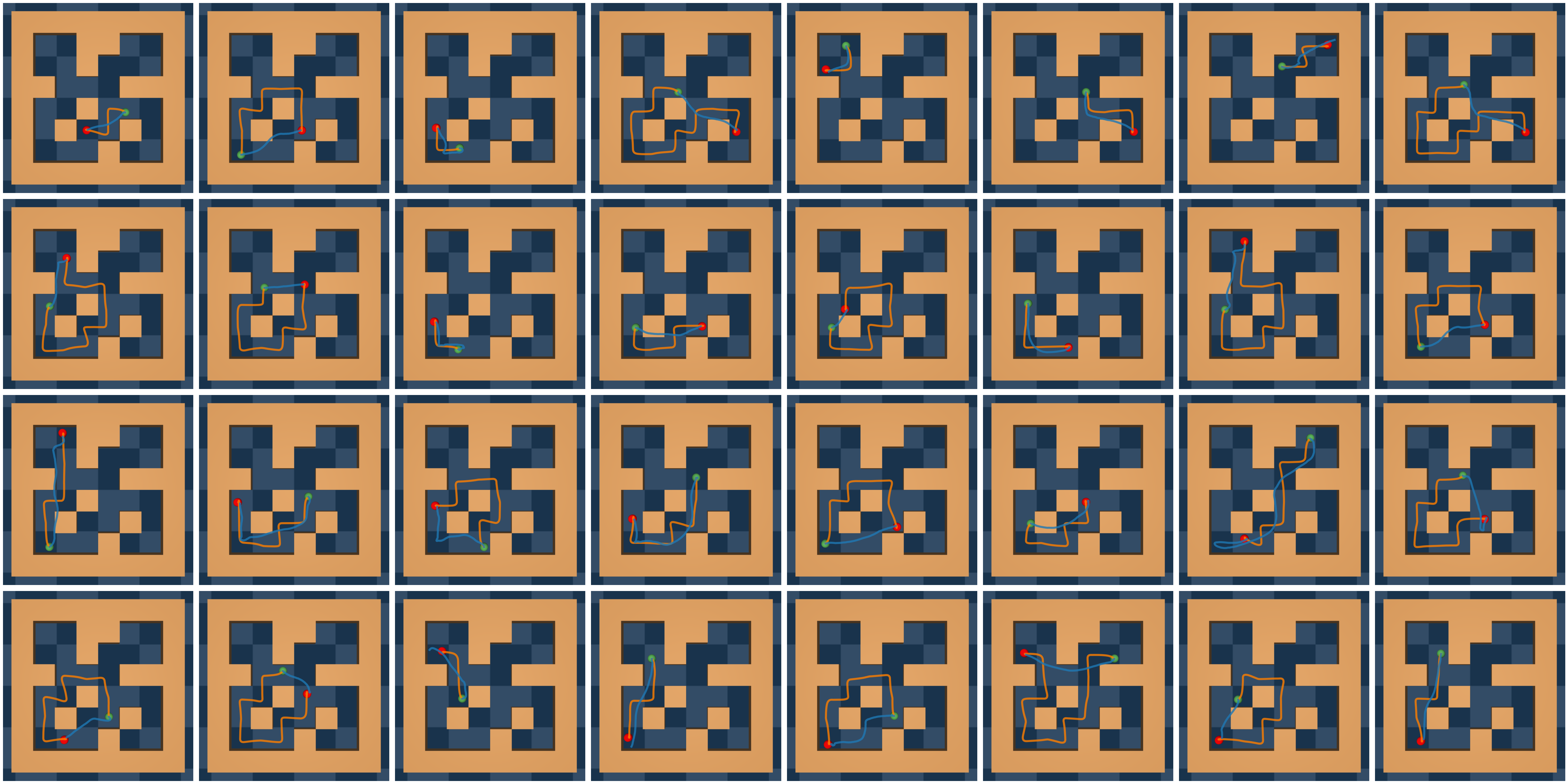}
    \caption{\myparagraph{Full result of generated trajectories under {\pointmaze}}
    The blue trace represents the generated trajectory and the orange trace represents the corresponding expert trajectory.
    DIFO can accurately model expert behavior and generate successful trajectories.}
    \label{fig:maze_traj_appendix}
\end{figure}

More results of the {\pointmaze} trajectory generation experiment in \cref{sec:generate_data} can be found in \cref{fig:maze_traj_appendix}.

\section{The stability of rewards}
\label{appendix:sample_timestep}

To further verify the stability of the rewards under different numbers of denoising step samples, we present the standard deviation to mean ratio of the rewards from 500 computations results
in Table~\ref{table:sample-timestep}. The values are averaged from a batch of 4096 transitions. The result shows that sampling a single denoising step is enough to produce a stable reward. 

\begin{table*}[!h]
\centering
% \small
\caption[Sample Timesteps]{\textbf{Reward ratio.} Standard deviation to mean rewards ratio over 500 computations, averaged from 4096 transitions. $n$ is the number of denoising step samples to compute the reward.}
\scalebox{0.9}{\begin{tabular}{@{}ccccc@{}}\toprule
\textbf{Learning Progress}
& $n=1$
& $n=2$ 
& $n=5$ 
& $n=10$ \\
% & Success Rate & Episode Length & Success Rate & Episode Length & Success Rate & Episode Length & Return\\
\cmidrule{1-5}
$20\%$  & 0.323 & 0.237 & 0.294 & 0.246 \\
$40\%$  & 0.234 & 0.199 & 0.206 & 0.230 \\
$60\%$  & 0.201 & 0.175 & 0.157 & 0.206 \\
$80\%$  & 0.157 & 0.152 & 0.150 & 0.190 \\
$100\%$ & 0.142 & 0.133 & 0.145 & 0.161 \\
\bottomrule
\end{tabular}
}
\label{table:sample-timestep}
\end{table*}

\section{Training details}
\label{appendix:training}

\begin{table*}[htbp]
\centering
\small
\caption[Hyperparameters]{\textbf{Hyperparameters.}
The overview of the hyperparameters used for all the methods in every task. We abbreviate "Discriminator" as "Disc." in this table.}
%\ra{1.3}
% \scalebox{0.6}{
\resizebox{\columnwidth}{!}{
\begin{tabular}{cccccccc}\toprule
\textbf{Method} & \textbf{Hyperparameter} & 
\pointmaze &
\antmaze &
\fetchpush &
\adroitdoor &
\walker &
\carracing \\ 
% & Success Rate & Episode Length & Success Rate & Episode Length & Success Rate & Episode Length & Return\\
% \cmidrule{1-8}
\midrule
\midrule
RL Algorithm & &SAC &SAC &SAC &SAC &PPO &PPO \\ 
\midrule
\midrule
% \hhline{========} 
% \cmidrule{1-8}
\multirow{4}{*}{BC} 
& Batch Size  &128 &128 &128 &128  &128  &128 \\
& Learning Rate &0.0003 &0.0003 &0.0003 &0.0001  &0.0003  &0.0003 \\
& L2 Weight  & 0.0001 &0.0001 &0.0001 &0.0001 &0.0001 &0.0001 \\
& Training Steps &\num{100000} &\num{500000} &\num{500000} &\num{500000}  &\num{500000}  &\num{500000} \\
% & \# Steps  & & & &  &  & \\
\cmidrule{1-8}
\multirow{6}{*}{BCO} 
& Batch Size &128 &128 &128 &128  &128  &128 \\
& Learning Rate  &0.0003 &0.0003 &0.0003 &0.0001  &0.0003  &0.0001 \\
& L2 Weight & 0.001 &0.001 &0.001 &0.001 &0.001 &0.001 \\
& Entropy Weight & 0.0001 &0.0001 &0.0001 &0.0001 &0.0001 &0.0001 \\
& $\alpha$ &0.01 &0.2 &1.0 &5.0  &1.0  &5.0 \\
% & Inv Hidden Dimensions &128 &128 &128 &128 &1024 &128 \\
% & $T$  & & & &  &  & \\
& \# Environment Steps  &\num{1000000} &\num{3000000} &\num{1500000} & \num{10000000} &\num{5000000} & \num{3000000} \\
\cmidrule{1-8}
\multirow{6}{*}{GAIfO} 
& Batch Size &64 &64 &64 &64  &64  &64 \\
& Disc. Learning Rate &0.0001 &0.0001 &0.0001 &0.0001  &0.00001  &0.0001 \\
% & Policy Learning Rate & & & & & & \\
& Disc. Hidden Dimensions & 128 & 128 & 128 & 128 & 128 & 128 \\
& Disc. Hidden Layers & 5 & 5 &5 &4 &5 &5 \\
& Disc. Buffer Size &\num{1000000} &\num{1000000} &\num{1000000} &\num{1000000} &\num{1000000} &\num{1000000} \\
& \# Environment Steps  &\num{1000000} &\num{3000000} &\num{1500000} & \num{10000000} &\num{5000000} & \num{3000000} \\
\cmidrule{1-8}
\multirow{8}{*}{WAIfO} 
& Batch Size &64 &64 &64 &64  &64  &64 \\
& Disc. Learning Rate  &0.0001 &0.0001 &0.0001 &0.0001  &0.00001  &0.0001 \\
% & Policy Learning Rate  & & & & & & \\
& Disc. Buffer Size &\num{1000000} &\num{1000000} &\num{1000000} &\num{1000000} &\num{1000000} &\num{1000000} \\
& Disc. Hidden Dimensions & 128 & 128 & 128 & 128 & 128 & 128 \\
& Disc. Hidden Layers & 5 & 5 &5 &4 &5 &5 \\
& Regularize Epsilon &0.001 &0.001 &0.001 &0.001  &0.001  &0.001 \\
& Regularize Function &$L_2$ &$L_2$ &$L_2$ &$L_2$  &$L_2$  &$L_2$ \\
& Gradient Clipping &1 &1 &1 &1 &1 &1 \\
& \# Environment Steps  &\num{1000000} &\num{3000000} &\num{1500000} & \num{10000000} &\num{5000000} & \num{3000000} \\
\cmidrule{1-8}
\multirow{6}{*}{AIRLfO} 
& Batch Size &64 &64 &64 &64  &64  &64 \\
& Disc. Learning Rate  &0.0001 &0.0001 &0.0001 &0.00001  &0.0001  &0.00001 \\
% & Policy Learning Rate  & & & &  &  & \\
& Disc. Buffer Size &\num{1000000} &\num{1000000} &\num{1000000} &\num{1000000} &\num{1000000} &\num{1000000} \\
& Disc. Hidden Dimensions & 128 & 128 & 128 & 128 & 128 & 128 \\
& Disc. Hidden Layers & 5 & 5 &5 &4 &5 &5 \\
& \# Environment Steps  &\num{1000000} &\num{3000000} &\num{1500000} & \num{10000000} &\num{5000000} & \num{3000000} \\
\cmidrule{1-8}
\multirow{4}{*}{DePO} 
& Batch Size               & 64& 64& 64& 64& 64& 64\\
& IDM. Batch Size               & 256& 256& 256& 256& 256& 256\\
& IDM. Learning Rate      & 0.0003& 0.0003& 0.0003& 0.0003& 0.0003& 0.0003\\
& \# Environment Steps     &\num{1000000} &\num{3000000} &\num{1500000} & \num{10000000} &\num{5000000} & \num{3000000} \\
\cmidrule{1-8}
\multirow{5}{*}{IQ-Learn} 
& Batch Size  &256 &256 &256 &256  &256  &256 \\
& Actor Learning Rate   &3$\times10^{-5}$ &3$\times10^{-5}$ &3$\times10^{-5}$ &3$\times10^{-5}$ &3$\times10^{-5}$ &3$\times10^{-5}$ \\
& Critic Learning Rate  &3$\times10^{-4}$ &3$\times10^{-4}$ &3$\times10^{-4}$ &3$\times10^{-4}$ &3$\times10^{-4}$ &3$\times10^{-4}$ \\
& Entropy Coefficient & 0.01 &0.01 &0.01 &0.01 &0.01 &0.01 \\
& \# Environment Steps  &\num{1000000} &\num{3000000} &\num{1500000} & \num{10000000} &\num{5000000} & \num{3000000} \\
\cmidrule{1-8}
\multirow{5}{*}{OT} 
& Batch Size               & 256& 256& 256& 256& 256& 256\\
& Reward Scale      & 100& 100& 100& 100& 100& 100\\
& \# Expert Samples        & 10& 10& 10& 10& 10& 10\\
& Learning Rate & 0.0003& 0.0003& 0.0003& 0.0003& 0.0003& 0.0003\\
& \# Environment Steps     &\num{1000000} &\num{3000000} &\num{1500000} & \num{10000000} &\num{5000000} & \num{3000000} \\
\cmidrule{1-8}
\multirow{12}{*}{\method{} (Ours)}
& Batch Size &64 &64 &64 &64  &64  &64 \\
& Disc. Learning Rate  &0.0001 &0.0001 &0.0001 &0.0001  &0.0001  &0.0001 \\
& Disc. Buffer Size &\num{1000000} &\num{1000000} &\num{1000000} &\num{1000000} &\num{1000000} &\num{1000000} \\
% & Policy Learning Rate  & & & &  &  & \\
& BCE Weight $\lambda_\text{BCE}$ &0.1 &0.01 &0.01 &0.001  &0.01  &0.1 \\
& MSE Weight $\lambda_\text{MSE}$ &1 &1 &1 &1  &1  &1 \\
& $\lambda_\sigma$ &10 &10 &10 &10  &10  &10 \\
& U-Net Units & (256, 256, 256) & (256, 256, 256) & (256, 256, 256) & (256, 256, 256) & (256, 256, 256) & (256, 256, 256) \\
& Embedding Dimension & 128 & 128 & 128 & 128 & 32& 128 \\
& Denoising Sample Range & (250, 750) & (250, 750)& (250, 750)& (250, 750)& (250, 750)& (250, 750) \\
& Denoising Timestep & 1000 & 1000 & 1000 & 1000 & 1000 & 1000 \\
% & Disc. training timesteps (for each update) &1000 &1000 &1000 &1000 &1000 &1000 \\ 
& Logit Scale  &10 &10 &10 &10 &10  &10 \\
& \# Environment Steps  &\num{1000000} &\num{3000000} &\num{1500000} & \num{10000000} &\num{5000000} & \num{3000000} \\
\cmidrule{1-8}
\multirow{12}{*}{DIFO-Uncond} 
& Batch Size &64 &64 &64 &64  &64  &64 \\
& Disc. Learning Rate  &0.0001 &0.0001 &0.0001 &0.0001  &0.0001  &0.0001 \\
& Disc. Buffer Size &\num{1000000} &\num{1000000} &\num{1000000} &\num{1000000} &\num{1000000} &\num{1000000} \\
% & Policy Learning Rate  & & & &  &  & \\
& BCE Weight $\lambda_\text{BCE}$ &0 &0 &0 &0 &0 &0 \\
& MSE Weight $\lambda_\text{MSE}$ &1 &1 &1 &1 &1 &1 \\
& $\lambda_\sigma$ &10 &10 &10 &10 &10 &10 \\
& U-Net Units & (256, 256, 256) & (256, 256, 256) & (256, 256, 256) & (256, 256, 256) & (256, 256, 256) & (256, 256, 256) \\
& Embedding Dimension & 128 & 128 & 128 & 128 & 128& 128 \\
& Denoising Sample Range & (250, 750) & (250, 750)& (250, 750)& (250, 750)& (250, 750)& (250, 750) \\
& Denoising Timestep & 1000 & 1000 & 1000 & 1000 & 1000 & 1000 \\
% & Disc. training timesteps (for each update) &1000 &1000 &1000 &1000 &1000 &1000 \\ 
& Logit Scale  &1 &1 &1 &1 &1 &1 \\
& \# Environment Steps  &\num{1000000} &\num{3000000} &\num{1500000} & \num{10000000} &\num{5000000} & \num{3000000} \\
\cmidrule{1-8}
\multirow{12}{*}{DIFO-NA} 
& Batch Size &64 &64 &64 &64  &64  &64 \\
& Disc. Learning Rate  &0.0001 &0.0001 &0.0001 &0.0001  &0.0001  &0.0001 \\
& Disc. Buffer Size &\num{1000000} &\num{1000000} &\num{1000000} &\num{1000000} &\num{1000000} &\num{1000000} \\
% & Policy Learning Rate  & & & &  &  & \\
& BCE Weight $\lambda_\text{BCE}$ &0.01 &0.01 &0.01 &0.01  &0.01  &0.01 \\
& MSE Weight $\lambda_\text{MSE}$ &1 &1 &1 &1  &1  &1 \\
& $\lambda_\sigma$ &10 &10 &10 &10  &10  &10 \\
& U-Net Units & (512, 512, 512, 512) & (512, 512, 512, 512) & (512, 512, 512, 512) & (512, 512, 512, 512) & (512, 512, 512, 512) & (256, 256, 256) \\
& Embedding Dimension & 128 & 256 & 256 & 256 & 256& 128 \\
& Denoising Sample Range & (250, 750) & (250, 750)& (250, 750)& (250, 750)& (250, 750)& (250, 750) \\
& Denoising Timestep & 1000 & 1000 & 1000 & 1000 & 1000 & 1000 \\
% & Disc. training timesteps (for each update) &1000 &1000 &1000 &1000 &1000 &1000 \\ 
& Logit Scale  &10 &10 &10 &10 &10  &10 \\
& \# Environment Steps  &\num{1000000} &\num{3000000} &\num{1500000} & \num{10000000} &\num{5000000} & \num{3000000} \\ 
\bottomrule
\end{tabular}
}
\label{table:hyperparameter}
\end{table*}

Our codebase inherits from \textit{Imitation}~\citep{gleave2022imitation}, an open-source imitation learning framework based on \textit{Stable Baselines3}~\citep{stable-baselines3}. 
We then describe the implementation of each baseline:
\begin{itemize}
    \item GAIfO, AIRLfO, and BC follow the original implementation of \textit{Imitation}.
    \item IQ-Learn and OT are modified from SAC in \textit{Stable Baselines3} with the loss function borrow from the official implementations~\citep{garg2021iq, papagiannis2022imitation}.
    \item We implemented our own BCO, WAIfO, and DePO.
\end{itemize}

\subsection{Model architectures}
\label{appendix:training:architecture}
\myparagraph{Vector-based observation space}
For tasks with 1D vectorized state space, we implement a U-Net with MLP layers as our backbone of the diffusion model. The conditions are applied on every layer.

\myparagraph{Image-based observation space}
For tasks with image observations, we implemented a 2D U-Net as the backbone of the diffusion model with the \textit{diffusers} package provided by \citet{von-platen-etal-2022-diffusers}, which originally proposed by \citet{ronneberger2015u}. The model contains 3 down-sampling blocks and 3 up-sampling blocks. Each block consists of 2 convolution residual layers, with group normalization applied using 4 groups. The conditions are applied on every layer.

\subsection{Hyperparameters}
\label{appendix:training:hp}
The hyperparameters of the policies and discriminators employed for all methods across all tasks are listed in Table~\ref{table:hyperparameter} and \ref{table:rl-hyperparameter}. 
We use Adam as the optimizer for all the experiments.

\begin{table*}[ht]
\centering
\small
\caption[SAC \& PPO training hyperparameters]{\textbf{SAC \& PPO training parameters.}}
\scalebox{0.75}{
\begin{tabular}{@{}ccccccccc@{}}\toprule
\textbf{Method} & \textbf{Hyperparameter} & h \\
% & Success Rate & Episode Length & Success Rate & Episode Length & Success Rate & Episode Length & Return\\
\cmidrule{1-3}
\multirow{3}{*}{SAC} 
& Learning Rate &0.0003 \\
& Batch Size  &256 \\
& Discount Factor $\gamma$  &0.99 \\
\cmidrule{1-3}
\multirow{9}{*}{PPO} 
& Learning Rate &0.0001 \\
& Batch Size &128 \\
& Discount Factor $\gamma$ &0.99 \\
& Clip &0.001 \\
& GAE $\lambda$ & 0.95 \\
& Value Function Coefficient & 0.5 \\
& Entropy Coefficient &0 \\
& Maximum gradient norm &0.6 \\
& \# Epochs &5 \\
\bottomrule
\end{tabular}
}
\label{table:rl-hyperparameter}
\end{table*}
\section{Computational resources}
\label{appendix:compute}

We used the workstations listed in \cref{table:computational-resouces}.
Our method takes approximately 3 hours for each task.
The shortest task is {\pointmaze} which takes around 1 hour.
The longest task is {\carracing} which takes around 6 hours.
As for reproducing all results including the baselines, it takes about 2000 GPU hours to run in series.
\begin{table*}[!ht]
\centering
% \small
\caption[Computational Resources]{\textbf{Computational resources.}}
\scalebox{0.9}{\begin{tabular}{@{}cccc@{}}\toprule
\textbf{Workstation} 
& CPU
& GPU
& RAM
\\
% & Success Rate & Episode Length & Success Rate & Episode Length & Success Rate & Episode Length & Return\\
\cmidrule{1-4}
Workstation 1
& Intel Xeon w7-2475X
& NVIDIA GeForce RTX 4090 x 2
& 125 GiB
\\
% \cmidrule{1-4}
Workstation 2
& Intel Xeon w5-2455X
& NVIDIA RTX A6000 x 2
& 125 GiB
\\
% \cmidrule{1-4}
Workstation 3
& Intel Xeon W-2255
& NVIDIA GeForce RTX 4070 Ti x 2
& 125 GiB
\\
% \cmidrule{1-4}
Workstation 4
& Intel Xeon W-2255
& NVIDIA GeForce RTX 4070 Ti x 2
& 125 GiB
\\
\bottomrule
\end{tabular}
}
\label{table:computational-resouces}
\end{table*}

\section{Limitations}
\label{appendix:limitation}

This work presents a novel learning from observation framework, {\method}, integrating a diffusion model into the AIL framework. 
Despite that {\method} achieves encouraging results in various domains, there are still some limitations.
Firstly, our method takes state-only transitions $(s,s')$ as the reward function, while the underlying optimal reward function could be in the form $r(s, a, s')$, where dynamics is involved.
This may lead to sub-optimal performance in tasks with delayed effects on actions.
Secondly, our method assumes the state spaces of the agent and expert are identical as they share the same model, which limits cross-embodiment applications to some extent.
Thirdly, our method highly relies on expert demonstrations, the presence of sub-optimal demonstrations may adversely impact the performance.
Finally, due to the learning nature focused on discrimination, {\method} may not incorporate well with environmental rewards even if they are accessible.

\section{Broader impact}
\label{appendix:impact}

Experimental results demonstrate that DIFO exhibits data efficiency, generalizability, and resistance to noisy environments, thereby enhancing its suitability for real-world applications. Learning from observation enables its deployment in scenarios where action data is costly or inaccessible.

However, our method inherits the nature of Adversarial Imitation Learning. One significant concern is the potential negative impact on safety when deployed in real-world settings, as the exploration process may lead to unsafe actions. Additionally, imitation learning may capture and reinforce bias present in expert demonstrations, causing trapping in sub-optimal behaviors. These issues highlight the necessity of further research focused on reducing these negative impacts.

% \clearpage
% \input{text/checklist}

\end{document}